\documentclass[11pt]{article}

\usepackage{graphicx}
\usepackage{amssymb}
\usepackage{amsthm}
\usepackage{booktabs}
\usepackage{rotating}
\usepackage{multirow}
\usepackage{eurosym}
\usepackage{amsfonts}
\usepackage{amsmath}

\usepackage{mathtools} 
\usepackage{tabularx}
\usepackage{umoline}
\usepackage{caption}
\usepackage{subcaption}
\usepackage{color}
\usepackage{url}
\usepackage{authblk}
\usepackage[english]{babel}

\usepackage{pbox}
\usepackage{footnote}
\usepackage{multirow}
\usepackage{changepage}
\usepackage{tablefootnote}
\usepackage{xcolor}

\usepackage[margin = 2cm]{geometry}
\usepackage{booktabs}
\usepackage{adjustbox} 
\usepackage{multirow} 
\usepackage{subcaption, epsfig} 
\usepackage{graphicx}
\usepackage{subcaption}
\usepackage{hyperref}
\usepackage[table,xcdraw]{xcolor}

\usepackage{breakurl}
\usepackage{hyperref}
\hypersetup{breaklinks=true}

\usepackage[natbibapa]{apacite}

\usepackage{rotating}
\usepackage{lscape}
\usepackage{float}
\floatstyle{plaintop}
\restylefloat{table} 
\usepackage{booktabs}
\usepackage{textcomp}
\usepackage{siunitx}
\usepackage{multirow}
\usepackage{makecell}
\usepackage{longtable}
\usepackage{setspace}
\usepackage{amssymb}
\usepackage{amsmath}
\usepackage{booktabs}
\usepackage{adjustbox} 
\usepackage{multirow} 
\usepackage{subcaption, epsfig} 
\usepackage{graphicx}
\usepackage{subcaption}
\usepackage{hyperref}
\usepackage{utfsym}
\usepackage{multicol}
\usepackage{rotating}

\usepackage{array}
\usepackage{natbib, ragged2e}
\usepackage{appendix}
\usepackage{pdflscape}
\usepackage{adjustbox}

\begin{document}
\newcolumntype{B}[1]{>{\small\hspace{0pt}\RaggedRight\bfseries}p{#1cm}}
\newcolumntype{C}[1]{>{\RaggedRight}p{#1cm}}

\title{Revealing Geography-Driven Signals in Zone-Level Claim Frequency Models: An Empirical Study using Environmental and Visual Predictors\footnote{\scriptsize NOTICE: This is the author´s version of a submitted work for publication. Changes resulting from the publishing process, such as editing, corrections, structural formatting, and other quality control mechanisms may not be reflected in this document. Changes may have been made to this work since it was submitted for publication. This work is made available
under a Creative Commons BY license. \textcopyright CC-BY-NC-ND}}

\author[1, 2]{Sherly Alfonso-Sánchez}
\author[1]{Cristi\'{a}n Bravo}
\author[1]{Kristina G. Stankova}

\affil[1]{Department of Statistical and Actuarial Sciences, Western University, 1151 Richmond Street, London, Ontario, N6A 5B7, Canada.}
\affil[2]{Departamento de Matemáticas, Universidad Nacional de Colombia. Ave Cra 30 \#45-03, Edificio 404. Bogotá, Colombia.}
\date{}
\maketitle
\begin{abstract}
Geographic context is often considered relevant for assessing motor insurance risk, yet public actuarial datasets typically provide limited location identifiers, constraining how this information can be incorporated in claim-frequency models. This study examines how geographic information from alternative data sources can be incorporated into actuarial models for Motor Third Party Liability (MTPL) claim prediction under such constraints.

Using the BeMTPL97 dataset, we adopt a zone-level modeling framework and evaluate predictive performance on postcodes not observed during training. Geographic information is introduced through two channels: environmental indicators from OpenStreetMap (2014) and CORINE Land Cover 2000, and orthoimagery from 1995 released by the Belgian National Geographic Institute for academic use. We evaluate the predictive contribution of coordinates, environmental features, and image embeddings across three baseline models: generalized linear models (GLMs), regularized GLMs, and gradient-boosted trees, while raw imagery is modeled using convolutional neural networks.

Our results show that augmenting actuarial variables with constructed geographic information improves predictive accuracy. Across experiments, both linear and tree-based models benefit most from combining coordinates with environmental features extracted at 5 km scale, while smaller neighborhoods also improve baseline specifications. Generally, image embeddings do not improve performance when structured environmental features are available; however, when such features are absent, pretrained vision-transformer embeddings enhance accuracy and stability for regularized GLMs. Our results show the predictive value of geographic information in zone-level MTPL frequency models depends less on model complexity than on how geography is represented, and illustrate that geographic context can be incorporated despite limited individual-level spatial information.
\end{abstract}

\begin{keywords}
Frequency prediction, Multimodal learning, Remote sensing, Alternative data.
\end{keywords}
\section{Introduction}
\label{sec:Introduction}
The rapid growth of new data sources has encouraged insurers to assess whether incorporating such information can improve profitability or the overall customer experience. Recent studies show that alternative data can improve different forms of risk assessment and decision-support in insurance operations. For example, \cite{oskarsdottir2022social} used the social network of a claim to improve fraud detection in motor insurance, demonstrating that relational features add predictive value beyond claim-specific attributes. Similarly, \cite{dubey2018smart} showed the usefulness of text data in underwriting by extracting embeddings from unstructured documents (e.g., personal details, medical history) using Natural Language Processing to support automated risk assessment. \cite{perez2024automated} used damaged vehicle images for automated car damage detection, showing that an ensemble of YOLOv5-based detectors outperforms state-of-the-art models in both accuracy and speed, replacing labor intensive manual inspections (i.e., manual analysis of images and physical investigation) with scalable automated processing that enables faster turnaround and an improved customer experience.

The growing availability of alternative data has been accompanied by rapid advances in artificial intelligence, allowing insurers to analyze complex risk profiles beyond traditional tabular information. These developments have expanded the range of data sources used in insurance, including geopolitical indicators, social media signals, Geospatial Information Systems (GIS), and real-time information streams \citep{jaiswal2023impact}. The rise of InsurTech firms has further accelerated this trend, making modern AI and machine-learning techniques accessible even in lower-income or emerging markets and improving customer engagement \citep{gupta2022artificial}. However, the adoption of these technologies also raises regulatory and ethical concerns. Current regulatory frameworks emphasize the need for transparent and explainable models to maintain public trust and to prevent discriminatory outcomes arising from opaque or biased data and algorithms \citep{bhattacharya2025ai}

Despite these technological advances, the use of AI within auto insurance is largely focused on a narrow set of tasks. According to a recent survey \citep{bhattacharya2025ai}, most applications of artificial intelligence in auto insurance focus on behavioral risk modeling through telematics \citep{henckaerts2022added, gao2022boosting, ayuso2019improving}, automated damage detection \citep{rababaah2023investigation, fouad2023automated}, fraud detection \citep{DING202551,benedek2023traditional}, or operational efficiency \citep{DONG202517, ayuso2019improving}. However, the use of AI or other forms of \textit{alternative data} for Motor Third-Party Liability (MTPL), which is the mandatory coverage that compensates third parties sustaining injuries and property damage caused by the policyholder, has received far less attention. With regard to the incorporation of geographic information, some researchers have enriched locations (e.g., latitude and longitude coordinates) with demographic or socioeconomic attributes. In this setting, \cite{tufvesson2019spatial} modeled frequency and severity in vehicle hull damage policies from a private insurer including as geographic data the area centroid coordinates and a neighborhood structure and improved with location related demographic variables, and \citep{blier2022geographic} earned spatial embeddings from census-derived socioeconomic variables to support geographic rate making. Although these studies demonstrate the potential value of geographic enrichment, they are based on either spatial adjacency structures or demographic context. In contrast, our study examines a different dimension of geography: the built environment, which refers to observable physical characteristics of infrastructure in the area surrounding each zone, including the structure of the road network and the composition of land-use, as well as the visual appearance of this area captured through images. To the best of our knowledge, these forms of geographic information have not been evaluated for the prediction of the frequency of MTPL claims at the zone-level, leaving this question empirically unexplored.

We use the publicly available beMTPL97 dataset, which contains Belgian MTPL policies and their associated postcodes or coordinates. In our research, we model the frequency of MTPL claims at the zone level. This choice is motivated both by the nature of the available data and by actuarial considerations. In the beMTPL97 dataset, individual policy records exhibit coarse geographic granularity, with many policies sharing the same postcode, which limits the ability to exploit meaningful location variation at the policy level. Aggregating to zones provides more stable exposure and claim counts and avoids sparsity issues in frequency modeling. Although modern auto insurance increasingly incorporates personalized or telematics-based scores, geographic variation in the built environment and traffic context remains an important determinant of MTPL risk. In this context, the use of the beMTPL97 dataset is primarily motivated by its public availability, while the dataset reflects a specific empirical context, the objective of this study is not tied to this particular dataset, but rather to develop and illustrate a general methodological framework and the types of insights it can provide. 

Zone-level modeling therefore offers a practical and interpretable framework for investigating the role of geography in MTPL claim-frequency prediction. Within this framework, geographic representations based on publicly observable built environment characteristics and images at the postcode level do not directly encode individual level socio-economic attributes, though they may act as indirect proxies; aggregation at the zone level mitigates ethical concerns commonly associated with the use of geographical variables. This scheme allows geographic context to be incorporated at the zone level alongside individual level actuarial variables, while preserving relevance for pricing applications. In this context, our study is guided by the following research questions.

\begin{enumerate}
\item How can useful geographic information be constructed or acquired from alternative data sources for MTPL insurance data when the available information is limited to policyholder coordinates or postcodes?
\item To what extent does incorporating both existing and constructed geographic information improve the prediction of the number of claims at the zone-level beyond traditional actuarial variables?
\item How sensitive are the resulting models to the choice of spatial scale, such as using narrower or broader neighborhoods when extracting environmental and visual features?
\end{enumerate}

The remainder of this article is organized as follows. Section~\ref{sec:LiteratureR} provides a detailed review of the related literature. 
Section~\ref{sec:Methodology}, describes the methodology created, including the introduction of background concepts, the description of the beMTPL97 data set, and the procedures used to construct geographic information from external data sources. Section~\ref{sec:Results} presents and discusses the empirical findings. Finally, Section~\ref{sec:Conclusions} concludes the article and outlines the limitations and implications of our research.

\section{Literature Review}
\label{sec:LiteratureR}
\subsection{Geographic Data Usages in Insurance}
The use of geographic information in insurance, understood as data that contain an implicit or explicit reference to a location \citep{ISO19109_2022}, has been explored in several areas. For example, \cite{blier2022geographic} rely on demographic variables linked to census polygons to represent how populations are distributed throughout space, using features such as population age structure, household composition, education, participation in the labor force, commuting patterns, and income distributions, while explicitly excluding protected characteristics such as ethnicity, language, or immigration status, to construct geographic embeddings. They apply these embeddings to predict fire counts, home insurance claims, and car accidents, and report lower bias and variance than standard spatial interpolation models. In a different direction, \cite{tufvesson2019spatial} analyze spatial variation in claims of hull damage using a Conditional Autoregressive model. Their geographic inputs are limited to area centroids and an insurer-supplied neighborhood structure that accounts for adjacent areas and natural barriers. Their results show improvements in predictive accuracy compared to a traditional GLM. Neither of these works incorporates information about the built environment or visual features.

Other studies consider geographic information extracted from images. \cite{kita2019google} link Google Street View (GSV) images with the home addresses of more than 20,000 auto insurance policyholders. They use convolutional neural networks to extract indicators of neighborhood condition, housing quality, and visible socioeconomic characteristics, and show that these image based measures improve accident risk prediction over the company’s internal model.

A similar approach is used in \cite{nguyen2022google}, where GSV images are combined with Intermountain Healthcare patient records. Deep learning models classify elements of the built environment, such as street greenery, crosswalks, and visible utility wires. These indicators are aggregated at the neighborhood level and are linked to several health outcomes. Interestingly, they show that people living in communities with high levels of street greenness and a lower proportion of single-family detached housing have 10 \ to 27\% lower diabetes, hypertension, and obesity adjusting by insurance status, white race, hispanic ethnicity, religion, marital status, and area deprivation index. In the context of home insurance, \cite{blier2024representation} construct a representation-learning framework that cleans and censors raw GSV images and then learns image embeddings by pre-training on property characterization tasks such as predicting construction year or number of floors. The authors find that these embeddings help predict the frequency of perils such as theft and wind.

Geographic information has also been used in agricultural insurance. \cite{islam2024damage} combine Landsat-8 satellite imagery with detailed land use maps to estimate flood related damage to boro rice crops in Bangladesh. They generate spatial damage maps at the subdistrict level and use them to estimate financial losses and appropriate premium levels. Their study shows that remote sensing can significantly improve crop damage assessment.

Finally, some applications use geographic information for operational tasks. \cite{asabere2024geo} develop a Geographic Information System (GIS) based platform called \textit{Geo-Insurance}, where a GIS is understood as ``a system specifically designed to capture, store, analyze, manage and present spatial and geographical data'' \citep{longley2015geographic}. The platform integrates geodatabases with web mapping tools to build a location-based recommendation system that helps customers identify nearby insurance companies that offer specific services. Although their work does not focus on risk modeling, it shows that geographic data and GIS tools can also support customer decision-making in insurance service selection.

Table~\ref{tab:summary_Geouses} contains a summary of some related work investigating the use of geographic data in insurance 
\begin{table}[h]
\begin{adjustbox}{max width=\textwidth}
\begin{tabular}{@{}llll@{}}
\toprule \hline
Authors (Year) & \begin{tabular}[c]{@{}l@{}}Insurance \\ Line\end{tabular} & \begin{tabular}[c]{@{}l@{}}Geographic \\ Information\end{tabular}                                    & Data source                                                                                                                                                                                                                                                        \\ \hline \midrule
\citet{blier2022geographic}              & Auto, fire and home.                                      & \begin{tabular}[c]{@{}l@{}}Spatial embeddings \\ from census data\end{tabular}                       & \begin{tabular}[c]{@{}l@{}}* Canadian census data\\ * Montréal's Collisions routi\`eres open dataset\\ * City of Toronto’s open dataset Fire Incidents\\ * Historical losses of home insurance contracts \\   \ \ \ from a Canadian P\&C insurance company.\end{tabular} \\
\hline
\cite{tufvesson2019spatial}              & Car                                                       & \begin{tabular}[c]{@{}l@{}}Area centroid coordinates\\ and a neighbourhood \\ structure\end{tabular} & * Commercial P\&C provider (Insightone)                                                                                                                                                                                                                          \\
\hline
\\
\cite{kita2019google}                    & Car accidents                                             & Google Street View Images                                                                            & * Poland motor insurance data set.                                                                                                                                                                                                                               \\
                                                          &                                                           &                                                                                                      &                                                                                                                                                                                                                                                                  \\ 
                                                          \hline
                                                          \\
                                                          \citet{nguyen2022google}
    & Healthcare                                                & Google Street View Images                                                                            & * Intermountain Healthcare                                                                                                                                                                                                                                       \\
    \hline
    \\
    \cite{blier2024representation} & Home                                                      & Google Street View Images                                                                            & * Canadian home insurance data set \\
    \hline
    \\
      \cite{islam2024damage}  & Crop                                                      & \begin{tabular}[c]{@{}l@{}}Remote sensing\\ images\end{tabular}                                   & \begin{tabular}[c]{@{}l@{}}* Landsat-8\\ * High-resolution land use land cover \\     \ \ \ map of a Survey of Bangladesh (SoB)\end{tabular} \\
    \hline
    \\
      
    \cite{asabere2024geo}  & Customer experience                                       & \begin{tabular}[c]{@{}l@{}}Geographic coordinates \\ of the potential client\end{tabular}            & \begin{tabular}[c]{@{}l@{}}* From client's mobile or desktop \\ * Some insurance companies in Accra, Ghana.\end{tabular} \\  
                                                    \bottomrule
\end{tabular}
\end{adjustbox}
\caption{Summary of related work}
\label{tab:summary_Geouses}
\end{table}

Across these studies, geographic information has shown promise in several insurance applications, but almost none of this work focuses on MTPL. Existing research on auto insurance either looks at telematics or uses images for private car risk, not liability claims. As a result, it is still unclear whether the environmental or visual features around a location are useful for predicting the frequency of MTPL claims. This opens a clear opportunity for our study, which tests these forms of geographic information at the zone level using the beMTPL97 dataset.

\subsection{Research in Motor Third Party Liability}

Research that examines MTPL claim frequency prediction using ML models remains limited but has gained attention in recent years. Using the widely known French MTPL dataset, available in the CASdatasets R package \citep{CASdatasets2024}, \cite{noll2020case} showed that tree-based models and neural networks (NN) can capture interactions between predictors that traditional GLMs cannot. To test the robustness of these findings on a different dataset, \cite{burka2021modelling} evaluated GLMs, Generalized Additive Models (GAMs), Random Forests (RF), and NN on a Hungarian MTPL data set. They reported that all models performed well, with the best results obtained from a weighted ensemble that combines predictions from the individual models.

Further work using the French MTPL dataset includes \cite{seyam2025predicting}, who compared Poisson GLMs, decision trees (DT), and GAMs. They found that GAMs provided the strongest combination of predictive performance and interpretability. With the same objective of comparing classical and modern approaches for MTPL frequency estimation, \cite{OndřejVít2025CFEi} used a commercial Czech dataset. Their results show that GLMs, when joined with expert informed preprocessing, remain highly competitive, even against more complex models such as neural networks and hurdle models. Although the French MTPL dataset is widely used in the literature, it contains only limited geographic information, namely the policyholder’s region. Because our analysis relies on spatial variables, we instead use the Belgian dataset (beMTPL97, available in the CASdatasets R package), which provides postcode-level information linked to municipality centroids and, therefore, allows for a finer spatial resolution.

More recently, \cite{ibrahim2024evaluating} extended the evaluation of XGBoost for pricing applications using the beMTPL97 dataset. They assessed the performance of the models using out-of-sample error metrics and model lift, finding that Extreme Gradient Boosting (XGBoost) and Gradient Boosting Machine (GBM) performed best for frequency prediction, while GAMs were strongest for severity. In addition, \cite{holvoet2025neural} investigated deep learning approaches for both frequency and severity in four public CAS datasets, including three MTPL datasets (beMTPL97, French and Australian). Their results show that a standard feed-forward neural network does not outperform well designed GLMs or GBMs, which remains consistent with broader findings that deep learning often underperforms in tabular settings. However, they also showed that hybrid architectures that fuse gradient-boosted trees with NN layers, which represent a specific instance of the broader Combined Actuarial Neural Network (CANN) framework, can outperform standalone tree-based models when applied to tabular insurance data.

In general, these studies compare traditional actuarial methods with modern ML techniques. However, none of them expand the feature space using external geographic information. Our work differs by incorporating built-environment variables and visual predictors related to the policyholder’s area of residence, offering a new perspective on MTPL claim frequency modeling. Table~\ref{tab:summary_MTPL} summarizes some of the research related to the MTPL insurance line.

\begin{table}[h]
\begin{adjustbox}{max width=\textwidth}
\begin{tabular}{@{}llll@{}}
\toprule \hline
Authors (Year) & MTPL task                                                                         & Models employed                       & Data 
\\ \hline \midrule
                                                         \cite{noll2020case} & \begin{tabular}[c]{@{}l@{}}Claim frequency\\ prediction\end{tabular}              & * GLM,  Tree-based models, NN         & * French MTPL (publicly available)                                                                                           \\ \midrule
                                                          \cite{burka2021modelling} & \begin{tabular}[c]{@{}l@{}}Claim frequency \\ prediction\end{tabular}             & * GLM, GAM, RF, NN                    & * Commercial Hungarian MTPL                                                                                                \\ \midrule
                                                          \cite{seyam2025predicting} & \begin{tabular}[c]{@{}l@{}}Claim frequency \\ prediction\end{tabular}             & * GLM,  DT, GAM                       & * French MTPL (publicly available)                                                                                           \\ \midrule
                                                          \cite{OndřejVít2025CFEi} & \begin{tabular}[c]{@{}l@{}}Claim frequency \\ prediction\end{tabular}             & * GLM, Hurdle models, Feed forward NN & * Commercial Czezh dataset                                                                                                 \\ \midrule
                                                          \cite{ibrahim2024evaluating} & \begin{tabular}[c]{@{}l@{}}Claim frequency and\\ severity prediction\end{tabular} & * XGB, GBM, GAM                       & \begin{tabular}[c]{@{}l@{}}* Belgian MTPL beMTPL97\\   \ \ (publicly available)\end{tabular}                                     \\ \midrule
                                                          \cite{holvoet2025neural} & \begin{tabular}[c]{@{}l@{}}Claim frequency and\\ severity prediction\end{tabular} & * GLM, GBM, Feed forward NN, CANN     & \begin{tabular}[c]{@{}l@{}}* Belgian MTPL beMTPL97\\ * French MTPL\\ * Australian MTPL\\    \ \ (publicly availables)\end{tabular} \\ \bottomrule
\end{tabular}
\end{adjustbox}
\caption{Summary of related work in ML applications in MTPL tasks}
\label{tab:summary_MTPL}
\end{table}

\section{Methodology}
\label{sec:Methodology}
Claim frequency estimation plays a central role in actuarial science, supporting underwriting, risk assessment, pricing, and reserving. In this setting, the objective is to predict the number of claims that occur over a given period using historical information. For the MTPL insurance line, this task has traditionally been performed using classical statistical models such as GLMs \citep{qazvini2019validation, haberman1996generalized, mccullagh2019generalized}. These models are transparent, interpretable, and suitable for actuarial practice, as they allow the inclusion of exposure as an offset and accommodate standard counting distributions. However, as discussed in the previous section, several studies have explored ML models to improve predictive performance.

Following \cite{holvoet2025neural} and \cite{zail2019predictive}, we model claim counts using a Poisson distribution. Importantly, we do not replicate the model exactly as is done in these studies. Instead, we retain the Poisson assumption while introducing alternative geographic and environmental predictors. Specifically, if we denote by
$y_i$ the observed claim count for the postcode $i$ (after aggregating claims at the postcode level), assumed to follow a Poisson distribution, this is: 
\begin{align*}
    y_i \sim Po(\lambda_i),
\end{align*}
\noindent where the mean parameter $\lambda_i$ is given by:
\begin{align}
   \lambda_i = \exp\{\ln(e_i)+f(x_i)\}.\label{eq:mean}
\end{align}
Here, $e_i$ denotes the exposure (i.e., the time the policy was in force), and $x_i$ represents the vector of predictors for observation $i$. In this formulation, our study examines the impact of incorporating geographic predictors in addition to traditional actuarial features such as policyholder age, vehicle age, bonus–malus class (a score reflecting the policyholder's history of claims) or type of coverage, which are aggregated at the postcode level. That is, we compare models where these geographic features are included or excluded from $x_i$, and we evaluate different approaches to estimating the function $f$, ranging from parametric specifications (as in GLMs) to non-parametric ones (as in XGBoost and NN).

In the following, we introduce the main modeling approaches employed in this study. We then describe the dataset, the aggregation of the data set by zone-levels, and the construction of geographic features from external data sources.
\begin{itemize}

\item \textit{Poisson Generalized Linear Models}

In this setup we assumed $f(x_i)$ in (\ref{eq:mean}) to be a linear function given by:
\begin{align*}
     f(x_i) = \beta_0 + \beta_1 x_{i,1} + \cdots + \beta_p x_{i,p},
\end{align*}
where $x_i = [x_{i,1}, \dots, x_{i,p}]^{T}$ denotes the vector of predictors for observation $i$, and  
$\beta = [\beta_0, \beta_1, \dots, \beta_p]^{T}$ is the corresponding vector of coefficients. This corresponds to the standard GLM linear predictor under a log link.
\newpage
\item \textit{Extreme Gradient Boosting Models (XGB)}

In contrast to GLMs, XGB models model $f(x_i)$ in a non-parametric way using additive ensemble of regression trees. Following \cite{chen2016xgboost}, the prediction function can be written as:
\begin{align*}
     f(x_i) = \sum_{k=1}^{K} f_k(x_i), \ f_k \in \mathcal{F},
\end{align*}
where $x_i = [x_{i,1}, \dots, x_{i,p}]^{T}$ is the vector of predictors for observation $i$, $\mathcal{F}=\{f(x)=w_{q(x)}\}$ is the space of regression trees. Where $q:\mathbb{R}^{p}\rightarrow T$, and $w \in \mathbb{R}^{T}$ represents the structure of the tree ($T$ being the number of leaves of the tree) and leaf weights, respectively. XGB builds trees sequentially, each one correcting the residuals of the previous ensemble. This forward stagewise procedure enables the model to capture nonlinear effects and interactions among predictors that cannot be represented by a linear specification. For more details refer to \cite{chen2016xgboost} and \cite{hastie2009elements}.

\item \textit{Neural Networks, CNNs and ResNet18}:

Feed-forward neural networks model nonlinear relationships by applying a sequence of affine 
transformations followed by nonlinear activation functions. If $z^{(m)}$ denotes the output of layer $m$, 
then a standard dense layer has the form
\[
z^{(m)} = \sigma\!\left(W^{(m)} z^{(m-1)} + b^{(m)}\right),
\]
where $W^{(m)}$ and $b^{(m)}$ are the learnable weights and biases, and $\sigma(\cdot)$ is a nonlinear 
activation function such as the Rectified Linear Unit (ReLU) \citep{bengio2017deep} function. Stacking such layers allows the network 
to approximate complex functions, thus providing a flexible estimator of $f(x_i)$ in (\ref{eq:mean}).

For image inputs, Convolutional Neural Networks (CNNs) are more appropriate because they preserve spatial structure. Instead of treating each pixel as an independent input, a convolutional layer applies a small learnable filter across the image. Each filter responds to visual patterns such as edges, corners, or textures. 

CNNs differ from fully connected networks in several important ways. First, CNNs employ local connections, meaning each neuron is connected only to a small spatial region of the previous layer. This reduces the number of parameters and improves convergence speed. Second, CNNs make use of weight sharing, where a group of connections across different spatial locations uses the same set of weights (i.e., convolutional filters), further decreasing the total number of learnable parameters. Finally, CNNs apply downsampling operations (e.g., pooling) that reduce spatial resolution while retaining useful information \citep{LecunY.1998Glat, li2021survey}.

In this study, we use ResNet18 \citep{he2016deep} to extract visual features from image tiles. The architecture follows the standard specification: an initial 7×7 convolution with batch normalization and max pooling, followed by four residual stages containing two residual blocks each, with 64, 128, 256, and 512 output channels, respectively. Each residual block consists of two 3×3
convolutional layers with identity skip connections. Following the scheme  described in \cite{zhang2023dive}, Figure~\ref{fig:process_met} illustrates the ResNet blocks with and without a 1×1 convolution, the latter being used to adjust channel dimensions before the skip connection is added.
\begin{figure}[th!]
\centering
	\includegraphics[width=0.6\textwidth]{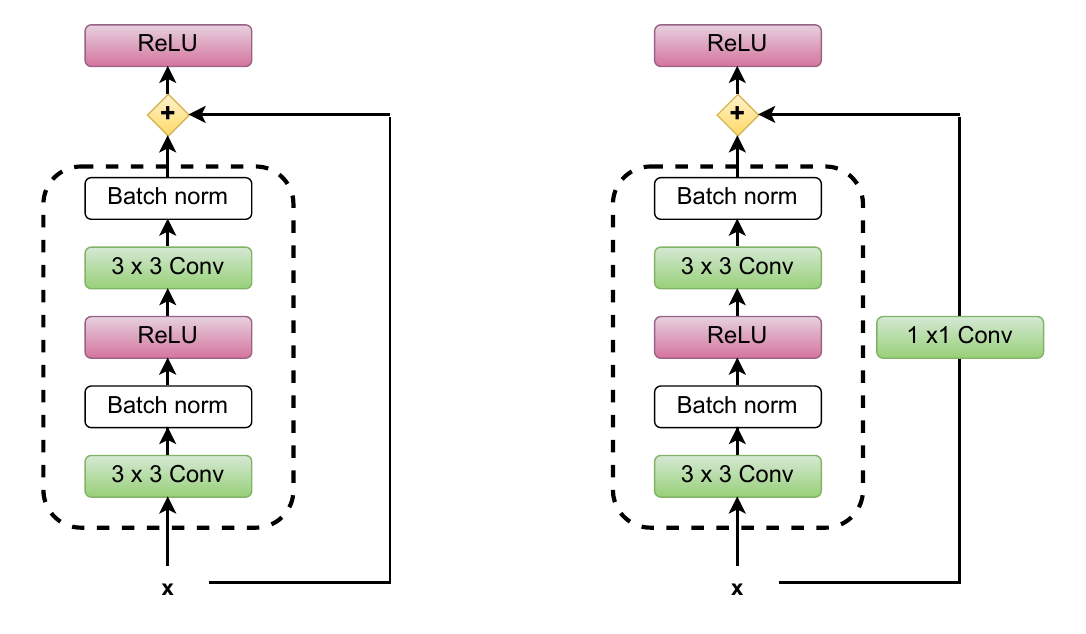}
	\caption{ResNet blocks.}
	\label{fig:process_met}
\end{figure}

Because the images used in our experiments are grayscale, we adapt the first convolutional layer from three input channels to one, by averaging the pretrained RGB weights across the channel dimension. All subsequent layers of the ResNet18 backbone remain unchanged. After the final residual stage, a global average pooling layer produces a 512 dimensional embedding that summarizes the visual information in each image. This embedding is used either on its own (image-only model) or concatenated with tabular predictors (multimodal model), and the resulting feature vector is passed through a small fully connected network that outputs the log-rate predictor for Poisson regression (Poisson Head MLP).

\begin{figure}[H]
\centering
	\includegraphics[width=0.95\textwidth]{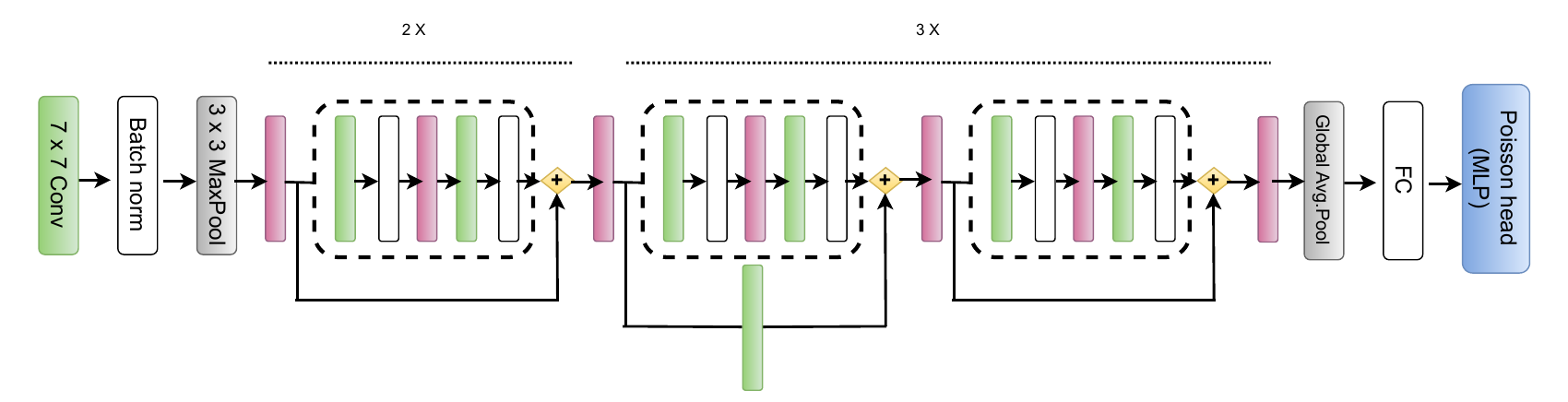}
	\caption{General scheme of the employed ResNet18 model.}
	\label{fig:ResNet18_architecture}
\end{figure}

For the regression model given in  (\ref{eq:mean}) with target variable $y$, the prediction for the observation $i$ is denoted by $\hat{y}_i$. The parameters of $f(\cdot)$ are those that minimize a given loss function $l(\cdot)$ over a training data set $X_{train}$, that is, those minimizing $\sum_{\{i:x_i \in X_{train}\}}l(y_i, \hat{y}_i)$. For the claim frequency task, we assume that the claim count is Poisson distributed, following previous works \citep{holvoet2025neural, clemente2023modelling}. Thus, the loss function is given by the Poisson negative log likelihood
\begin{align*}
     l(y_i, \hat{y}_i) = \dfrac{1}{n}\left(\sum_{i} \hat{y}_i-y_i\ln(\hat{y}_i)-\ln(y_i!)\right),
\end{align*}
where $n$ is the number of observations in $X_{train}$. To evaluate the model performance, we calculate the out-of-sample error of the trained model over the test set $X_{test}$, using the Root Mean Square Error (RMSE), since we are interested in the point estimate of the expected number of claims by zone level.

In addition, to reduce overfitting and to mitigate multicollinearity among  predictors, we also consider Elastic Net regularization within the GLM framework. Elastic Net combines the $L_1$ penalty of lasso with the $L_2$ ridge penalty \citep{zou2005regularization}, allowing both coefficient shrinkage and variable selection. The penalized log-likelihood is defined as
\begin{align}
l_{\text{pen}}(\beta) 
= l(\beta) 
+ \eta \left[
\alpha \|\beta\|_{1} 
+ \frac{1-\alpha}{2}\|\beta\|_{2}^{2}
\right], \label{eq:ELasticNet}
\end{align}
where $\eta \ge 0$ controls the overall amount of regularization and 
$\alpha \in [0,1]$ determines the relative weight of the $L_1$ and $L_2$ components.

Additionally, to evaluate all models in a consistent and statistically reliable way, we follow a multifold out-of-sample procedure created by \citet{henckaerts2021boosting}. Instead of relying on a single train–validation–test split, the dataset $X$ is first partitioned into six stratified and disjointed sets, denoted by $X_1, \cdots, X_6$. Each subset serves once as an external test sample. For a given fold $j$, $j \in \{1, \cdots, 6\}$, $X_j$ is held out, and the remaining observations $X-X_j$ are internally divided into five parts for a separate cross-validation routine. Within this inner loop, four parts are used to fit the model and the remaining one part is used for validation, rotating over the five possible choices.

The average validation error across these five inner folds guides the selection of the hyperparameter configuration  (e.g., regularization strength for GLMs, trees depth, dropout parameters for neural networks). After identifying the parameters' set, we refit the model using all data in $X-X_j$ and compute its predictive performance on the untouched test subset $X_j$ . Repeating this procedure for $j \in \{1, \cdots, 6\}$ produces six independent out-of-sample predictions, ensuring that every observation in the dataset is assessed without ever being used for its own model training.

This evaluation design provides greater stability and significantly lower variance than a single train–test split. Cross-validation, which is done in the inner loop, is known to mitigate sensitivity to arbitrary data partitions and to yield more reliable estimates of predictive accuracy \citep{hastie2009elements, ArlotSylvain2010Asoc}. The following Figure~\ref{fig:ExtendedCV} is an adaptation from the schema given by \citet{henckaerts2021boosting}:
\begin{figure}[H]
\centering
	\includegraphics[width=0.9\textwidth]{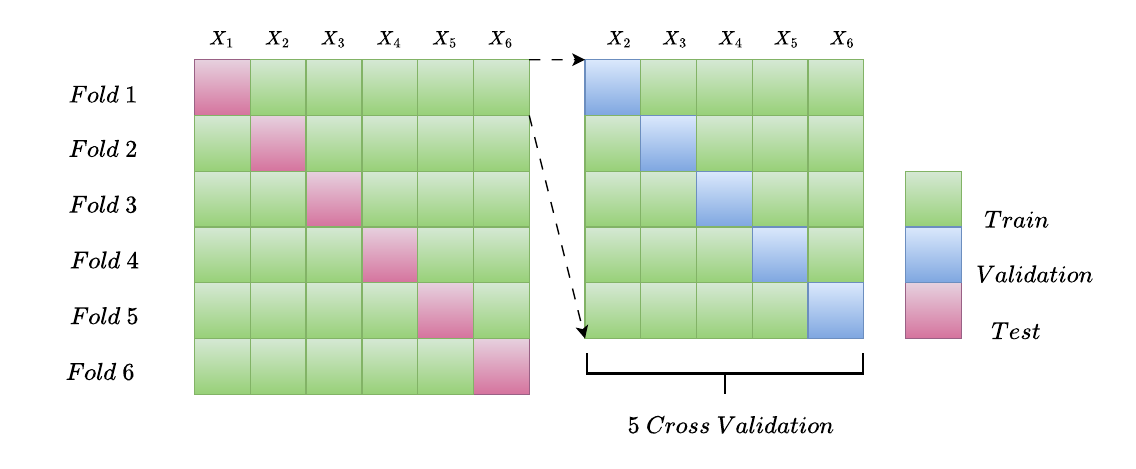}
	\caption{Extended cross validation scheme.}
	\label{fig:ExtendedCV}
\end{figure}
\end{itemize}
In the following subsections, we introduce the tabular data used in the analysis, the construction of the environmental features, the acquisition of the imagery, and the role of image embeddings in the modeling framework.

\subsection{Tabular Data}
\label{subsec:Tabular Data}
The original dataset used in our research is the 1997 Belgian MTPL dataset (beMTPL97), available in the R package CASdatasets \citep{dutang2024insurance}. The data contain information on 163,212 policyholders, each observed over a certain period of exposure. Exposure represents the fraction of the year during which a policyholder was exposed to risk, with a maximum value of one. The variables used in this study are listed in Table~\ref{Table:VariableDescription}.

\begin{table}[htb!]
\centering
\begin{tabular}{lp{0.75\textwidth}}
\toprule \hline
\textbf{Variable} & \textbf{Description}                       \\ \hline \midrule
expo     & Numeric variable to describe the fraction of the year the policy was exposed.                                                                                                                     \\ \hline
coverage & \begin{tabular}[c]{@{}l@{}}Categorical variable for the insurance coverage level: TPL only third party \\ liability, TPL+ limited material damage , TPL++ comprehensive material \\ damage.\end{tabular} \\ \hline
ageph    & Numeric for the policyholder age.                                                                                                                                                                    \\ \hline
sex      & categorical variables for female and male.                                                                                                                                                           \\ \hline
bm       & \begin{tabular}[c]{@{}l@{}}Integer for the level occupied in the former compulsory Belgian bonus-malus \\ scale. From 0 to 22, a higher level indicates worse claim history.\end{tabular}            \\ \hline
power    & Numeric variable for horsepower of the vehicle in Kilowatt.                                                                                                                                          \\ \hline
agec     & Numeric variable for the car in years.                                                                                                                                                               \\ \hline
fuel     & Categorical variable for the type of the vehicle gasoline or diesel.                                                                                                                                 \\ \hline
use      & Categorical variable for the use of the vehicle: private or work.                                                                                                                                    \\ \hline
fleet    & Categorical variable for indicating whether the vehicle is part of a fleet.                                                                                                                          \\ \hline
postcode & Postal code of the municipality of policyholder's residency.                                                                                                                                         \\ \hline
lat      & Numeric variable for the latitude coordinate of the center of the municipality where the policyholder resides.                                                                                       \\ \hline
long     & Numeric variable for the longitude coordinate of the center of the municipality where the policyholder resides.                                                                                      \\ \hline
nclaims  & Numeric variable for the number of claims.                                                                                                                                                           \\ \hline
\end{tabular}
\caption{Variable descriptions.}
\label{Table:VariableDescription}
\end{table}

After exploring the data, we find that the proportion of policyholders with more than three claims is extremely small, that is, less than 0.012\% of the dataset. For this reason, we restrict attention to policyholders with at most three claims. 

We also observed that the 163,193 policyholders correspond to only 583 unique postcodes. With such limited geographic variation at the individual level, it is not feasible to study the impact of built environment features or visual geographic predictors directly at the policyholder level. To obtain a meaningful geographic variation, we aggregated the data at the postcode level. This aggregation is consistent with the granularity at which geographic information is available in beMTPL97, and remains actuarially relevant, as territorial effects are still widely used in MTPL underwriting and helps explain spatial variation in claim frequency patterns.

Aggregating to the area level is also consistent with actuarial practice in situations where confidentiality constraints prevent the employment of individual data. For example, studies using the publicly released long term care (LTC) incidence experience of the Society of Actuaries rely on pre-grouped exposure and claim counts, because the data provider does not provide policy-level observations. Examples include \cite{SOA2015LTC} and \cite{zail2019predictive}, both of which analyze grouped LTC records rather than individual policies.

In our research, to create the aggregated dataset at the postcode level, we used the original variables listed in Table~\ref{Table:VariableDescription}. Exposure and \texttt{nclaims} were aggregated as sums of all individual observations within each postcode. For numerical variables, we compute the mean, median and standard deviation within each postcode. For categorical variables, we constructed proportion-based features that capture the share of each category in that postcode. For clarity, we use the suffixes \texttt{\_mean}, \texttt{\_median}, and \texttt{\_sd} for numerical summaries and \texttt{\_prop} for categorical proportions. For example, the aggregated variables include \texttt{ageph\_mean}, \texttt{ageph\_median}, \texttt{ageph\_sd}, \texttt{coverage\_TPL\_prop}, and \texttt{coverage\_TPL+\_prop}. Additionally, we define the variable \texttt{postcode\_2}, which corresponds to the first two digits of the original \texttt{postcode} variable and represents the region in Belgium. We also include the variables \texttt{latitude} and \texttt{longitude}.

\begin{figure}[htb!]
     \centering
     \begin{subfigure}[b]{0.45\textwidth}
         \includegraphics[width=\textwidth]{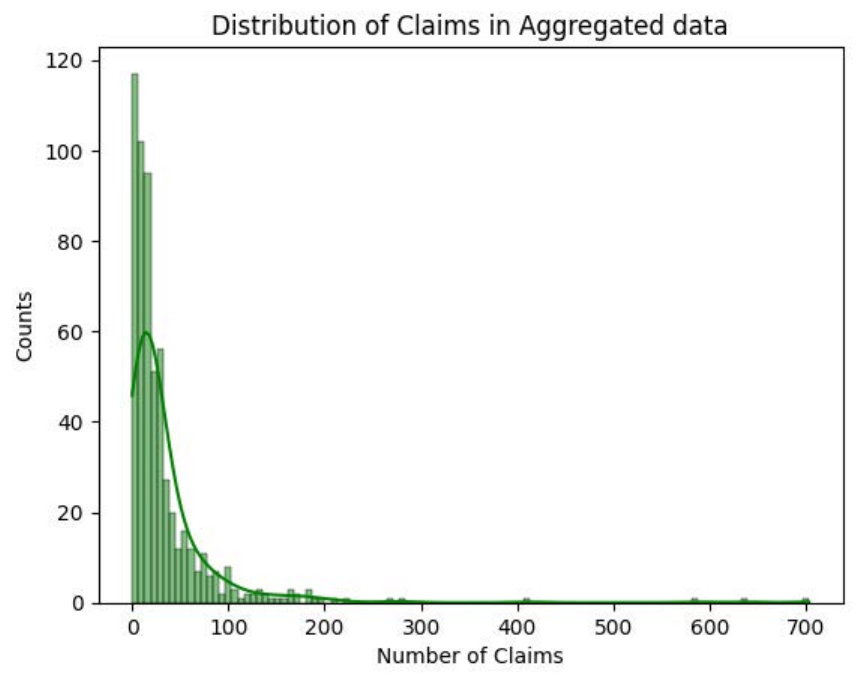}
     \end{subfigure}
     \hfill
     \begin{subfigure}[b]{0.45\textwidth}
         \includegraphics[width=\textwidth]{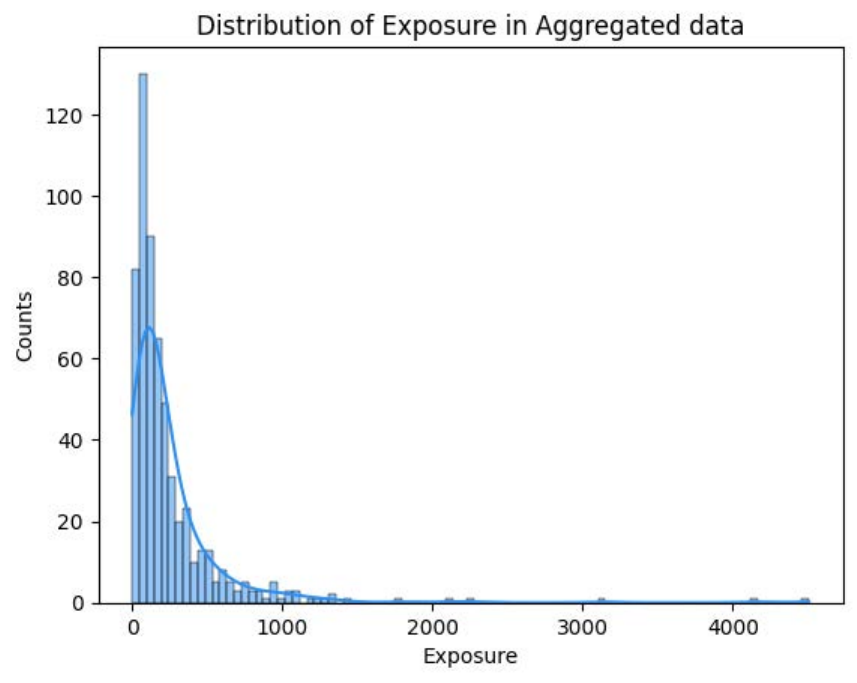}
     \end{subfigure}
     \caption{Histogram of the aggregated number of claims and exposure.}
     \label{fig:Histogram_claims_expo.}
\end{figure}

\begin{figure}[htb!]
    \centering
    \begin{subfigure}[t]{0.45\textwidth}
        \centering
        \vspace{0pt} 
        \includegraphics[width=\textwidth]{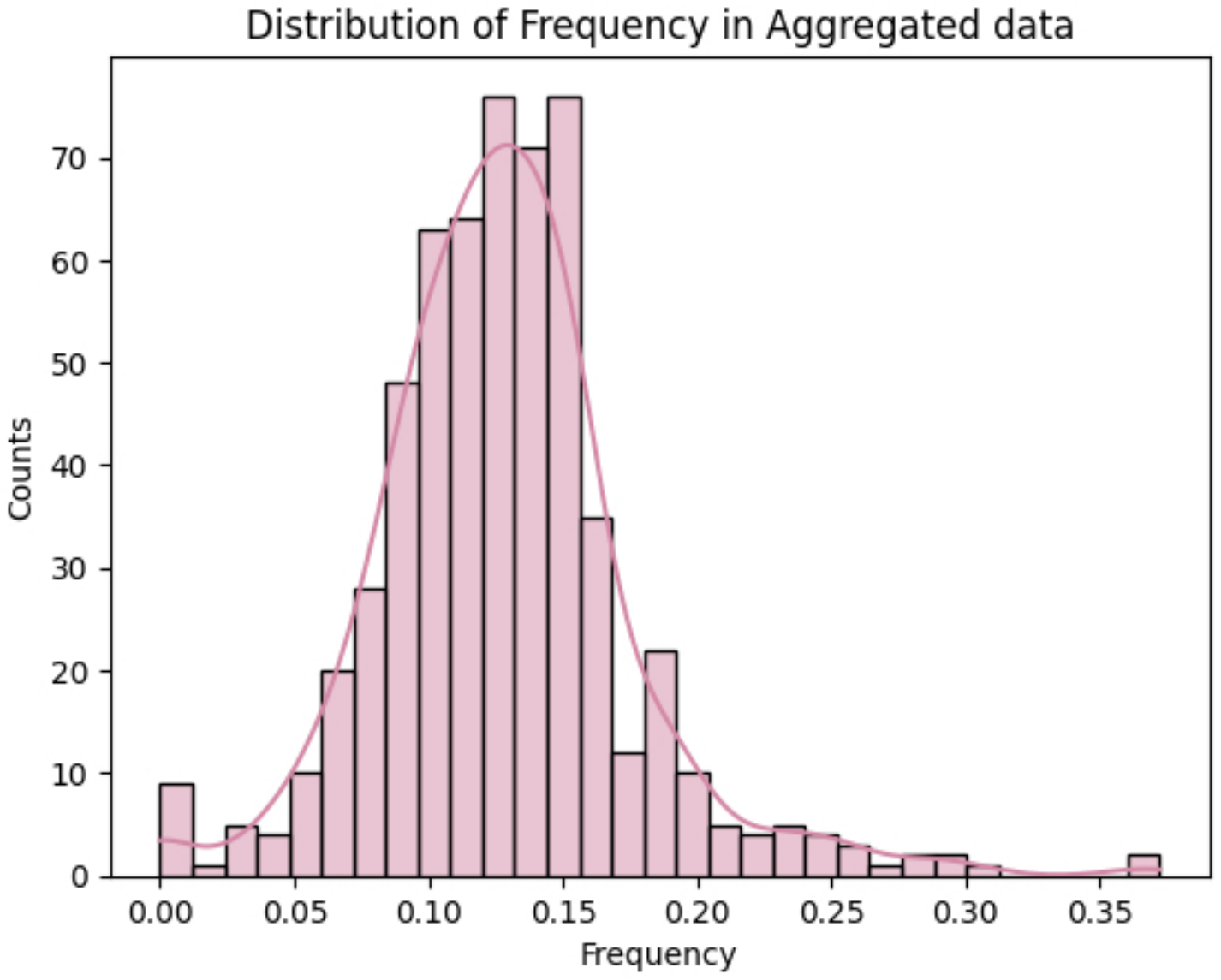}
        \caption{Histogram of the frequency in the aggregated data.}
        \label{subfig:Histogram_freq}
    \end{subfigure}
    \hfill
    \begin{subfigure}[t]{0.5\textwidth}
    \small
        \centering
        \vspace{1.78cm}
        \vspace{0pt} 
        \begin{tabular}{@{}cccccc@{}}
        \toprule \hline
        \textbf{Variable} & \textbf{Mean} & \textbf{Median} & \textbf{SD} & \textbf{Min} & \textbf{Max} \\ \hline \midrule
        \texttt{nclaims$_{\texttt{ag}}$}   & 34.54  & 18.00  & 60.43 & 0.00 & 703.00  \\ \midrule
        \texttt{expo}$_{\texttt{ag}}$      & 249.06 & 141.12 & 379.52 & 1.10 & 4505.73 \\ \midrule
        \texttt{freq} & 0.13   & 0.13   & 0.05  & 0.00 & 0.37    \\ \bottomrule
        \\
        \\
        \\
        \end{tabular}
        \caption{Summary statistics of the variables.}
        \label{Tab:summary_statistics}
    \end{subfigure}
    \caption{Histogram of frequency and summary statistics.}
    \label{fig:Histogram_claims_expo}
\end{figure}

Figures~\ref{fig:Histogram_claims_expo.} and~\ref{subfig:Histogram_freq} and Table~\ref{Tab:summary_statistics} summarize the postcode level distributions of the aggregated number of claims (\texttt{nclaims$_{\texttt{ag}}$}), exposure (\texttt{expo}$_{\texttt{ag}}$), and claim frequency (\texttt{freq} = \texttt{nclaims$_{\texttt{ag}}$/expo$_{\texttt{ag}}$}). The aggregated variables \texttt{nclaims$_{\texttt{ag}}$} and \texttt{expo}$_{\texttt{ag}}$ show a strong right skewness. Most postcodes record relatively low totals, with medians of 18 claims and 141.12 units of exposure. However, a limited number of postcodes correspond to Belgium urban postal zones with high population density, such as 2000 (Antwerp), 4000 (Liège), 6000 (Charleroi) and 9000 (Ghent). These postal zones cover major municipalities and their associated districts, which naturally have much larger insured populations and traffic volumes. As a result, they generate substantially higher aggregated values, with a maximum of 703 claims and more than 4,500 units of exposure. Therefore, these high totals reflect structural differences between postcodes rather than anomalies, and all aggregated values correspond to valid regions. The histogram of the claim frequency shows a roughly unimodal distribution centered between 0.10 and 0.15. The left tail reflects postcodes with very low frequencies, while the right tail includes a small number of postcodes with higher values, up to approximately 0.36. These higher frequencies are genuine postcode-level results and were retained.


\subsection{Construction of the Environmental Features}
\label{subsec:EnvironmentalF}
The built environment around a residence is often considered relevant for assessing road traffic risk. Factors such as road density, intersection layout, traffic control devices, and the concentration of commercial or public amenities can influence local traffic volume and traffic complexity and may therefore be associated with accident occurrence. Similarly, the presence of schools, healthcare facilities, or fuel stations is linked to distinct activity patterns and potentially different exposure to risk conditions. These considerations suggest that neighborhood characteristics could help explain the variation in claim frequency at the postcode level, for example, \cite{STEVENSON2022378} showed that the greenery and other characteristics of the neighborhoods influence the value of the assets of the people who inhabit them, helping in the determination of losses. We aim to empirically evaluate whether these features contribute predictive information beyond the tabular variables.

To characterize the built environment around each postcode, we use OpenStreetMap (OSM) feature data \citep{openstreetmap, osmBelgium2025} together with CORINE Land Cover 2000 data \citep{CLC2000}. These two datasets provide complementary information, OSM provides detailed representations of roads, amenities, and other local infrastructure, while CORINE offers a consistent Europe-wide land cover classification around the year 2000.

OSM is a collaboratively curated geospatial database, it is also known as a volunteered geographic information system, which started in 2004. In our setting, OSM features are used as proxy indicators of the spatial context surrounding each postcode, particularly for long lived infrastructure such as major roads and public facilities, which tend to be stable over time. In addition, CORINE 2000 provides a 100m resolution map classified into 44 land cover categories. We retain only cells labeled as artificial surfaces (codes 111–142), covering urban fabric, industrial and commercial units, and transport infrastructure. These represent areas that were already developed by 2000.

The BeMTPL97 insurance dataset refers to year 1997, whereas detailed OSM feature coverage is only available from later snapshots, we use the 2014 extract, which is the version available to the public that is closer in time. To reduce the temporal mismatch, we partially \textit{backdate} the OSM layers using CORINE 2000. Specifically, we retain only the OSM features that fall within the artificial surface cells of CORINE, thus restricting the analysis to locations that were already urbanized by 2000. Although the CORINE mask does not identify individual buildings or roads present in 1997, it restricts the retained OSM features to locations that were urbanized prior to 2000, then reducing the influence of features located in newly developed areas that emerged after the accident period. We note that the temporality of the data is not the relevant part of our analysis. We do not intend to model if the specific area within Belgium at that time was predictable; our research studies if modern AI techniques allow analysts to incorporate urbanization data in general.

For each postcode, we take the latitude and longitude of the location and project it to the Belgian Lambert 72 coordinate system, so that distances and areas were measured in meters. Around each postcode center, we draw circular neighborhoods (buffers) with radius of 0.5 km, 1 km, 3 km, and 5 km. These radius are chosen to capture both the immediate surroundings and the progressively wider neighborhoods of the center location of the insured municipality, consistent with evidence that many road accidents occur close to the place of residence \citep{thiran1997accidents}.

From the masked OSM layers, we derive measures that describe the local road network, traffic control devices, amenity concentrations, and the presence of key public facilities. All variables are calculated separately for each radius, with suffixes \texttt{\_r0.5}, \texttt{\_r1}, \texttt{\_r3} , and \texttt{\_r5} denoting the corresponding buffer sizes in kilometers. The entire extraction pipeline produces a set of candidate metrics from roads, buildings, and amenities. For the modeling stage, we focus on the subset listed in Table~\ref{tab:env_vars}, which provides the most interpretable and informative summary of the local traffic environment around each postcode.

\begin{table}[t]
\centering
\begin{tabular}{ll}
\toprule \hline
\textbf{Category} & \textbf{Variable description} \\
\hline \midrule
\textbf{Road network} &
\begin{tabular}[t]{@{}l@{}}
Road length per km$^2$ (\texttt{road\_len\_km\_per\_km2\_r*}) \\
Intersection count per km$^2$ (\texttt{intersection\_count\_per\_km2\_r*}) \\
Roundabout count per km$^2$ (\texttt{roundabout\_count\_per\_km2\_r*}) \\
Traffic-signal count per km$^2$ (\texttt{traffic\_signal\_count\_per\_km2\_r*})
\end{tabular}
\\[2mm]
\hline
\textbf{Amenities} &
\begin{tabular}[t]{@{}l@{}}
Retail count per km$^2$ (\texttt{retail\_count\_per\_km2\_r*}) \\
Tourism count per km$^2$ (\texttt{tourism\_count\_per\_km2\_r*}) \\
Parking facilities count per km$^2$ (\texttt{parking\_count\_per\_km2\_r*})
\end{tabular}
\\[2mm]
\hline
\textbf{Public facilities} &
\begin{tabular}[t]{@{}l@{}}
Presence of schools (\texttt{has\_education\_r*}) \\
Presence of healthcare facilities (\texttt{has\_healthcare\_r*}) \\
Presence of fuel stations (\texttt{has\_fuel\_station\_r*}) \\
School count per km$^2$ (\texttt{school\_count\_per\_km2\_r*}) \\
Healthcare count per km$^2$ (\texttt{healthcare\_count\_per\_km2\_r*}) \\
Fuel-station count per km$^2$ (\texttt{fuel\_count\_per\_km2\_r*})
\end{tabular}
\\

\bottomrule
\end{tabular}
\caption{Environmental variables extracted from masked OSM layers. 
Each variable is computed for four radii (0.5 km, 1 km, 3 km, 5 km), 
with suffixes \texttt{\_r0.5}, \texttt{\_r1}, \texttt{\_r3} and 
\texttt{\_r5} indicating the neighborhood size.}
\label{tab:env_vars}
\end{table}

\begin{figure}[H]
    \centering
    \begin{subfigure}[t]{0.32\textwidth}
        \centering
        \vspace{0pt}
        \includegraphics[width=\textwidth]{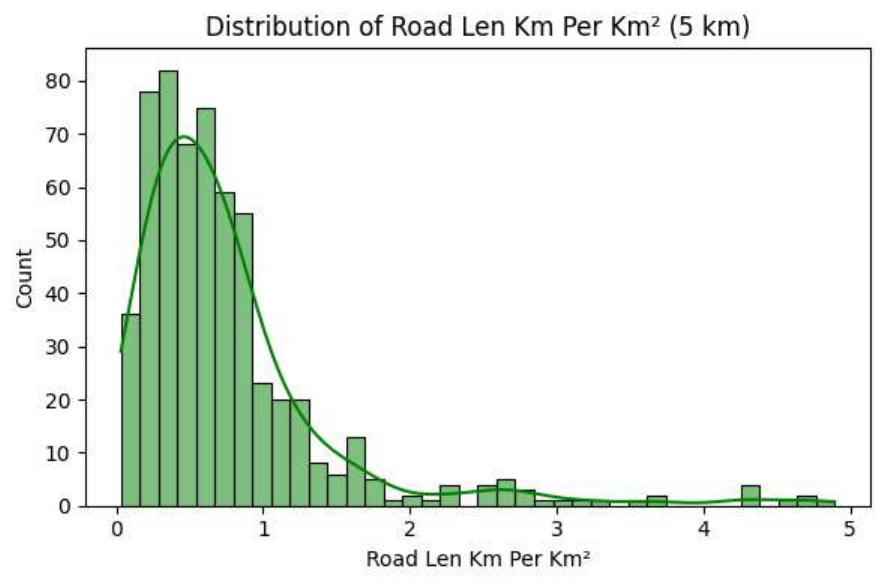}
        \caption{Histogram of the road length}
        \label{subfig:Histogram_roadlen}
    \end{subfigure}
    \begin{subfigure}[t]{0.32\textwidth}
        \centering
        \vspace{0pt} 
        \includegraphics[width=\textwidth]{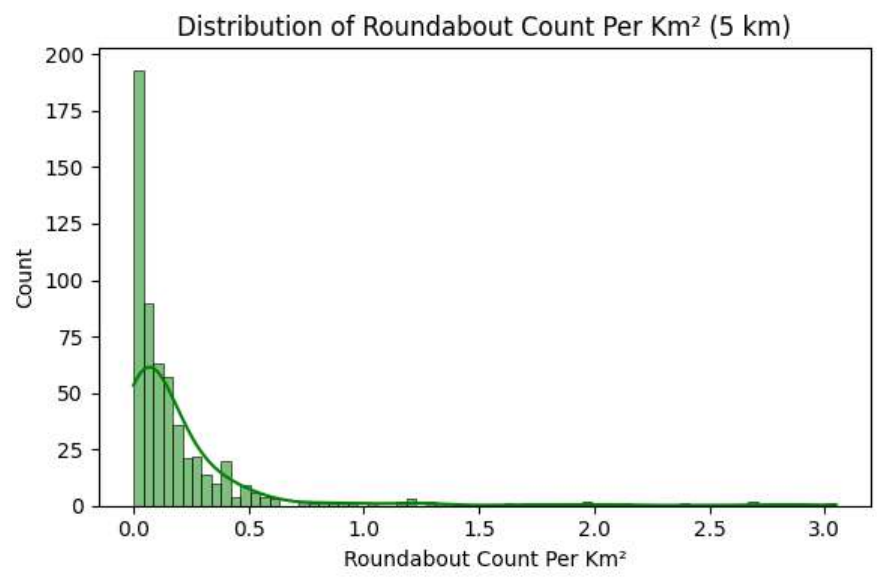}
        \caption{Histogram of the roundabout count}
        \label{subfig:Histogram_round}
    \end{subfigure}
    \begin{subfigure}[t]{0.32\textwidth}
        \centering
        \vspace{0pt} 
        \includegraphics[width=\textwidth]{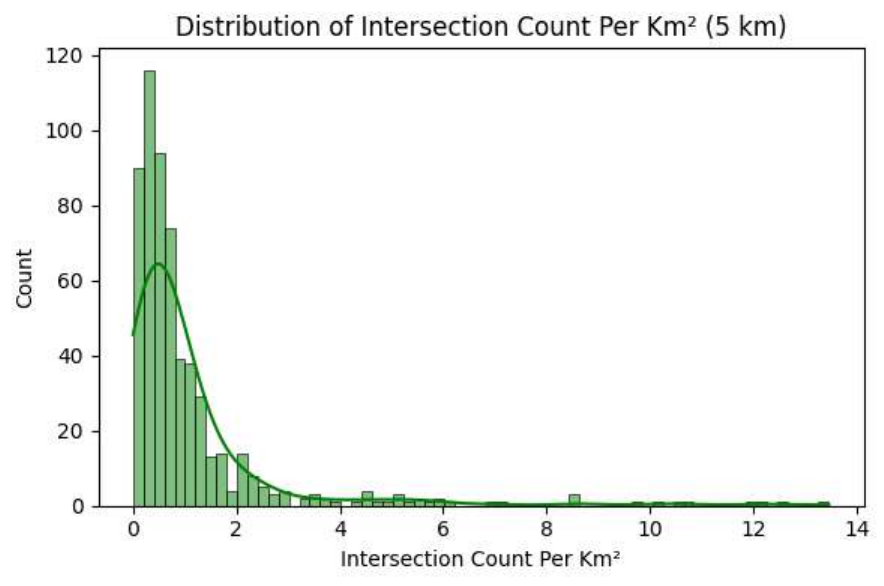}
        \caption{Histogram of the Intersection Count}
        \label{subfig:Histogram_Intersection}
    \end{subfigure}
    \begin{subfigure}[t]{0.32\textwidth}
        \centering
        \vspace{0pt} 
        \includegraphics[width=\textwidth]{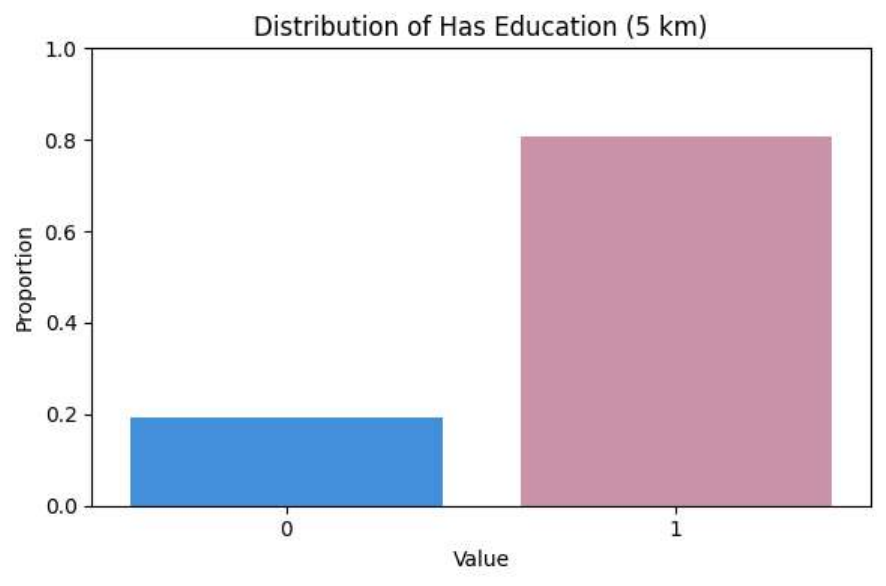}
        \caption{Plot bar of eduction indicator variable}
        \label{subfig:Plot_Education}
    \end{subfigure}
     \begin{subfigure}[t]{0.32\textwidth}
        \centering
        \vspace{0pt} 
        \includegraphics[width=\textwidth]{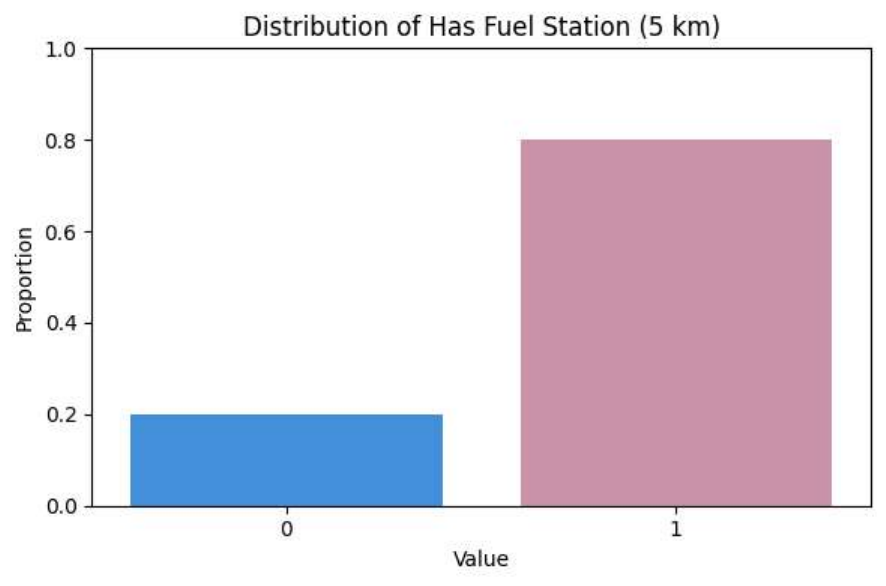}
        \caption{Plot bar for fuel indicator variable}
        \label{subfig:Plot_Fuel}
    \end{subfigure}
     \begin{subfigure}[t]{0.32\textwidth}
        \centering
        \vspace{0pt}
        \includegraphics[width=\textwidth]{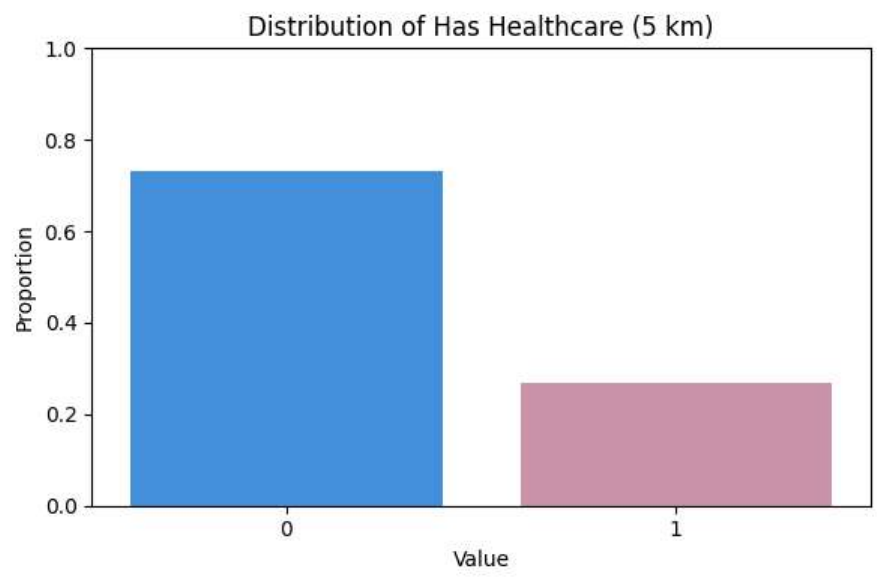}
        \caption{Plot bar for healthcare indicator variable.}
        \label{subfig:Plot_Education}
    \end{subfigure}
    \caption{Histogram and bar plots of some environmental features at radius 5 km}
    \label{fig:Histogram_r500}
\end{figure}

For illustration, we show histograms and bar plots of selected variables for the buffer radius of 5 km in Figure~\ref{fig:Histogram_r500}. These distributions reveal several distinct patterns across environmental indicators. Road density, roundabout counts, and intersection counts all display right skewed continuous distributions. The mean values within the 5 km buffer are approximately 60.1 km of road length, 15.9 roundabouts, and 85.8 intersections. The amenity indicators show different behaviors. The presence of educational facilities and fuel stations is relatively common in postcodes, whereas healthcare facilities are absent in a large proportion of areas. In general, these patterns illustrate how different types of infrastructure vary between postcodes and how they provide complementary information to the local built environment.

\subsection{Imagery}
\label{subsec:Imagery}

Obtaining imagery suitable for assessing the contribution of visual geographic information to claim frequency prediction is challenging because the BeMTPL97 dataset refers to the year 1997, while historical, high-resolution imagery is not uniformly available for that period. Public insurance datasets that include georeferenced policyholder information are rare, and the datasets commonly used in actuarial research (see Section~\ref{sec:LiteratureR}) do not provide geographic coordinates. The BeMTPL97 data is therefore one of the few datasets that enables such an analysis, though geographic information is only available at aggregated level, with multiple policyholders sharing the same postcode rather than, for example, having individual residential locations.

We initially explored several potential imagery sources, including OpenStreetMap basemaps and Google Earth historical satellite views. OpenStreetMap cartographic tiles reflect the present day appearance of the landscape and therefore do not align with the temporal reference of our data. Google Earth provides historical imagery, but the available dates vary substantially between locations, and the most usable images starting only around 2015–2016. In addition, automated or bulk downloading of Google Earth imagery is prohibited under Google's terms of service, which prevents its use for large scale acquisition.

Following an extensive search, we identified a suitable historical dataset: the national orthoimagery for Belgium produced by the National Geographic Institute (NGI Belgium). The 1995 NGI orthophotos constitute the closest available representation of the built environment around the 1997 claim period. We accessed the imagery through the tiled web service provided by ©Institut géographique national (NGI Belgium) via Cartesius.be \citep{CartesiusPortal} for academic research purposes. In line with NGI’s recommendation, the imagery was retrieved dynamically through their Tiled Service Layer rather than by downloading full raster datasets. This method allows for the extraction of only the geographic areas required for each postcode buffer, integrates with standard GIS software (in our case, ArcGIS Pro) and ensures efficient and compliant use of the data. Therefore, all imagery used in this study was obtained on demand from the NGI tile service in accordance with these guidelines.

For each postcode, we extracted square orthoimage tiles centered on its coordinates, with apothem lengths of 0.5 km, 1 km, and 3 km. We initially considered a 5 km tile as well. However, approximately 12\% of these large tiles extended beyond the coverage of the 1995 NGI orthophotos, resulting in incomplete images. We therefore restrict the analysis to the fully observed tile sizes (0.5 km, 1 km and 3 km), for which imagery is complete for all postcodes. Note that larger tiles are associated with lower resolutions. As a result, using different spatial resolutions allows us to examine whether more detailed geographic imagery (with smaller coverage) improves predictive performance, or whether broader but less detailed postcode coverage yields better results, and how sensitive the results are to the choice of spatial scale. The imagery extraction was conducted in ArcGIS Pro using its interface, with Python scripts automating the retrieval of the square tiles for each postcode.

\begin{figure}[htb!]
    \centering
    \begin{subfigure}[t]{0.32\textwidth}
        \centering
        \vspace{0pt} 
        \includegraphics[width=\textwidth]{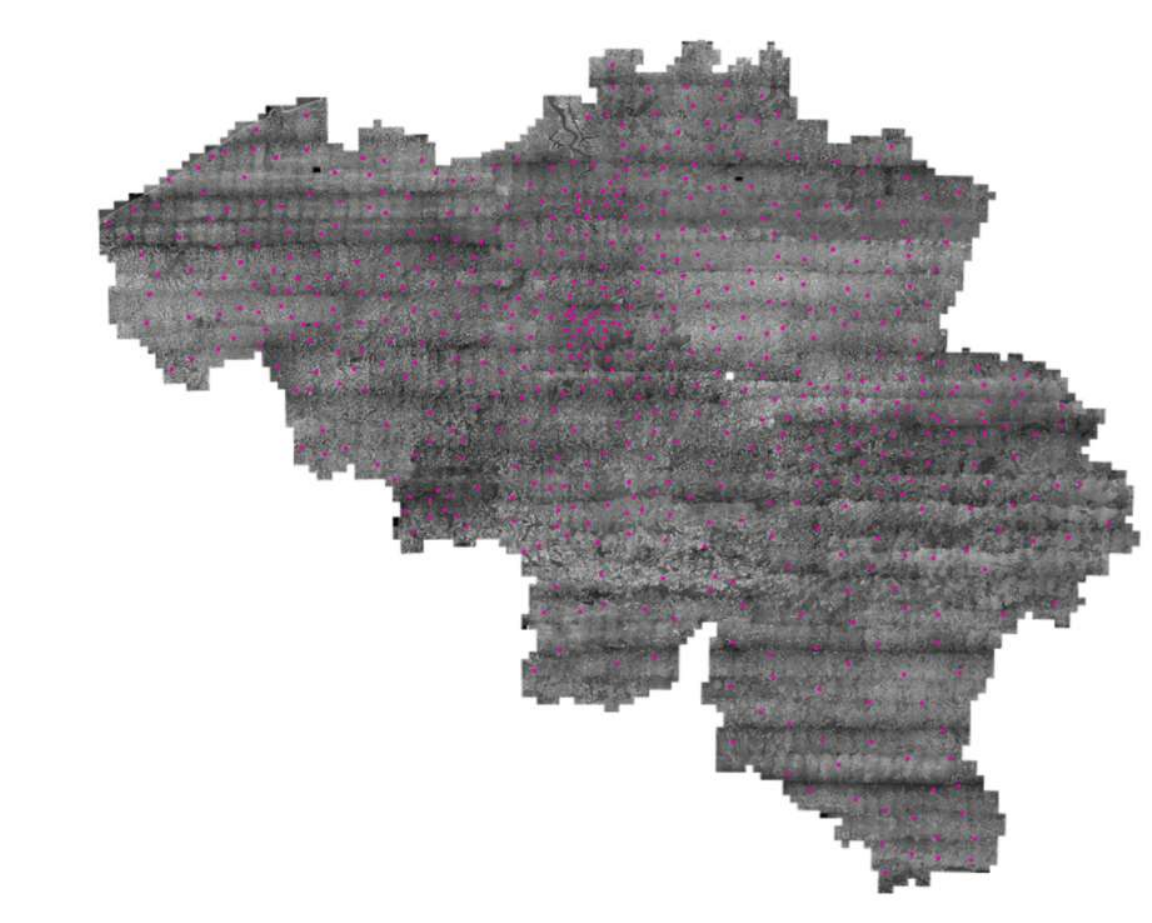}
    \end{subfigure}
    \begin{subfigure}[t]{0.32\textwidth}
        \centering
        \vspace{0pt} 
        \includegraphics[width=\textwidth]{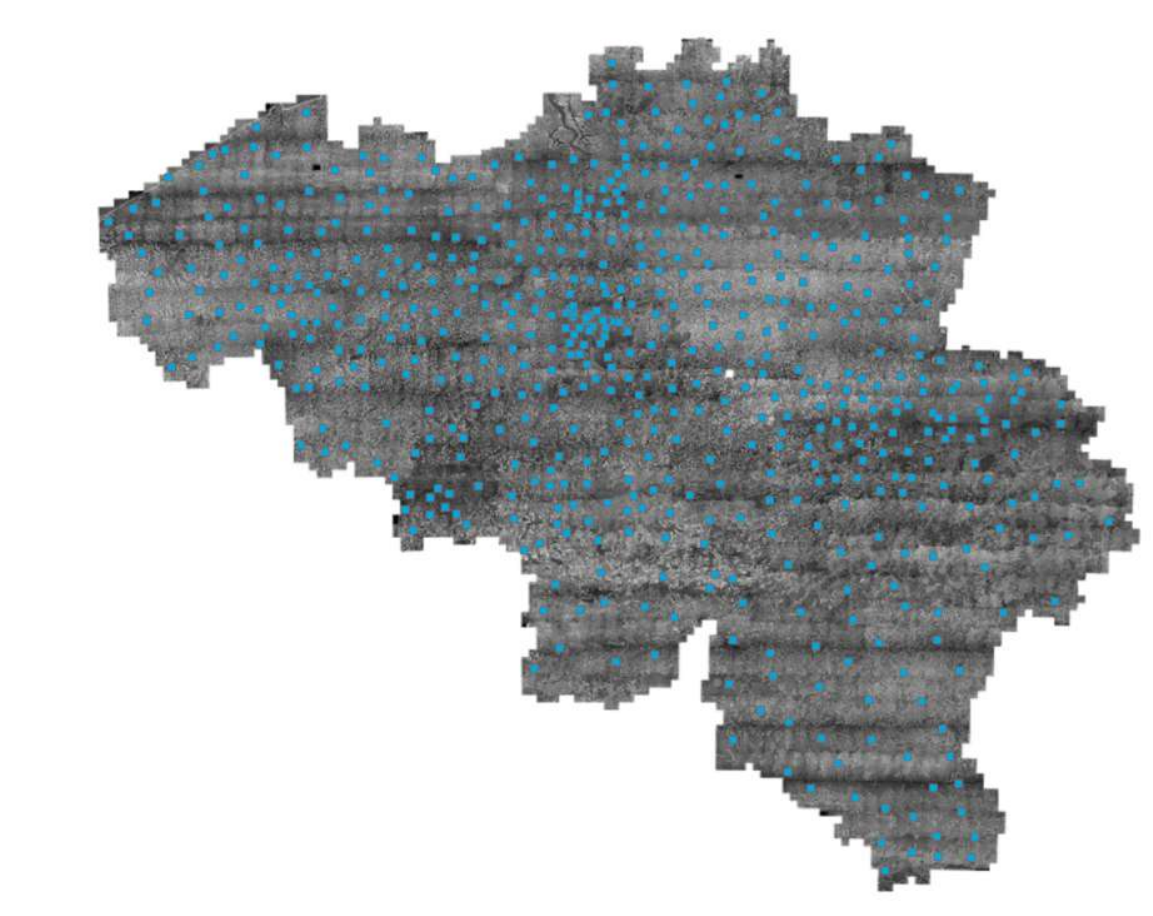}
    \end{subfigure}
    \begin{subfigure}[t]{0.32\textwidth}
        \centering
        \vspace{0pt} 
        \includegraphics[width=\textwidth]{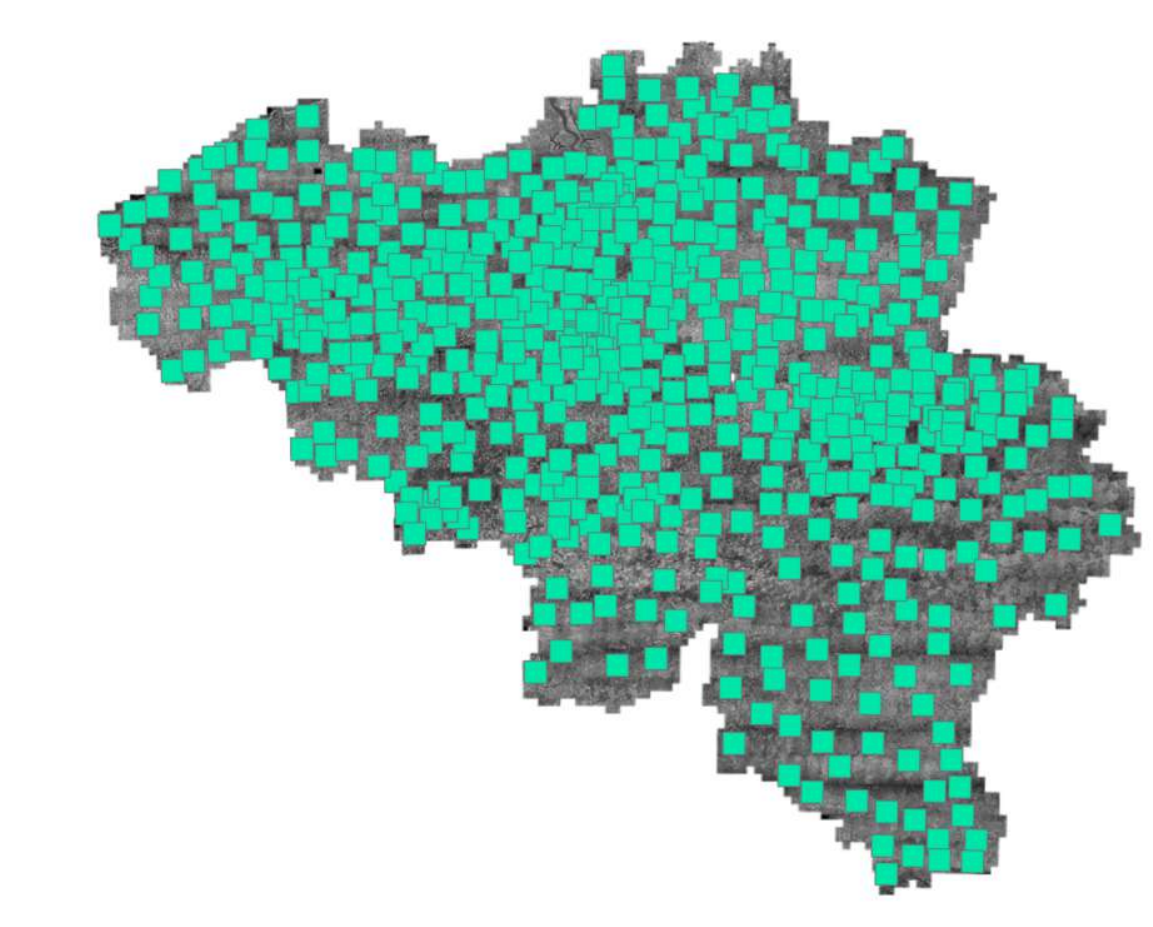}
    \end{subfigure}
    \caption{Orthoimages tiles centered on the given postcodes' coordinates, with apothem lengths of 0.5 km, 1 km and 3 km.}
    \label{fig:Ortho95_squemas}
\end{figure}

\begin{figure}[htb!]
    \centering
    \begin{subfigure}[t]{0.32\textwidth}
        \centering
        \vspace{0pt} 
        \includegraphics[width=\textwidth]{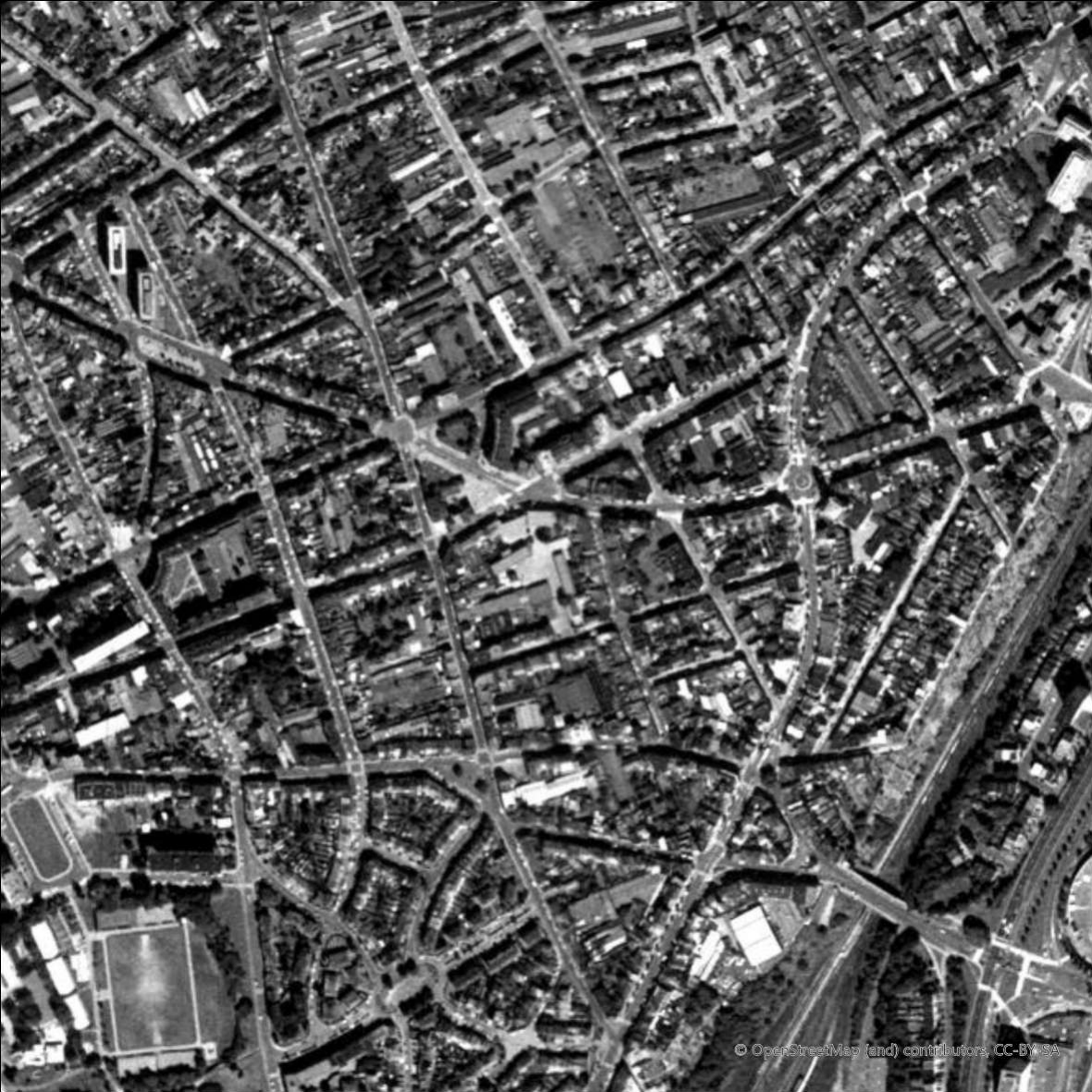}
    \end{subfigure}
    \begin{subfigure}[t]{0.32\textwidth}
        \centering
        \vspace{0pt} 
        \includegraphics[width=\textwidth]{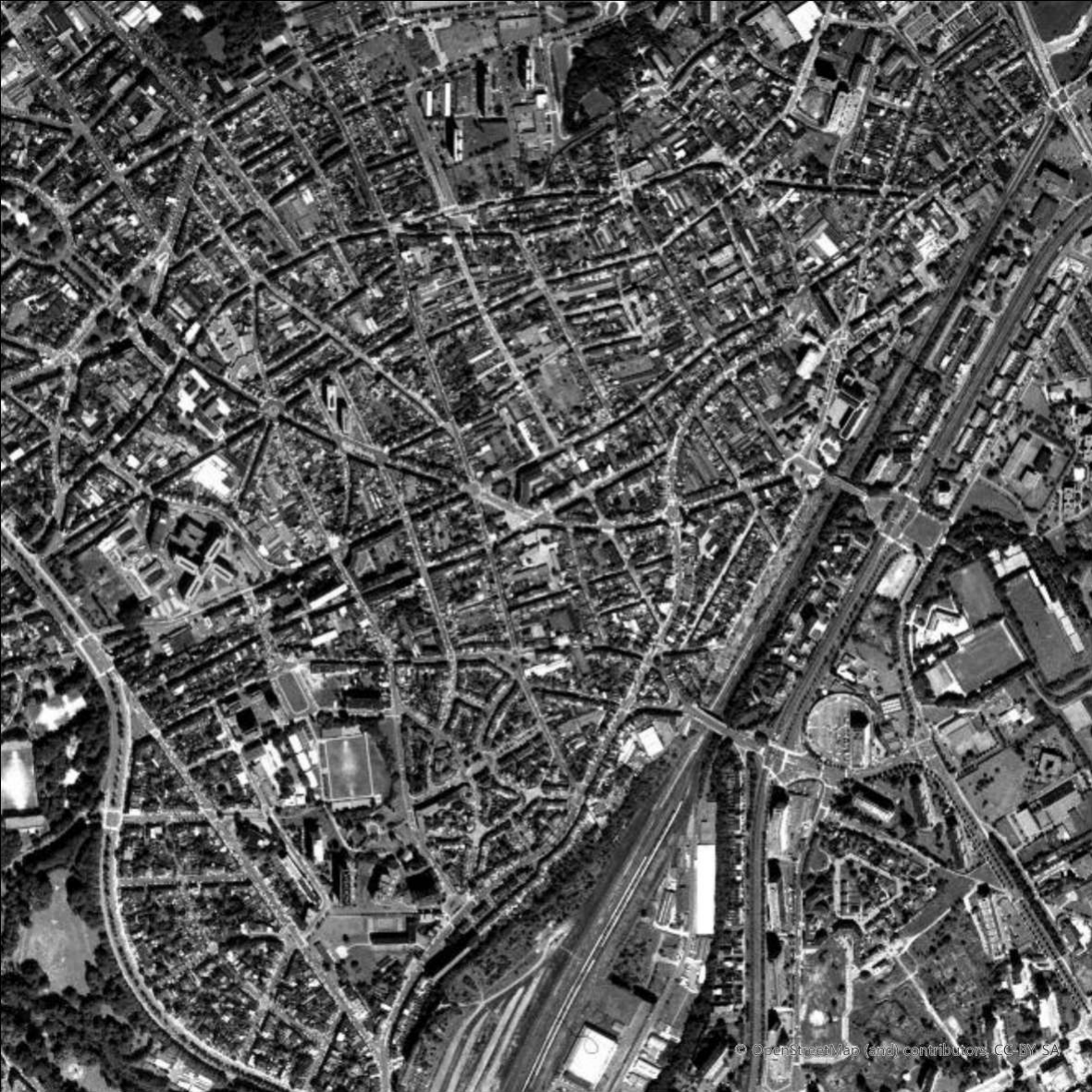}
    \end{subfigure}
    \begin{subfigure}[t]{0.32\textwidth}
        \centering
        \vspace{0pt} 
        \includegraphics[width=\textwidth]{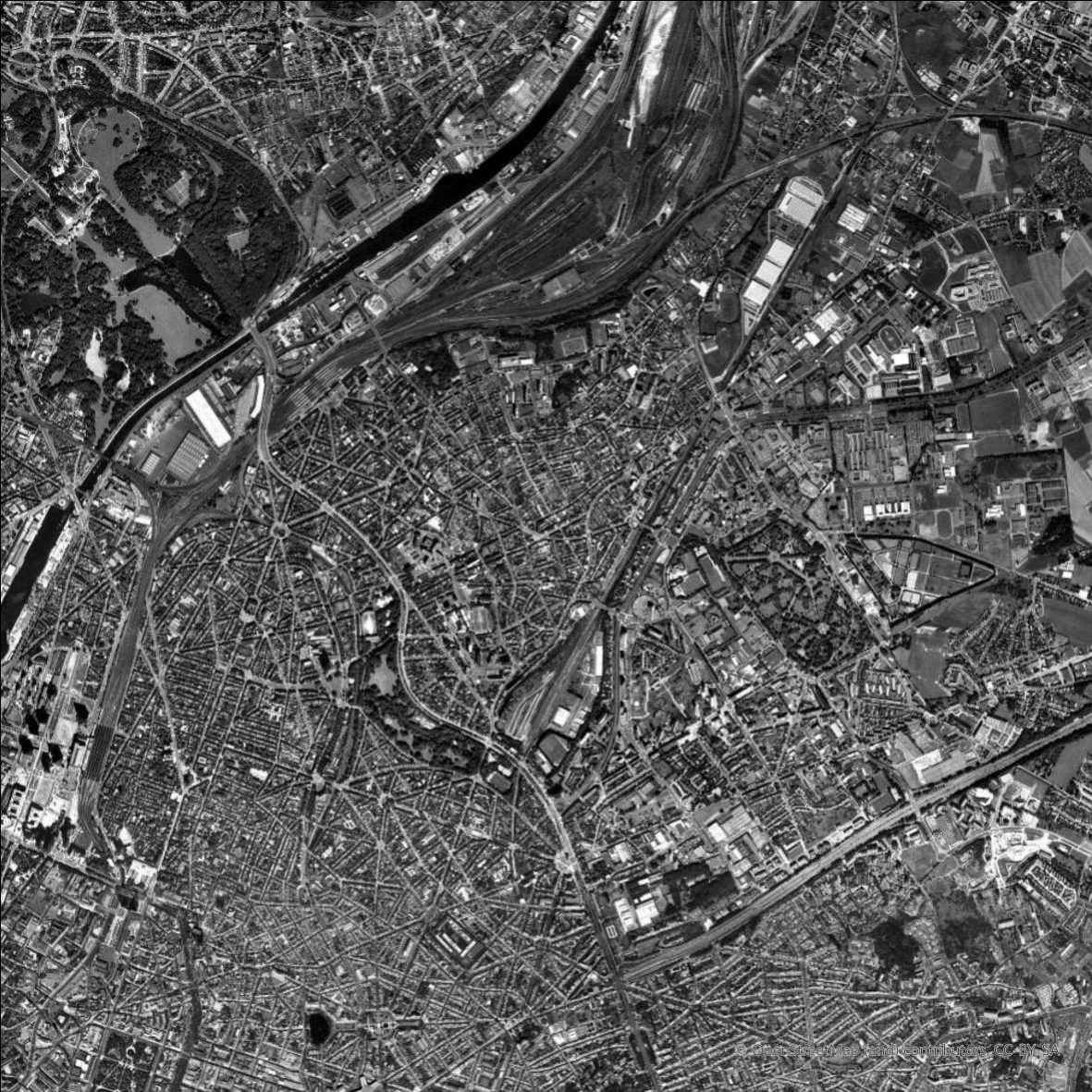}
    \end{subfigure}
    \caption{Detailed example for the squares generated for the postcode 1140 with latitude and longitude, 50.87064 N and 4.39674 E, respectively.}
    \label{fig:Ortho95_example}
\end{figure}

Figure~\ref{fig:Ortho95_squemas} illustrates the square orthoimage tiles used in the study, with apothem lengths of 0.5 km, 1 km, and 3 km. For illustration, 
Figure~\ref{fig:Ortho95_example} shows an example of these tiles for a selected postcode in the dataset.


Figure~\ref{fig:Ortho95_example} shows an example of these tiles for a selected postcode in the dataset. The choice of apothem determines the spatial scale represented in each image and therefore affects the level of visual detail captured. Smaller apothems correspond to a more local, zoomed-in view of the built environment, where fine grained features such as road geometry and nearby structures are more clearly visible. Larger apothems provide a broader spatial context, but necessarily reduce the prominence of local details, as a fixed image size represents a larger geographic area. Importantly, the technical image resolution is held constant across apothems; the variation lies in the spatial scale rather than in image quality.

\subsection{Visual Features}
\label{subsec:VisualFeatures}

In addition to the image based ResNet18 model, we also extract visual embeddings from each orthoimage tile and use them as predictors in tabular models (GLM, regularized GLM, and XGB). This enables us to assess the standalone predictive value of visual geographic information, independently of training a CNN for the Poisson regression step.

We construct visual features in two ways. First, we use the trained ResNet18 backbone, which was described in Section~\ref{sec:Methodology}. From the model achieving the best mean score of the six 5-fold cross-validation RMSE and using only the images, we retain only the convolutional backbone and discard the Poisson regression head. Each image tile is then passed through the backbone to obtain a 512 dimensional embedding that summarizes its visual structure (e.g., road layout, building and street density, land-cover patterns). These embeddings serve as additional covariates in the tabular models.

Second, to complement CNN-based features, we also generate 768-dimensional embeddings using the \texttt{nomic\allowbreak-embed\allowbreak-vision\allowbreak-v1.5} model, a pretrained Vision Transformer optimized for semantic similarity \citep{nussbaum2024nomic}. Each image tile is resized and processed through the transformer, and the resulting embedding provides a compact representation of the local visual environment. These embeddings are evaluated as standalone geographic predictors in the GLM and XGB models.

\subsection{Postal Code Representations}
\label{sec:PCRep}

As described in Subsections~\ref{subsec:EnvironmentalF} and~\ref{subsec:Imagery}, the incorporation of alternative geographic information (whether derived from OSM features, raw images, or image embeddings) is based on the location of the municipality centroid associated with each postcode. Circular buffers or square tiles are constructed around this centroid and used as spatial representations of the postcode. A natural question is therefore how well these constructed neighborhoods capture the underlying postcode zones.

To assess this, we rely on a postcode boundary map provided by Esri \citep{esri_belgium_postcodes}, which reports the surface area (in km$^2$) of Belgian postcodes as of 2022. Although this information postdates the BeMTPL97 dataset, postcode boundaries in Belgium have remained largely stable over time, making this a reasonable proxy for the postcode area. We compare the area of each postcode with the area of the constructed circular buffers and square tiles in order to quantify the maximum proportion of the postcode that can be covered for a given radius or tile size.

Based on this analysis, for circular neighborhoods with radius of 0.5~km, 1~km, 3~km, and 5~km, the average maximum proportion of the area of the postcode covered is 4.96\%, 15.38\%, 76.81\% and 97.58\%, respectively. For square image tiles with side lengths of 0.5~km, 1~km, and 3~km, the corresponding average coverage proportions are 5.17\%, 9.26\% and 84.42\%. These results indicate that smaller neighborhoods capture only a limited fraction of most postcode areas, while larger buffers and tiles can approximate nearly complete coverage.

If we focus on postcodes with high exposure and a large number of claims, such as 2000 (Antwerp), 4000 (Li\`ege), 6000 (Charleroi) and 9000 (Ghent), the maximum proportion of the postcode area covered by circular buffers increases substantially with the radius. Specifically, the coverage ranges from 15.71\% to 100\% for postcode 2000, from 9.52\% to 100\% for postcode 4000, from 19.64\% to 100\% for postcode 6000, and from 8.06\% to 100\% for postcode 9000 when moving from a 0.5~km to a radius of 5~km. 

This indicates that postcodes associated with high exposure and claim counts are not necessarily poorly represented by the constructed neighborhoods. Rather, these postcodes correspond to highly populated urban areas with a relatively small to medium geographical extent, so that larger buffers can achieve near complete coverage.

Rather than restricting the analysis to neighborhood sizes that achieve near-complete postcode coverage, we deliberately evaluate model performance across multiple spatial scales. This allows us to assess whether partial geographic information (captured through smaller buffers or tiles) can still improve predictive performance relative to baseline models and to examine the sensitivity of results to the chosen spatial representation.

\section{Results}
\label{sec:Results}
In this section, we present the predictive results obtained using the publicly available BeMTPL97 dataset, with the objective of forecasting the number of claims at the postcode level, using this empirical setting as a concrete basis for examining how the proposed approach performs and what it reveals in practice. Because our focus lies in evaluating the contribution of geography-driven signals, we first examine geography-related variables available or deduced directly from the tabular data, namely the geographic coordinates (\texttt{latitude} and \texttt{longitude}) and the regional categorization encoded by the first two digits of the postcode (\texttt{postcode\_2}). This initial analysis assesses whether the spatial structure already present in the data provides predictive value.

We then extend the feature set by incorporating the environmental indicators described in Subsection~\ref{subsec:EnvironmentalF}, which summarize local characteristics of the built environment derived from OSM and CORINE. For each model specification, performance is assessed on the six test folds, and we report the six corresponding test RMSE values together with their mean and standard deviation. Hyperparameters were selected using 5-fold cross-validation performed on the training portion of each fold (i.e., the complement of the held-out test set). Results are organized by model family, GLMs (with and without regularization), and XGB, for clarity of presentation. In all tables, \texttt{+ lat\_long} denotes the inclusion of latitude and longitude, while \texttt{+ osm\_r} refers to the addition of environmental features computed at radius \texttt{r} and used in conjunction with non-geographic tabular variables.

\begin{table}[htb!]
\begin{adjustbox}{max width=\textwidth}
\begin{tabular}{@{}lcccccccc@{}}
\toprule \hline
\multicolumn{1}{c}{\textbf{}}                                                           & \multicolumn{6}{c}{\textbf{Test RMSE}}                                                                                                                                                                                                                                                                                                                                                                                                                                                                                      & \multicolumn{1}{c}{\textbf{}}     & \multicolumn{1}{c}{\textbf{}}    \\ \hline \midrule
\multicolumn{1}{c}{\textbf{\begin{tabular}[c]{@{}c@{}}GLM\\ Model Family\end{tabular}}} & \multicolumn{1}{c}{\textbf{\begin{tabular}[c]{@{}c@{}}Data fold\\ 0\end{tabular}}} & \multicolumn{1}{c}{\textbf{\begin{tabular}[c]{@{}c@{}}Data fold\\ 1\end{tabular}}} & \multicolumn{1}{c}{\textbf{\begin{tabular}[c]{@{}c@{}}Data fold\\ 2\end{tabular}}} & \multicolumn{1}{c}{\textbf{\begin{tabular}[c]{@{}c@{}}Data fold\\ 3\end{tabular}}} & \multicolumn{1}{c}{\textbf{\begin{tabular}[c]{@{}c@{}}Data fold\\ 4\end{tabular}}} & \multicolumn{1}{c}{\textbf{\begin{tabular}[c]{@{}c@{}}Data fold\\ 5\end{tabular}}} & \multicolumn{1}{c}{\textbf{Mean}} & \multicolumn{1}{c}{\textbf{std}} \\
\hline
\\
Only tabular feature (not geography-related)                                                                                        & 6.8547                                                                             & 16.0352                                                                            & 6.8899                                                                             & 7.6411                                                                             & 7.8392                                                                             & 7.7015                                                                             & 8.8270                            & 3.5567                           \\
\texttt{+ lat\_long}                                                                             & 7.1231                                                                             & 14.0282                                                                            & 7.1434                                                                             & 8.8950                                                                             & 7.4595                                                                             & 7.6076                                                                             & 8.7095                            & 2.6859                           \\
\texttt{+ postcode\_2}                                                                           & 10.4139                                                                            & 24.4254                                                                            & 9.0655                                                                             & 8.8513                                                                             & 7.7978                                                                             & 8.3658                                                                             & 11.4866                           & 6.3987                           \\
\texttt{+ lat\_long + postcode\_2}                                                              & 10.3685                                                                            & 23.9839                                                                            & 8.9836                                                                             & 8.0294                                                                             & 8.0141                                                                             & 8.5130                                                                             & 11.3154                           & 6.2670                           \\
\hline
\texttt{+ osm\_r0.5}                                                                             & 7.7693                                                                             & 15.6286                                                                            & 8.0011                                                                             & 7.5884                                                                             & 12.9756                                                                            & 8.2571                                                                             & 10.0367                           & 3.4162                           \\
\texttt{+ osm\_r1 }                                                                              & 7.3571                                                                             & 12.8703                                                                            & 8.2306                                                                             & 6.9783                                                                             & 8.0361                                                                             & 7.9985                                                                             & 8.5785                            & 2.1552                           \\
\texttt{+ osm\_r3}                                                                               & 8.0033                                                                             & 20.4238                                                                            & 7.4257                                                                             & 6.3266                                                                             & 8.6313                                                                             & 7.7377                                                                             & 9.7581                            & 5.2802                           \\
\texttt{+ osm\_r5}                                                                             & 7.0056                                                                             & 15.5886                                                                            & 7.0527                                                                             & 5.9760                                                                             & 8.1034                                                                             & 6.9535                                                                             & 8.4466                            & 3.5631                           \\
\texttt{+ osm\_rALL}                                                                             & 11.7540                                                                            & 16.9653                                                                            & 6.7479                                                                             & 10.2338                                                                            & 24.7798                                                                            & 7.9811                                                                             & 13.0770                           & 6.7518                           \\\hline
\texttt{+ postcode\_2 + osm\_r0.5}                                                               & 11.6850                                                                            & 23.1043                                                                            & 8.5513                                                                             & 9.4961                                                                             & 15.2280                                                                            & 8.9336                                                                             & 12.8330                           & 5.6076                           \\
\texttt{+ postcode\_2 + osm\_r1}                                                                 & 9.3478                                                                             & 20.8771                                                                            & 9.2848                                                                             & 7.2474                                                                             & 9.4127                                                                             & 10.4006                                                                            & 11.0951                           & 4.9017                           \\
\texttt{+ postcode\_2 + osm\_r3}                                                                & 10.6179                                                                            & 24.6793                                                                            & 9.8646                                                                             & 8.0089                                                                             & 8.6515                                                                             & 7.8766                                                                             & 11.6165                           & 6.4884                           \\
\texttt{+ postcode\_2 + osm\_r5}                                                                & 9.8460                                                                             & 24.4621                                                                            & 10.0381                                                                            & 7.2279                                                                             & 7.9949                                                                             & 8.3781                                                                             & 11.3245                           & 6.5267                           \\
\texttt{+ postcode\_2 + osm\_rALL}                                                              & 14.7904                                                                            & 50.1174                                                                            & 10.8492                                                                            & 11.3951                                                                            & 39.2400                                                                            & 12.4620                                                                            & 23.1424                           & 17.0864                          \\\hline
\texttt{+ lat\_long + osm\_r0.5}                                                                 & 7.8593                                                                             & 14.0310                                                                            & 7.9618                                                                             & 8.9427                                                                             & 12.7350                                                                            & 8.0921                                                                             & 9.9370                            & 2.7276                           \\
\texttt{+ lat\_long + osm\_r1}                                                                   & 7.1247                                                                             & 10.5546                                                                            & 8.2807                                                                             & 8.6122                                                                             & 7.6839                                                                             & 7.9899                                                                             & 8.3743                            & 1.1837                           \\
\texttt{+ lat\_long + osm\_r3}                                                                  & 7.9893                                                                             & 18.4028                                                                            & 7.5347                                                                             & 6.5009                                                                             & 8.5329                                                                             & 7.6281                                                                             & 9.4315                            & 4.4455                           \\
\texttt{+ lat\_long + osm\_r5 }                                                                 & 7.1155                                                                             & 14.8002                                                                            & 7.1251                                                                             & 5.9370                                                                             & 7.9822                                                                             & 6.9886                                                                             & \textbf{8.3248}                            & \textbf{3.2384}                           \\
\texttt{+ lat\_long + osm\_rALL}                                                                 & 11.5226                                                                            & 20.4139                                                                            & 6.8656                                                                             & 10.5638                                                                            & 25.6137                                                                            & 7.8657                                                                             & 13.8075                           & 7.5136                           \\\hline
\texttt{+ lat\_long + postcode\_2 + osm\_r0.5}                                                     & 11.3376                                                                            & 22.4444                                                                            & 8.4040                                                                             & 7.5357                                                                             & 15.0367                                                                            & 8.8560                                                                             & 12.2691                           & 5.6755                           \\
\texttt{+ lat\_long + postcode\_2 + osm\_r1}                                                       & 9.2410                                                                             & 20.7916                                                                            & 9.2704                                                                             & 7.1959                                                                             & 9.3802                                                                             & 10.3889                                                                            & 11.0447                           & 4.8871                           \\
\texttt{+ lat\_long + postcode\_2 + osm\_r3}                                                       & 10.4435                                                                            & 24.1354                                                                            & 9.8066                                                                             & 9.1656                                                                             & 9.0526                                                                             & 7.9378                                                                             & 11.7569                           & 6.1216                           \\
\texttt{+ lat\_long + postcode\_2 + osm\_r5}                                                       & 9.7223                                                                             & 24.4353                                                                            & 10.0306                                                                            & 7.2250                                                                             & 8.4153                                                                             & 8.5194                                                                             & 11.3913                           & 6.4691                           \\
\texttt{+ lat\_long + postcode\_2 + osm\_rALL}                                                    & 14.5198                                                                            & 51.7616                                                                            & 10.8524                                                                            & 11.6299                                                                            & 38.1968                                                                            & 12.6059                                                                            & 23.2611                           & 17.4044                          \\
\hline
\end{tabular}
\end{adjustbox}
\caption{Test RMSE of the GLM model with different predictor sets: non-geographic features only, and non-geographic features plus (\texttt{lat\_long}), (\texttt{postcode\_2}), or (\texttt{osm\_r*}).}
\label{tab:GLM_base}
\end{table}

Table~\ref{tab:GLM_base} reports the test RMSEs for the classical GLM without regularization. The baseline model, which uses only the non-geographic tabular variables from Table~\ref{Table:VariableDescription} is taken as baseline. Adding geographic coordinates produces a modest but consistent improvement, the mean RMSE decreases by 1.33\%, and the variance between folds drops by 24.48\%, indicating that latitude and longitude capture a meaningful spatial structure absent from the baseline set.

In contrast, incorporating the regional identifier \texttt{postcode\_2} leads to a deterioration in performance, both alone and in combination with coordinates. This suggests that the administrative divisions encoded in the first two postcode digits do not align with the finer scale variation in claims captured by the dataset.

Instead, environmental variables introduce clearer gains. Using OSM-derived features at a radius of 1 km and 5 km reduces the mean RMSE by 2.82\% and 4.31\%, respectively. The 1 km radius also yields a notable reduction in variability across folds, whereas the 5 km radius remains comparably stable. These results indicate that characteristics of the built environment contribute predictive information, particularly when measured on intermediate or broader spatial scales.

The GLM specification with the lowest mean of the test RMSE errors enhances both accuracy and stability, combining latitude and longitude with the 5 km environmental features reduces the mean RMSE by 5.69\% and the standard deviation by 8.95\% relative to the baseline. In contrast, models that combine all radius simultaneously (\texttt{osm\_rALL}) perform poorly, likely reflecting redundancy and reduced model stability when multiple highly correlated spatial representations are included.

Overall, these findings indicate that geographic information improves GLM performance, but the gains depend markedly on the type and spatial scale of the included features.
\begin{table}[htb!]
\begin{adjustbox}{max width=\textwidth}
\begin{tabular}{@{}lcccccccc@{}}
\toprule \hline
\multicolumn{1}{c}{\textbf{}}                                                           & \multicolumn{6}{c}{\textbf{Test RMSE}}                                                                                                                                                                                                                                                                                                                                                              & \textbf{}     & \textbf{}    \\ \hline \midrule
\multicolumn{1}{c}{\textbf{\begin{tabular}[c]{@{}c@{}}GLM\\ Model Family (With regularization)\end{tabular}}} & \textbf{\begin{tabular}[c]{@{}c@{}}Data fold\\ 0\end{tabular}} & \textbf{\begin{tabular}[c]{@{}c@{}}Data fold\\ 1\end{tabular}} & \textbf{\begin{tabular}[c]{@{}c@{}}Data fold\\ 2\end{tabular}} & \textbf{\begin{tabular}[c]{@{}c@{}}Data fold\\ 3\end{tabular}} & \textbf{\begin{tabular}[c]{@{}c@{}}Data fold\\ 4\end{tabular}} & \textbf{\begin{tabular}[c]{@{}c@{}}Data fold\\ 5\end{tabular}} & \textbf{Mean} & \textbf{std} \\ \hline \\Only tabular feature (not geography-related) 
                                                                                   & 6.4995                                                         & 16.3370                                                        & 6.9553                                                         & 7.8492                                                         & 7.9105                                                         & 7.5218                                                         & 8.8455        & 3.7099       \\
\texttt{+ lat\_long}                                                                       & 6.7611                                                         & 14.8204                                                        & 6.8001                                                         & 8.5897                                                         & 7.4681                                                         & 7.1742                                                         & 8.6023        & 3.1186       \\
\texttt{+ postcode2}                                                                      & 6.1943                                                         & 16.8350                                                        & 6.7913                                                         & 7.4813                                                         & 7.8883                                                         & 7.6918                                                         & 8.8137        & 3.9795       \\
\texttt{+ lat\_long + postcode\_2}                                                        & 6.5750                                                         & 14.7952                                                        & 6.8351                                                         & 8.1355                                                         & 7.1885                                                         & 7.2001                                                         & 8.4549        & 3.1508       \\\hline
\texttt{+ osm\_r0.5}                                                                       & 7.5550                                                         & 16.7596                                                        & 7.1647                                                         & 8.1867                                                         & 8.5897                                                         & 7.7226                                                         & 9.3297        & 3.6736       \\
\texttt{+ osm\_r1 }                                                                        & 7.6805                                                         & 13.0719                                                        & 7.9142                                                         & 8.5749                                                         & 8.4469                                                         & 7.8244                                                         & 8.9188        & 2.0654       \\
\texttt{+ osm\_r3}                                                                         & 6.8655                                                         & 14.7952                                                        & 7.5619                                                         & 6.3216                                                         & 8.3297                                                         & 7.6729                                                         & 8.5911        & 3.1174       \\
\texttt{+ osm\_r5}                                                                         & 6.6572                                                         & 14.4736                                                        & 6.7364                                                         & 7.0260                                                         & 7.6901                                                         & 7.0377                                                         & 8.2702        & 3.0607       \\
\texttt{+ osm\_rALL}                                                                       & 8.4069                                                         & 13.5257                                                        & 7.8237                                                         & 9.5501                                                         & 8.6974                                                         & 7.1808                                                         & 9.1974        & 2.2666       \\\hline
\texttt{+ postcode\_2 + osm\_r0.5}                                                         & 7.8824                                                         & 15.2873                                                        & 7.7370                                                         & 8.5572                                                         & 7.9727                                                         & 7.7679                                                         & 9.2008        & 2.9967       \\
\texttt{+ postcode\_2 + osm\_r1}                                                           & 6.9523                                                         & 13.0029                                                        & 8.4276                                                         & 8.9051                                                         & 8.3445                                                         & 7.7784                                                         & 8.9018        & 2.1174       \\
\texttt{+ postcode\_2 + osm\_r3 }                                                          & 7.2840                                                         & 15.0676                                                        & 7.6767                                                         & 6.2570                                                         & 8.3297                                                         & 7.8079                                                         & 8.7371        & 3.1778       \\
\texttt{+ postcode\_2 + osm\_r5}                                                           & 6.5845                                                         & 9.1200                                                         & 6.9612                                                         & 6.2283                                                         & 7.8292                                                         & 7.3133                                                         & 7.3394        & 1.0352       \\
\texttt{+ postcode\_2 + osm\_rALL}                                                         & 10.3595                                                        & 12.1160                                                        & 7.6833                                                         & 7.8122                                                         & 8.3337                                                         & 7.8079                                                         & 9.0188        & 1.8202       \\\hline
\texttt{+ lat\_long + osm\_r0.5}                                                           & 7.5495                                                         & 13.2794                                                        & 7.9892                                                         & 8.1892                                                         & 8.9551                                                         & 7.3370                                                         & 8.8832        & 2.2263       \\
\texttt{+ lat\_long + osm\_r1}                                                             & 6.8704                                                         & 12.7286                                                        & 7.9837                                                         & 8.3562                                                         & 8.3414                                                         & 7.3506                                                         & 8.6051        & 2.1026       \\
\texttt{+ lat\_long + osm\_r3}                                                             & 6.8604                                                         & 15.1655                                                        & 7.6767                                                         & 7.7165                                                         & 8.3297                                                         & 7.5457                                                         & 8.8824        & 3.1135       \\
\texttt{+ lat\_long + osm\_r5}                                                             & 6.5845                                                         & 8.5698                                                         & 6.7096                                                         & 6.1252                                                         & 7.8143                                                         & 7.0832                                                         & \textbf{7.1477}        & \textbf{0.8973}       \\
\texttt{+ lat\_long + osm\_rALL}                                                           & 8.7468                                                         & 13.3690                                                        & 7.3739                                                         & 7.5801                                                         & 8.5448                                                         & 7.4829                                                         & 8.8496        & 2.2884       \\\hline
\texttt{+ lat\_long + postcode\_2 + osm\_r0.5}                                               & 8.0964                                                         & 13.4093                                                        & 7.8332                                                         & 8.4675                                                         & 8.9460                                                         & 7.3617                                                         & 9.0190        & 2.2176       \\
\texttt{+ lat\_long + postcode\_2 + osm\_r1}                                                 & 6.9523                                                         & 13.0029                                                        & 8.4276                                                         & 8.9051                                                         & 8.3445                                                         & 7.7784                                                         & 8.9018        & 2.1174       \\
\texttt{+ lat\_long + postcode\_2 + osm\_r3}                                                 & 6.8725                                                         & 12.0947                                                        & 7.1005                                                         & 8.8382                                                         & 8.2045                                                         & 7.2996                                                         & 8.4017        & 1.9552       \\
\texttt{+ lat\_long + postcode\_2 + osm\_r5}                                                 & 6.8069                                                         & 8.6032                                                         & 6.9231                                                         & 5.9905                                                         & 7.8143                                                         & 7.0996                                                         & 7.2063        & 0.9000       \\
\texttt{+ lat\_long + postcode\_2 + osm\_rALL}                                               & 9.4244                                                         & 12.6254                                                        & 7.2289                                                         & 8.7376                                                         & 8.3008                                                         & 7.3587                                                         & 8.9460        & 1.9847     \\
  \hline
\end{tabular}
\end{adjustbox}
\caption{Test RMSE of the regularized GLM model using different predictor sets. The baseline model uses only non-geographic features. Each alternative model adds additional predictor group to this baseline,  latitude and longitude coordinates (\texttt{lat\_long}), regional categorization (\texttt{postcode\_2}), or environmental features (\texttt{+ osm\_r*}).}
\label{tab:GLM_reg}
\end{table}

Table~\ref{tab:GLM_reg} reports the results for the GLM with ElasticNet regularization. Relative to the unregularized GLM, regularization does not improve performance when only the non-geographic tabular variables are used. However, once geographic coordinates are included, the regularized model performs better than the unregularized baseline, reducing the mean RMSE of the test and its variability by 2.55\% and 12.32\%, respectively. Regularization also stabilizes the contribution of the regional identifier \texttt{postcode\_2}, while adding \texttt{lat\_long} and \texttt{postcode\_2} previously degraded performance, the regularized model achieves reductions of 4.22\% in mean RMSE and 11.41\% in standard deviation relative to the baseline of unregularized GLM. This highlights the value of ElasticNet in managing collinearity and censoring coefficients for weak categorical predictors.

Within the regularized GLM family itself, the inclusion of either \texttt{lat\_long}, \texttt{postcode\_2}, or their combination yields incremental gains. Compared to the regularized baseline, these specifications reduce the mean RMSE by up to 4.42\% and reduce the standard deviation in folds by up to 15.94\%. This suggests that claim frequency at the postcode level has a clear spatial component and that location features capture predictive signal beyond what is explained by traditional actuarial variables, while also improving model stability.

Turning to the environmental features, the 3 km and 5 km radius provide consistent improvements. Relative to the regularized baseline, the 3 km radius decreases the mean RMSE by 2.88\% and its variability by 15.97\%, while the 5 km radius achieves reductions of 6.50\% and 17.50\%, respectively. Notably, combining \texttt{postcode\_2} with the 5 km environmental features produces a substantial improvement, with a 17.03\% reduction in mean RMSE and a 72.10\% reduction in standard deviation. This suggests that broader scale built environment indicators help extract meaningful structure from postcode level administrative categories.

The best overall performance is achieved when \texttt{lat\_long} and the 5 km environmental features are used together. Compared to the regularized baseline, this specification reduces the mean RMSE of the test by 19.19\% and the standard deviation in the folds by 75.81\%. Improvements are also substantial compared to the unregularized GLM baseline, with reductions of 19.02\% in mean RMSE and 74.77\% in variability. Overall, these results show that ElasticNet regularization helps the model make better use of spatial features by reducing overfitting and stabilizing the coefficients.

\begin{table}[H]
\begin{adjustbox}{max width=\textwidth}
\begin{tabular}{@{}lcccccccc@{}}
\toprule \hline
\multicolumn{1}{c}{\textbf{}}                                                           & \multicolumn{6}{c}{\textbf{Test RMSE}}                                                                                                                                                                                                                                                                                                                                                              & \textbf{}     & \textbf{}    \\ \hline \midrule
\multicolumn{1}{c}{\textbf{\begin{tabular}[c]{@{}c@{}}XGB\\ Model Family\end{tabular}}} & \textbf{\begin{tabular}[c]{@{}c@{}}Data fold\\ 0\end{tabular}} & \textbf{\begin{tabular}[c]{@{}c@{}}Data fold\\ 1\end{tabular}} & \textbf{\begin{tabular}[c]{@{}c@{}}Data fold\\ 2\end{tabular}} & \textbf{\begin{tabular}[c]{@{}c@{}}Data fold\\ 3\end{tabular}} & \textbf{\begin{tabular}[c]{@{}c@{}}Data fold\\ 4\end{tabular}} & \textbf{\begin{tabular}[c]{@{}c@{}}Data fold\\ 5\end{tabular}} & \textbf{Mean} & \textbf{std} \\ \hline \\Only tabular feature (not geography-related) 
                                                                                        & 7.3506                                                         & 21.0880                                                        & 7.8936                                                         & 9.3553                                                         & 7.3553                                                         & 6.6217                                                         & 9.9441        & 5.5354       \\
\texttt{+ lat\_long}                                                                             & 7.0220                                                         & 20.6761                                                        & 8.2752                                                         & 9.5827                                                         & 7.1793                                                         & 6.0245                                                         & 9.7933        & 5.4681       \\
\texttt{+ postcode\_2 }                                                                          & 6.7627                                                         & 20.1609                                                        & 7.9394                                                         & 8.7377                                                         & 7.4438                                                         & 6.3582                                                         & 9.5671        & 5.2581       \\
\texttt{+ lat\_long + postcode\_2 }                                                              & 6.9450                                                         & 20.3917                                                        & 8.4242                                                         & 7.9689                                                         & 7.1131                                                         & 5.9933                                                         & 9.4727        & 5.4157       \\ \hline
\texttt{+ osm\_r0.5 }                                                                            & 8.4985                                                         & 21.8860                                                        & 8.4325                                                         & 9.6472                                                         & 7.4518                                                         & 6.2382                                                         & 10.3590       & 5.7617       \\
\texttt{+ osm\_r1}                                                                               & 6.4865                                                         & 17.7601                                                        & 8.6516                                                         & 7.5292                                                         & 8.0547                                                         & 6.5519                                                         & 9.1723        & 4.2907       \\
\texttt{+ osm\_r3}                                                                               & 7.2139                                                         & 13.3785                                                        & 9.2327                                                         & 7.1156                                                         & 8.9856                                                         & 6.3032                                                         & 8.7049        & 2.5580       \\
\texttt{+ osm\_r5}                                                                               & 6.7091                                                         & 14.7900                                                        & 7.8815                                                         & 8.2657                                                         & 7.4943                                                         & 5.8285                                                         & 8.4949        & 3.2049       \\
\texttt{+ osm\_rALL  }                                                                           & 6.4763                                                         & 15.8528                                                        & 7.3577                                                         & 9.1074                                                         & 7.6594                                                         & 6.0134                                                         & 8.7445        & 3.6432       \\ \hline
\texttt{+ postcode\_2 + osm\_r0.5 }                                                              & 8.0432                                                         & 20.4022                                                        & 8.5606                                                         & 8.4692                                                         & 7.2048                                                         & 5.9562                                                         & 9.7727        & 5.2973       \\
\texttt{+ postcode\_2 + osm\_r1}                                                                 & 6.4325                                                         & 17.7433                                                        & 8.8194                                                         & 7.1778                                                         & 8.1291                                                         & 6.3755                                                         & 9.1129        & 4.3350       \\
\texttt{+ postcode\_2 + osm\_r3 }                                                                & 7.6820                                                         & 16.2360                                                        & 8.2319                                                         & 7.2402                                                         & 8.3549                                                         & 6.4577                                                         & 9.0338        & 3.5960       \\
\texttt{+ postcode\_2 + osm\_r5}                                                                 & 6.4397                                                         & 15.9249                                                        & 7.3728                                                         & 7.7777                                                         & 7.3442                                                         & 5.8877                                                         & 8.4578        & 3.7233       \\
\texttt{+ postcode\_2 + osm\_rALL }                                                              & 6.6269                                                         & 15.9843                                                        & 7.3475                                                         & 7.5941                                                         & 7.6691                                                         & 6.0445                                                         & 8.5444        & 3.6981       \\ \hline
\texttt{+ lat\_long + osm\_r0.5}                                                                 & 7.8943                                                         & 20.8628                                                        & 8.3087                                                         & 8.1702                                                         & 7.3352                                                         & 5.7803                                                         & 9.7253        & 5.5335       \\
\texttt{+ lat\_long + osm\_r1}                                                                   & 6.4865                                                         & 17.7601                                                        & 8.6516                                                         & 7.5292                                                         & 8.0547                                                         & 6.5519                                                         & 9.1723        & 4.2907       \\
\texttt{+ lat\_long + osm\_r3    }                                                               & 7.5996                                                         & 14.9472                                                        & 8.4330                                                         & 6.8024                                                         & 8.2400                                                         & 6.4131                                                         & 8.7392        & 3.1414       \\
\texttt{+ lat\_long + osm\_r5}                                                                   & 7.2411                                                         & 16.3658                                                        & 7.3381                                                         & 7.6368                                                         & 6.4273                                                         & 5.9605                                                         & 8.4949        & 3.9063       \\
\texttt{+ lat\_long + osm\_rALL }                                                                & 6.3242                                                         & 15.5611                                                        & 7.5313                                                         & 7.9427                                                         & 7.2261                                                         & 5.8486                                                         & \textbf{8.4057}        & \textbf{3.5900}       \\ \hline
\texttt{+ lat\_long + postcode2 + osm\_r0.5 }                                                    & 8.0432                                                         & 20.4022                                                        & 8.5606                                                         & 8.4692                                                         & 7.2048                                                         & 5.9562                                                         & 9.7727        & 5.2973       \\
\texttt{+ lat\_long + postcode2 + osm\_r1 }                                                      & 7.0164                                                         & 18.7171                                                        & 8.2524                                                         & 7.2968                                                         & 8.2006                                                         & 5.7033                                                         & 9.1978        & 4.7558       \\
\texttt{+ lat\_long + postcode2 + osm\_r3 }                                                      & 7.5898                                                         & 18.2558                                                        & 8.8232                                                         & 6.8659                                                         & 7.0991                                                         & 6.6203                                                         & 9.2090        & 4.5002       \\
\texttt{+ lat\_long + postcode2 + osm\_r5}                                                       & 6.4733                                                         & 16.9851                                                        & 7.4738                                                         & 8.5570                                                         & 7.1035                                                         & 5.8091                                                         & 8.7336        & 4.1479       \\
\texttt{+ lat\_long + postcode2 + osm\_rALL}                                                     & 6.4714                                                         & 16.5456                                                        & 7.8926                                                         & 7.7930                                                         & 6.7124                                                         & 5.9188                                                         & 8.5556        & 3.9889       \\ \bottomrule
\end{tabular}
\end{adjustbox}
\caption{Test RMSE results for the XGB model using different predictor sets: non-geographic features only, and non-geographic features plus (\texttt{lat\_long}), (\texttt{postcode\_2}), or (\texttt{osm\_r*}).}
\label{tab:XGB}
\end{table}

Table~\ref{tab:XGB} reports the predictive performance of the XGB model across the six folds. None of the XGB specifications outperform the best-performing model identified so far, namely the regularized GLM with \texttt{lat\_long + osm\_r5}. Nevertheless, several patterns emerge that are consistent with the behavior observed in the GLM framework.

First, incorporating geographic coordinates or the regional identifier improves performance compared with the XGB baseline that uses only the non-geographic tabular features. Including \texttt{lat\_long + postcode\_2} reduces the mean test RMSE and its standard deviation by 4.74\% and 2.16\%, respectively. This suggests that XGB can extract a useful spatial structure from coarse geographic attributes. 

In contrast to the GLM results, environmental features alone provide meaningful improvements across a wider-range-of-radius models using OSM-CORINE-derived variables from 1 km up to the combined \texttt{rALL} specification that performs better than the baseline. Among these, the combination of \texttt{postcode\_2 + osm\_r5} delivers the second-best performance within the XGB family.

The strongest XGB model is obtained by combining coordinates with all environmental radius, this is the baseline XGB model \texttt{+ lat\_long + osm\_rALL}, which reduces the mean RMSE by 15.47\% and the standard deviation by 35.14\% relative to the baseline. Thus, unlike the GLM framework, where a single broader radius (5 km) provided the clearest benefit, XGB appears to leverage multi-scale environmental information more effectively, indicating that tree-based models may better accommodate overlapping spatial features. This reflects the ability of the XGB to exploit complementary information on multiple spatial scales rather than being affected by redundancy, indicating that feature selection strategies should differ by model, in this experiment, while GLMs benefit from selecting a single representative radius, XGB benefits from retaining several spatial scale features.

\begin{table}[htb!]
\begin{adjustbox}{max width=\textwidth}
\begin{tabular}{@{}llcccccccc@{}}
\toprule \hline
                                                                        & \multicolumn{1}{c}{\textbf{}}                                                           & \multicolumn{6}{c}{\textbf{Test RMSE}}                                                                                                                                                                                                                                                                                                                                                              & \textbf{}     & \textbf{}    \\ \hline \midrule
                                                                        & \multicolumn{1}{c}{\textbf{\begin{tabular}[c]{@{}c@{}}CNN\\ Model Family\end{tabular}}} & \textbf{\begin{tabular}[c]{@{}c@{}}Data fold\\ 0\end{tabular}} & \textbf{\begin{tabular}[c]{@{}c@{}}Data fold\\ 1\end{tabular}} & \textbf{\begin{tabular}[c]{@{}c@{}}Data fold\\ 2\end{tabular}} & \textbf{\begin{tabular}[c]{@{}c@{}}Data fold\\ 3\end{tabular}} & \textbf{\begin{tabular}[c]{@{}c@{}}Data fold\\ 4\end{tabular}} & \textbf{\begin{tabular}[c]{@{}c@{}}Data fold\\ 5\end{tabular}} & \textbf{Mean} & \textbf{std} \\ \midrule
                                                                        &                                                                                         &                                                                &                                                                &                                                                &                                                                &                                                                &                                                                &               &              \\
\multirow{4}{*}{\begin{tabular}[c]{@{}l@{}}Only \\ Images\end{tabular}} & \texttt{+ img\_0.5}                                                             & 16.7106                                                        & 11.7379                                                        & 11.7519                                                        & 17.5654                                                        & 10.6739                                                        & 13.1902                                                        & 13.6050       & 2.8640       \\
                                                                        & \texttt{+ img\_1}                                                               & 8.5441                                                         & 36.7624                                                        & 11.8488                                                        & 13.3001                                                        & 13.7133                                                        & 11.8641                                                        & 16.0055       & 10.3298      \\
                                                                        & \texttt{+ img\_3}                                                               & 7.5001                                                         & 29.2607                                                        & 7.9710                                                         & 11.1879                                                        & 9.3834                                                         & 9.9939                                                         & 12.5495       & 8.2962       \\
                                                                        & \texttt{+ img\_ALL}                                                                & 8.2133                                                         & 8.6910                                                         & 9.0714                                                         & 12.9878                                                        & 11.5601                                                        & 10.7732                                                        & 10.2161       & 1.8680       \\ \midrule
\multirow{4}{*}{Multimodal}                                             & \texttt{+ img\_0.5}                                                            & 9.5394                                                         & 21.8794                                                        & 10.1902                                                        & 14.2065                                                        & 8.4701                                                         & 9.5166                                                         & 12.3004       & 5.0968       \\
                                                                        & \texttt{+ img\_1}                                                              & 6.6800                                                         & 9.4210                                                         & 9.7421                                                         & 12.7323                                                        & 10.7801                                                        & 7.7372                                                         & 9.5154        & 2.1545       \\
                                                                        & \texttt{+ img\_3}                                                               & 10.4765                                                        & 12.8569                                                        & 8.4087                                                         & 7.2116                                                         & 8.0258                                                         & 12.0133                                                        & 9.8321        & 2.3012       \\
                                                                        & \texttt{+ img\_ALL}                                                                & 5.5693                                                         & 9.4121                                                         & 9.0685                                                         & 8.8177                                                         & 8.6116                                                         & 12.1636                                                        & \textbf{8.9405}        & \textbf{2.1031}       \\ \bottomrule
\end{tabular}
\end{adjustbox}
\caption{Test RMSE results using CNN model, namely ResNet18, using images only or concatenated with tabular predictors that are not associated to geography. The image tiles with a \texttt{r} km apothem length are represented by \texttt{img\_r}.}
\label{tab:CNN_1}
\end{table}

With respect to the image-based models, we first evaluate the predictive performance of the CNN using only imagery, and then in a multimodal setting where the learned image representations are concatenated with the non-geographic tabular predictors (Table~\ref{tab:CNN_1}). As expected, models trained solely on images perform noticeably worse than all tabular, regularized GLM, and XGB specifications. Their RMSEs remain high across folds, indicating that visual information alone is not sufficient to explain the variation in claim counts at the postcode level. Nevertheless, the standard deviations for the 0.5 km and ALL-radius configurations are lower than those observed for several XGB models, suggesting more stable behavior despite weaker accuracy.

A more informative pattern emerges once non-geographic tabular features are added. Across all radius configurations, multimodal models achieve lower mean RMSE than their image only counterparts. For example, the model combining all radii achieves a 12.49\% reduction in mean RMSE relative to the use of images alone. The multimodal CNN with all radii also outperforms the XGB models that rely on non-geographic features or combinations of \texttt{lat\_long}, \texttt{postcode\_2}, or both.

Compared to regularized GLMs fitted with tabular or geographic features, excluding OSM-derived environmental variables, multimodal CNN using images at all spatial scales remains less accurate in terms of mean RMSE: its error increases by up to 5.74\% relative to the Elastic Net model with \texttt{lat\_long + postcode\_2}. However, the CNN exhibits substantially greater stability across folds, reducing the standard deviation of the test RMSE by between 32.56\% and 47.15\%. Although this improvement in robustness is notable, the multimodal CNN still does not surpass the best performing regularized GLM, which incorporates the constructed environmental indicators derived from OSM and CORINE.

After analyzing the image-only and multimodal models (including tabular non-geographic-related features), we proceeded to evaluate configurations that combine the Ortho95 imagery with geography-related features. We did not test all possible predictor combinations. Instead, to avoid any form of data leakage, the selection of candidate configurations was guided exclusively by the within fold 5-fold cross-validation results, without reference to the held-out test folds. Table~\ref{tab:CV_image_tab_selection} reports the 5-CV RMSEs for each fold, together with their mean and standard deviation between folds.

Across all model families, also including non-geographic tabular features, the environmental features computed using a 5 km radius consistently achieved the lowest average 5-CV RMSE, including in the Poisson MLP. We therefore selected the 5 km radius for the subsequent experiments. For the imagery component, tiles with 3 km apothem length obtained the lowest average 5-CV RMSE across folds, and this configuration was retained for the combined tabular and imagery models.

\begin{table}[]
\begin{adjustbox}{max width=\textwidth}
\begin{tabular}{@{}clcccccccc@{}}
\toprule \hline
\multicolumn{1}{l}{}                                                                     &                             & \multicolumn{6}{c}{\textbf{5-CV RMSEs scores}}                                                                                                    & \multicolumn{1}{l}{}              & \multicolumn{1}{l}{}             \\ \hline \midrule
\multicolumn{2}{c}{\textbf{Model}}                                                                                     & \textbf{Data fold   0}         & \textbf{Data fold 1} & \textbf{Data fold 2} & \textbf{Data fold 3} & \textbf{Data fold 4} & \textbf{Data fold 5} & \multicolumn{1}{l}{\textbf{Mean}} & \multicolumn{1}{l}{\textbf{std}} \\ \midrule
                                                                                         & \texttt{+ osm\_r0.5}                 & 11.8860                        & 9.2865               & 9.9218               & 11.0611              & 11.2619              & 11.0658              & 10.7472                           & 0.9567                           \\
                                                                                         & \texttt{+ osm\_r1}                   & 8.7681                         & 8.2109               & 8.4321               & 9.4886               & 9.2520               & 8.9381               & 8.8483                            & 0.4831                           \\
                                                                                         & \texttt{+ osm\_r3}                   & 10.0644                        & 7.3636               & 9.9460               & 10.5157              & 9.9932               & 10.8230              & 9.7843                            & 1.2346                           \\
                                                                                         & \texttt{+ osm\_r5 }                  & 9.7167                         & 7.0466               & 8.3415               & 8.8130               & 8.0089               & 9.0475               & \textbf{8.4957}                   & 0.9227                           \\
\multirow{-5}{*}{\textbf{GLM}}                                                           & \texttt{+ osm\_rALL}                 & 16.0267                        & 11.7381              & 14.2717              & 16.8146              & 14.2404              & 15.9705              & 14.8437                           & 1.8379                           \\ \midrule
                                                                                         &\texttt{+ osm\_r0.5}                 & \cellcolor[HTML]{FFFFFF}8.5905 & 7.8946               & 8.2345               & 9.0884               & 8.7557               & 9.6476               & 8.7019                            & 0.6212                           \\
                                                                                         &\texttt{+ osm\_r1}                   & 8.0272                         & 7.5131               & 7.8240               & 8.5143               & 8.1487               & 8.3749               & 8.0670                            & 0.3658                           \\
                                                                                         & \texttt{+ osm\_r3 }                  & 7.7187                         & 6.9107               & 7.4877               & 7.6342               & 7.1249               & 7.5977               & 7.4123                            & 0.3217                           \\
                                                                                         &\texttt{+ osm\_r5}                   & 7.5968                         & 6.9646               & 7.2765               & 7.4759               & 7.0757               & 7.4175               & \textbf{7.3012}                   & 0.2434                           \\
\multirow{-5}{*}{\textbf{\begin{tabular}[c]{@{}c@{}}Regularized\\ GLM\end{tabular}}}     & \texttt{+ osm\_rALL}                 & 8.0346                         & 7.5659               & 7.0848               & 8.3448               & 7.8601               & 8.5178               & 7.9013                            & 0.5245                           \\ \midrule
                                                                                         & \texttt{+ osm\_r0.5}                 & 9.7680                         & 7.2390               & 9.7675               & 8.9525               & 9.0127               & 9.6095               & 9.0582                            & 0.9627                           \\
                                                                                         & \texttt{+ osm\_r1 }                  & 9.2999                         & 7.1636               & 8.4835               & 9.2896               & 8.8327               & 9.2827               & 8.7253                            & 0.8328                           \\
                                                                                         & \texttt{+ osm\_r3}                   & \cellcolor[HTML]{FFFFFF}7.6593 & 7.2876               & 8.5180               & 8.3234               & 8.1741               & 8.7391               & 8.1169                            & 0.5461                           \\
                                                                                         & \texttt{+ osm\_r5}                   & 7.9869                         & 6.8839               & 7.3060               & 7.9489               & 6.9716               & 7.9350               & \textbf{7.5054}                   & 0.5146                           \\
\multirow{-5}{*}{\textbf{XGB}}                                                           & \texttt{+ osm\_rALL}               & \cellcolor[HTML]{FFFFFF}7.6082 & 7.1273               & 7.4254               & 7.8124               & 7.1596               & 8.1094               & 7.5404                            & 0.3824                           \\ \midrule
                                                                                         & \texttt{+ osm\_r0.5}                 & 14.0608                        & 14.8741              & 14.6908              & 17.3455              & 15.1118              & 15.9372              & 15.3367                           & 1.1581                           \\
                                                                                         &\texttt{+ osm\_r1}                   & 15.3614                        & 16.5362              & 16.0515              & 15.7480              & 15.8154              & 15.5550              & 15.8446                           & 0.4120                           \\
                                                                                         & \texttt{+ osm\_r3 }                  & 17.9654                        & 10.7951              & 20.1036              & 16.4659              & 19.9642              & 16.6464              & 16.9901                           & 3.4135                           \\
                                                                                         & \texttt{+ osm\_r5}                   & 12.0575                        & 12.5549              & 14.9637              & 11.9644              & 15.6071              & 15.6627              & \textbf{13.8017}                  & 1.7914                           \\
\multirow{-5}{*}{\textbf{MLP}}                                                           & \texttt{+ osm\_rALL }                & 18.8669                        & 16.1258              & 22.8786              & 18.1097              & 25.6471              & 22.3702              & 20.6664                           & 3.5496                           \\ \midrule
                                                                                         & \texttt{+ img\_0.5 }& 12.1341                        & 12.4189              & 12.2882              & 13.1753              & 11.4482              & 12.5700              & 12.3391                           & 0.5652                           \\
                                                                                         & \texttt{+ img\_1}  & 10.9387                        & 11.7947              & 10.5285              & 12.9036              & 10.9100              & 11.3613              & 11.4061                           & 0.8521                           \\
                                                                                         & \texttt{+ img\_3}  & 10.3863                        & 9.3735               & 10.1485              & 9.9428               & 9.9427               & 9.0853               & \textbf{9.8132}                   & 0.4894                           \\
\multirow{-4}{*}{\textbf{\begin{tabular}[c]{@{}c@{}}CNN \\ (Only images )\end{tabular}}} &\texttt{+ img\_ALL}   & 10.1841                        & 9.9870               & 10.3609              & 10.2221              & 10.1732              & 10.7859              & 10.2855                           & 0.2728                           \\ \bottomrule
\end{tabular}
\end{adjustbox}
\caption{Six-fold cross-validation results using tabular predictors (only tabular non-geographic and OSM–CORINE features) for GLM, regularized GLM, XGB, and MLP. The CNN results use image data only.}
\label{tab:CV_image_tab_selection}
\end{table}

After identifying the 3 km image tiles and the 5 km OSM–CORINE radius as the best performing configurations in the cross-validation analysis, we evaluated several CNN models that combine images with different geographic predictors. Table \ref{tab:CNN_2} reports the test RMSEs across the six folds for the selected specifications, where all models include the non-geographic tabular features.

\begin{table}[htb!]
\begin{adjustbox}{max width=\textwidth}
\begin{tabular}{@{}clcccccccc@{}}
\toprule \hline
\textbf{}                                                                                                                & \multicolumn{1}{c}{\textbf{}} & \multicolumn{8}{c}{\textbf{Test RMSE}}                                                                                                                                                                                                                                                                                                                                                                                                                                                                                                                                                                   \\ \hline \midrule
\multicolumn{2}{c}{\textbf{Model}}                                                                                                                       & \multicolumn{1}{c}{\textbf{\begin{tabular}[c]{@{}c@{}}Data fold \\ 0\end{tabular}}} & \multicolumn{1}{c}{\textbf{\begin{tabular}[c]{@{}c@{}}Data fold \\ 1\end{tabular}}} & \multicolumn{1}{c}{\textbf{\begin{tabular}[c]{@{}c@{}}Data fold \\ 2\end{tabular}}} & \multicolumn{1}{c}{\textbf{\begin{tabular}[c]{@{}c@{}}Data fold \\ 3\end{tabular}}} & \multicolumn{1}{c}{\textbf{\begin{tabular}[c]{@{}c@{}}Data fold \\ 4\end{tabular}}} & \multicolumn{1}{c}{\textbf{\begin{tabular}[c]{@{}c@{}}Data fold \\ 5\end{tabular}}} & \multicolumn{1}{c}{\textbf{Mean}} & \multicolumn{1}{c}{\textbf{std}} \\ \midrule
\multirow{7}{*}{\textbf{\begin{tabular}[c]{@{}c@{}}CNN\\ with \texttt{img\_3 +}\\ non-geographic \\ tabular \\ features\end{tabular}}} &       \texttt{+ lat\_long}                        & 8.5535                                                                              & 12.0836                                                                             & 8.0545                                                                              & 12.7207                                                                             & 8.3715                                                                              & 11.9141                                                                             & 10.2830                           & 2.1659                           \\
                                                                                                                         &  \texttt{+ postcode\_2}                             & 8.4339                                                                              & 21.3710                                                                             & 6.3233                                                                              & 19.2682                                                                             & 7.7500                                                                              & 11.0043                                                                             & 12.3584                           & 6.3854                           \\
                                                                                                                         &        \texttt{+ postcode\_2 + lat\_long}                        & 8.1223                                                                              & 13.3367                                                                             & 7.0583                                                                              & 7.0249                                                                              & 7.9078                                                                              & 10.7980                                                                             & \textbf{9.0413}                            & \textbf{2.5165}                           \\
                                                                                                                         &        \texttt{+ osm\_r5}                        & 6.1239                                                                              & 17.0030                                                                             & 7.4897                                                                              & 7.1114                                                                              & 9.2627                                                                              & 8.5322                                                                              & 9.2538                            & 3.9514                           \\
                                                                                                                         &       \texttt{+ osm\_r5 + lat\_long}                        & 9.0063                                                                              & 9.7907                                                                              & 6.8490                                                                              & 11.0816                                                                             & 9.2256                                                                              & 8.5524                                                                              & 9.0843                   & 1.3996                           \\
                                                                                                                         &       \texttt{+ osm\_r5 + postcode\_2}                         & 8.3520                                                                              & 14.7986                                                                             & 7.0970                                                                              & 11.8137                                                                             & 8.3244                                                                              & 13.8385                                                                             & 10.7040                           & 3.2257                           \\
                                                                                                                         &         \texttt{+ osm\_r5 + postcode\_2 + lat\_long}                 & 7.3423                                                                              & 11.7226                                                                             & 7.1467                                                                              & 12.4166                                                                             & 8.6854                                                                              & 10.9374                                                                             & 9.7085                            & 2.2851                           \\ \bottomrule
\end{tabular}
\end{adjustbox}
\caption{Test RMSEs for convolutional neural network models using 3 km orthoimage tiles, combined with non-geographic tabular features and selected geographic predictors (\texttt{lat\_long}, \texttt{postcode\_2}, and \texttt{osm\_r5}). For each specification, RMSEs are shown across the six test folds, along with the mean and standard deviation.}
\label{tab:CNN_2}
\end{table}

The results show that augmenting the image-only CNN with any individual geographic variable, \texttt{lat\_long}, \texttt{postcode\_2}, or the 5 km environmental features does not outperform the multimodal model using image tiles from all radius and the non-geographic tabular features (Table \ref{tab:CNN_1}). Although combinations with \texttt{+ osm\_r5 + lat\_long}, yield a reduction in variability, no combination improves the predictive accuracy of the best GLM with regularization and 5 km environmental features. In short, while image information contributes additional structure, CNN models, whether unimodal or multimodal, do not surpass the performance achieved using tabular environmental indicators derived from the 5 km radius, showcasing the need for additional data engineering for better accuracy, instead of relying exclusively on automated methods.

Given that CNN models that use images directly, alone or in multimodal form, do not outperform the GLM, regularized GLM or XGB models, we next examined whether image embeddings could provide a stronger and more concentrated spatial signal (see Subsection~\ref{subsec:VisualFeatures}). The corresponding test RMSE results are reported in Table~\ref{tab:Embeddings}.

\begin{table}[H]
\begin{adjustbox}{max width=\textwidth}
\begin{tabular}{@{}lccccccccccc@{}}
\toprule \hline
                                                                                                      & \multicolumn{1}{l}{} & \multicolumn{10}{c}{\textbf{Test RMSE}}                                                                                                                                                                                                                                                                                                                                                                                                                                                                                     \\ \hline \midrule
                                                                                                      & \multicolumn{1}{l}{} & \multicolumn{1}{l}{}                                           & \multicolumn{1}{l}{}                                           & \multicolumn{1}{l}{}                                           & \multicolumn{1}{l}{}                                           & \multicolumn{1}{l}{}                                           & \multicolumn{1}{l}{}                                           & \multicolumn{1}{l}{} & \multicolumn{1}{l}{} & \multicolumn{2}{l}{\textbf{Without Embeddings}} \\ \midrule
\multicolumn{1}{c}{\textbf{Model}}                                                                    & \textbf{Embeddings}  & \textbf{\begin{tabular}[c]{@{}c@{}}Data fold\\ 0\end{tabular}} & \textbf{\begin{tabular}[c]{@{}c@{}}Data fold\\ 1\end{tabular}} & \textbf{\begin{tabular}[c]{@{}c@{}}Data fold\\ 2\end{tabular}} & \textbf{\begin{tabular}[c]{@{}c@{}}Data fold\\ 3\end{tabular}} & \textbf{\begin{tabular}[c]{@{}c@{}}Data fold\\ 4\end{tabular}} & \textbf{\begin{tabular}[c]{@{}c@{}}Data fold\\ 5\end{tabular}} & \textbf{Mean}        & \textbf{std}         & \textbf{Mean}                       & \textbf{std}                      \\ \midrule
                                                                                                      & \texttt{ResNet18\_emb}          & 6.7747                                                         & 7.8011                                                         & 7.4498                                                         & 6.9819                                                         & 8.7097                                                         & 7.6757                                                         & 7.5655               & 0.6865               & 7.1477                              & 0.8973                            \\
\multirow{-2}{*}{\textbf{\begin{tabular}[c]{@{}l@{}}Regularized\\ GLM \texttt{+ lat\_long + osm\_r5}\end{tabular}}} & \texttt{Nomicv15\_emb}         & 6.4529                                                         & 12.6302                                                        & 6.5770                                                         & 7.7490                                                         & 7.8969                                                         & 7.4570                                                         & 8.1272               & 2.2860               & 7.1477                              & 0.8973                            \\
                                                                                                      & \texttt{ResNet18\_emb}          & 7.1135                                                         & 16.3616                                                        & 8.0293                                                         & 11.8236                                                        & 7.8120                                                         & 9.1947                                                         & 10.0558              & 3.5051               & 8.4949                              & 3.9063                            \\
\multirow{-2}{*}{\textbf{XGB \texttt{+ lat\_long + osm\_r5}}}                                                         & \texttt{Nomicv15\_emb}          & 6.5822                                                         & 15.2512                                                        & 7.9744                                                         & 7.6572                                                         & 6.7252                                                         & 7.1014                                                         & 8.5486               & 3.3267               & 8.4949                              & 3.9063                            \\ \midrule
                                                                                                      & \texttt{ResNet18\_emb}          & 6.2986                                                         & 15.5075                                                        & 7.9461                                                         & 7.4211                                                         & 8.5647                                                         & 7.6814                                                         & 8.9032               & 3.3203               & 8.6023                              & 3.1186                            \\
\multirow{-2}{*}{\textbf{\begin{tabular}[c]{@{}l@{}}Regularized \\ GLM \texttt{+ lat\_long}\end{tabular}}}        & \texttt{Nomicv15\_emb}          & 6.1835                                                         & 11.7991                                                        & 6.0264                                                         & 10.1658                                                        & 8.4974                                                         & 6.4155                                                         & \textbf{8.1813}      & \textbf{2.4033}      & 8.6023                              & 3.1186                            \\
                                                                                                      & \texttt{ResNet18\_emb}          & 6.9960                                                         & 17.7494                                                        & 8.9422                                                         & 10.8737                                                        & 8.6479                                                         & 9.0698                                                         & 10.3798              & 3.8154               & 9.7933                              & 5.4681                            \\
\multirow{-2}{*}{\textbf{XGB \texttt{+ lat\_long}}}                                                              & \texttt{Nomicv15\_emb}          & 8.6472                                                         & 16.9038                                                        & 8.6920                                                         & 12.4303                                                        & 8.9521                                                         & 8.0626                                                         & 10.6147              & 3.4553               & 9.7933                              & 5.4681                            \\ \midrule
                                                                                                      & \texttt{ResNet18\_emb}          & 6.2739                                                         & 15.0133                                                        & 7.9978                                                         & 8.5322                                                         & 8.1097                                                         & 7.4356                                                         & 8.8937               & 3.0980               & 8.8455                              & 3.7099                            \\
\multirow{-2}{*}{\textbf{Regularized GLM}}                                                            & \texttt{Nomicv15\_emb}          & 6.2931                                                         & 13.0527                                                        & 6.1718                                                         & 9.7725                                                         & 8.5981                                                         & 7.0254                                                         & \textbf{8.4856}      & \textbf{2.6389}      & 8.8455                              & 3.7099                            \\
                                                                                                      & \texttt{ResNet\_18emb}          & 6.6286                                                         & 19.7049                                                        & 9.0149                                                         & 10.3526                                                        & 9.0736                                                         & 9.4319                                                         & 10.7011              & 4.5799               & 9.9441                              & 5.5354                            \\
\multirow{-2}{*}{\textbf{XGB}}                                                                        & \texttt{Nomicv15\_emb}          & 8.5232                                                         & 15.4379                                                        & 8.6180                                                         & 12.2891                                                        & 8.1439                                                         & 8.0338                                                         & 10.1743              & 3.0342               & 9.9441                              & 5.5354                            \\ \bottomrule
\end{tabular}
\end{adjustbox}
\caption{Test RMSEs for GLM and XGB models, whose predictors are the non-geographic tabular features or accompanied by geographic ones (\texttt{lat\_long}, \texttt{postcode\_2}, \texttt{osm\_r5}, and the image embeddings \texttt{ResNet\_18emb} or \texttt{Nomicv15\_emb}). For each specification, RMSEs are shown across the six test folds, along with the mean and standard deviation, as well as the mean and std upon the model without the embeddings.}
\label{tab:Embeddings}
\end{table}

The initial predictor sets were chosen using the mean 5-fold CV scores from the previous experiments. Across GLM, regularized GLM, and XGB , the configuration combining non-geographic tabular variables with latitude, longitude, and the 5 km OSM–CORINE features (\texttt{+ lat\_long + osm\_r5}) systematically achieved the best validation performance. We therefore took this as a starting point and then progressively removed components (first \texttt{osm\_r5}, then \texttt{lat\_long}) to assess the incremental value of the ResNet18 and Nomic v1.5 embeddings. The last set of specifications uses embeddings as the only geographic information, together with the non-geographic tabular features.

For models that already include the 5 km OSM–CORINE features, adding ResNet18 or Nomic embeddings does not improve the mean RMSE over the test set. In these configurations, the mean RMSE increases. However, in several cases, the cross-fold variability decreases substantially, with reductions ranging from approximately 10.27\% to 23.49\% relative to the corresponding baseline. These reductions indicate that embeddings can make the spatial signal of the model more stable across partitions, although they do not improve predictive accuracy when strong environmental predictors are already present.

The picture is different when we remove the OSM–CORINE features. For the regularized GLM with latitude and longitude but without \texttt{osm\_r5}, the addition of Nomic embeddings (\texttt{Nomicv15\_emb}) improves both accuracy and stability, the mean test RMSE decreases by  4.89\% and the standard deviation by 22.94\% compared to the baseline without embeddings. A similar pattern appears when we add Nomic embeddings to the regularized GLM with only non-geographic tabular features, where both mean RMSE and variance are reduced by 4.07\% and 28.87\%, respectively. These findings show that transformer-based embeddings encode a meaningful spatial structure that becomes valuable precisely when explicit environmental features are absent.

In contrast, ResNet18 embeddings (\texttt{ResNet18\_emb}), despite being trained directly on the task and the image representations selected via cross-validation, do not systematically improve the mean RMSE of the GLM or XGB frameworks, and their impact on variability is limited. Overall, while image embeddings do not outperform the best-performing specification that includes OSM–CORINE features (regularized GLM with \texttt{+ lat\_long + osm\_r5}), Nomic embeddings provide clear gains in both accuracy and stability in settings where structured environmental information is absent. This suggests that pretrained vision transformers can serve as effective proxies for geographic context when explicit spatial features cannot be constructed.

To illustrate the type of relationships captured by the specification with the lowest mean test RMSE and variance (regularized GLM with latitude, longitude, and 5 km environmental features), Table \ref{tab:glm_example_coef} reports the non-zero coefficients for a representative fold (fold 3), where the best ElasticNet hyperparameters were $\eta=0.2$ and $\alpha = 1.0$ in Equation(\ref{eq:ELasticNet}) (Lasso penalty). 

\begin{table}[H]
\centering
\begin{tabular}{lr}
\toprule \hline
\textbf{Variable} & \textbf{Coefficient} \\
\hline \midrule
Intercept                          & -2.0460  \\
$\texttt{bm\_mean}$                &  0.1119  \\
$\texttt{road\_len\_km\_per\_km2\_r5000}$ &  0.0654 \\
$\texttt{coverage\_TPL\_prop}$     & -0.0215 \\
$\texttt{has\_healthcare\_r5000}$  &  0.0167  \\
$\texttt{lat}$                     &  0.0100 \\
$\texttt{power\_median}$           &  0.0087 \\
$\texttt{parking\_count\_per\_km2\_r5000}$ & -0.0070 \\
\midrule
\multicolumn{2}{l}{Non-selected variables (coefficients shrunk to 0 by ElasticNet).} \\
\bottomrule
\end{tabular}
\caption{Non-zero coefficients for the regularized GLM with latitude and 5 km environmental features in fold 3. All predictor variables were standardised prior to estimation.}
\label{tab:glm_example_coef}
\end{table}

In this model, a higher average bonus–malus level (\texttt{bm\_mean}) and greater road density within 5 km are associated with an increased frequency of claims, while a higher share of TPL coverage is associated with a lower frequency. The presence of healthcare facilities, as well as latitude and median engine power, also contribute positively, while more parking amenities are associated with a reduction in expected claims.

For this fold, the model achieves a test RMSE of 6.13 and a mean predicted claim count of 37.54 versus an observed mean of 37.10, illustrating that a sparse set of covariates can provide both competitive predictive accuracy and an interpretable link between claim frequency and local traffic environment.

\section{Robustness Analysis}
\label{sec:RobustnessAnalysis}
In this section, we assess the robustness of our findings to the way data are partitioned into training and test sets. Specifically, we examine whether the conclusions reported above remain stable when varying the fold assignments used in the evaluation procedure described in Section~\ref{sec:Methodology}. To this end, we repeated the full set of experiments under four additional split configurations, resulting in five distinct sets of six-fold partitions.

Across these alternative splits, the exact combination of predictors that achieve the lowest mean test RMSE varies for both the GLM family (with and without regularization) and the XGB models. This indicates that model selection at a fine level is sensitive to how geographic zones are allocated across folds. Nevertheless, a consistent pattern emerges, in all split configurations, the inclusion of OSM-derived environmental features leads to improvements over models relying only on traditional actuarial variables. In particular, across all five split configurations, the model achieving the lowest average test RMSE across the six folds is a regularized GLM in which environmental features constructed within a 5 km radius are included among the predictors. This suggests that while the optimal predictor set is not invariant to the specific data split, the importance of geographic context is robust to the choice of fold partition.

Table~\ref{tab:TableRAGLM} summarizes the robustness analysis across the five experimental splits for the GLM family. On average, the configuration achieving both the lowest mean test RMSE across folds and the lowest variability in test RMSE is the regularized GLM that includes latitude-longitude coordinates and environmental features of the OSM constructed within a 5~km radius (\texttt{+ lat\_long + osm\_r5}). This result is consistent with the findings reported in Section~\ref{sec:Results}.

\begin{table}[H]
\begin{adjustbox}{max width=\textwidth}
\begin{tabular}{@{}llccccc@{}}
\toprule \hline
\multirow{2}{*}{}              & \multirow{2}{*}{\textbf{Model}}                                 & \multicolumn{1}{c}{\multirow{3}{*}{\textbf{\begin{tabular}[c]{@{}c@{}}Avg. fold-mean \\ test RMSE \\ (5 experiments)\end{tabular}}}} & \multicolumn{1}{c}{\multirow{3}{*}{\textbf{\begin{tabular}[c]{@{}c@{}}Avg. fold-std \\ test RMSE \\ (5 experiments)\end{tabular}}}} & \multicolumn{1}{l}{\multirow{2}{*}{}}                                                                                    & \multicolumn{1}{c}{\multirow{3}{*}{\textbf{\begin{tabular}[c]{@{}c@{}}Avg. fold-mean \\ test RMSE\\ (5 experiments)\end{tabular}}}} & \multicolumn{1}{c}{\multirow{3}{*}{\textbf{\begin{tabular}[c]{@{}c@{}}Avg. fold-std \\ test RMSE\\ (5 experiments)\end{tabular}}}} \\
\\
                               &                                                                 & \multicolumn{1}{c}{}                                                                                                              & \multicolumn{1}{c}{}                                                                                                             & \multicolumn{1}{l}{}                                                                                                     & \multicolumn{1}{c}{}                                                                                                             & \multicolumn{1}{c}{}                                                                                                            \\
                               \hline \midrule
\multirow{24}{*}{\textbf{GLM}} & Only tabular feature (not geography-related)                    & 8.4725                                                                                                                            & 2.5960                                                                                                                           & \multirow{24}{*}{\textbf{\begin{tabular}[c]{@{}c@{}}R\\ e\\ g\\ u\\ l\\ a\\ r\\ i\\ z\\ e\\ d\\ \\ \\ GLM\end{tabular}}} & 8.6832                                                                                                                           & 2.6910                                                                                                                          \\
                               & \texttt{+ lat\_long}                           & 8.4345                                                                                                                            & 2.0071                                                                                                                           &                                                                                                                          & 8.4222                                                                                                                           & 2.3639                                                                                                                          \\
                               & \texttt{+ postcode2}                           & 11.2980                                                                                                                           & 5.4623                                                                                                                           &                                                                                                                          & 9.0487                                                                                                                           & 3.7511                                                                                                                          \\
                               & \texttt{+ lat\_long + postcode\_2}             & 11.1144                                                                                                                           & 5.3909                                                                                                                           &                                                                                                                          & 8.3377                                                                                                                           & 2.2644                                                                                                                          \\
                               \cline{2-4}
\cline{6-7}
                               & \texttt{+ osm\_r0.5}                           & 9.3277                                                                                                                            & 2.2832                                                                                                                           &                                                                                                                          & 9.1743                                                                                                                           & 2.2846                                                                                                                          \\
                               & \texttt{+ osm\_r1}                             & 8.5421                                                                                                                            & 1.8187                                                                                                                           &                                                                                                                          & 8.6127                                                                                                                           & 1.7994                                                                                                                          \\
                               & \texttt{+ osm\_r3}                             & 9.5132                                                                                                                            & 3.5069                                                                                                                           &                                                                                                                          & 8.1576                                                                                                                           & 2.1214                                                                                                                          \\
                               & \texttt{+ osm\_r5}                             & 8.1451                                                                                                                            & 1.9082                                                                                                                           &                                                                                                                          & 7.8205                                                                                                                           & 1.6207                                                                                                                          \\
                               & \texttt{+ osm\_rALL}                           & 11.8759                                                                                                                           & 5.4864                                                                                                                           &                                                                                                                          & 8.5958                                                                                                                           & 2.1518                                                                                                                          \\
                                                              \cline{2-4}
\cline{6-7}
                               & \texttt{+ postcode\_2 + osm\_r0.5}             & 12.1943                                                                                                                           & 4.7193                                                                                                                           &                                                                                                                          & 9.8287                                                                                                                           & 2.6871                                                                                                                          \\
                               & \texttt{+ postcode\_2 + osm\_r1}               & 11.2954                                                                                                                           & 4.3082                                                                                                                           &                                                                                                                          & 8.0721                                                                                                                           & 1.3777                                                                                                                          \\
                               & \texttt{+ postcode\_2 + osm\_r3}              & 11.8348                                                                                                                           & 5.2023                                                                                                                           &                                                                                                                          & 7.9870                                                                                                                           & 1.8502                                                                                                                          \\
                               & \texttt{+ postcode\_2 + osm\_r5}               & 11.9266                                                                                                                           & 4.9663                                                                                                                           &                                                                                                                          & 7.5412                                                                                                                           & 1.1597                                                                                                                          \\
                               & \texttt{+ postcode\_2 + osm\_rALL}             & 20.8211                                                                                                                           & 14.1991                                                                                                                          &                                                                                                                          & 8.4486                                                                                                                           & 2.1774                                                                                                                          \\
                                                              \cline{2-4}
\cline{6-7}
                               & \texttt{+ lat\_long + osm\_r0.5}               & 9.3589                                                                                                                            & 2.0165                                                                                                                           &                                                                                                                          & 8.8358                                                                                                                           & 1.8137                                                                                                                          \\
                               & \texttt{+ lat\_long + osm\_r1}                 & 8.4258                                                                                                                            & 1.3718                                                                                                                           &                                                                                                                          & 8.3288                                                                                                                           & 1.6237                                                                                                                          \\
                               & \texttt{+ lat\_long + osm\_r3}                 & 9.1291                                                                                                                            & 2.9214                                                                                                                           &                                                                                                                          & 8.3436                                                                                                                           & 2.1576                                                                                                                          \\
                               & \texttt{+ lat\_long + osm\_r5}                 & 7.9486                                                                                                                            & 1.7185                                                                                                                           &                                                                                                                          & \textbf{7.3895}                                                                                                                  & \textbf{1.0099}                                                                                                                 \\
                               & \texttt{+ lat\_long + osm\_rALL}               & 12.0162                                                                                                                           & 5.6423                                                                                                                           &                                                                                                                          & 8.7218                                                                                                                           & 2.1546                                                                                                                          \\
                                                              \cline{2-4}
\cline{6-7}
                               & \texttt{+ lat\_long + postcode\_2 + osm\_r0.5} & 12.2358                                                                                                                           & 5.3979                                                                                                                           &                                                                                                                          & 8.9372                                                                                                                           & 1.9985                                                                                                                          \\
                               & \texttt{+ lat\_long + postcode\_2 + osm\_r1}   & 11.5129                                                                                                                           & 4.3460                                                                                                                           &                                                                                                                          & 8.0721                                                                                                                           & 1.3777                                                                                                                          \\
                               & \texttt{+ lat\_long + postcode\_2 + osm\_r3}   & 12.1775                                                                                                                           & 4.8428                                                                                                                           &                                                                                                                          & 7.9306                                                                                                                           & 1.7331                                                                                                                          \\
                               & \texttt{+ lat\_long + postcode\_2 + osm\_r5}   & 12.4041                                                                                                                           & 5.0057                                                                                                                           &                                                                                                                          & 7.4752                                                                                                                           & 1.2428                                                                                                                          \\
                               & \texttt{+ lat\_long + postcode\_2 + osm\_rALL} & 19.5148                                                                                                                           & 12.3666                                                                                                                          &                                                                                                                          & 8.3354                                                                                                                           & 1.8324       \\\bottomrule                                                                                                                  
\end{tabular}
\end{adjustbox}
\caption{Robustness analysis over five repeated experiments with different random seeds. In each experiment, test RMSE is averaged across six folds.
The table reports the mean of these per-experiment mean RMSEs and the mean of the corresponding fold-level RMSE standard deviations across the GLM model family.}
\label{tab:TableRAGLM}
\end{table}

Beyond the previous fact, the table also shows that incorporating partial neighborhood information can already yield meaningful improvements. In particular, regularized GLMs that include OSM features constructed at 1~km or 3~km radius reduce both prediction error and variability relative to models that rely only on traditional actuarial variables combined with coordinates and postcode indicators (i.e., \texttt{+ lat\_long + postcode\_2}). A similar pattern is observed when environmental features at a 3~km radius are included in the absence of other geographic covariates, leading to improved performance over purely non-geographic specifications. In contrast, very local information extracted at a 0.5~km radius does not lead to systematic performance gains on average in all experiments.

Regarding XGB models, the robustness results reported in Table~\ref{tab:TableRAXGB} show that, on average, in all experiments, the inclusion of OSM-derived environmental features is associated with lower test RMSEs and reduced variability compared to specifications relying only on traditional actuarial variables or basic location identifiers. However, unlike the GLM family, the optimal spatial scale is not invariant across experimental splits: while the specific radius that yields the lowest error varies, in the majority of experiments, the lowest average test RMSE is obtained when environmental features constructed within a 5-~km radius are included. Moreover, in contrast to linear models, XGB specifications benefit more consistently from partial or multiscale geographic information, as features constructed at smaller radii (0.5 to 3~km) or aggregated across all radii are, on average, associated with improvements in both predictive accuracy and stability relative to XGB models without geographic information or those that only include coordinates and/or postcode indicators (\texttt{postcode\_2}).

Turning to the convolutional neural network (CNN) models, the robustness analysis leads to conclusions consistent with those reported in Section~\ref{sec:Results}. As shown in Table~\ref{tab:RA_CNN_nongeo}, raw images alone provide limited predictive power for claim frequency, resulting in relatively high errors and unstable performance. However, augmenting the CNN with traditional actuarial variables substantially reduces both the test RMSE and, in most cases, its variability, indicating that image-based information is more informative when combined with standard risk factors. Furthermore, following the same scale-selection strategy as in the main results—namely choosing the image scale and OSM radius that minimize cross-validation error. Table~\ref{tab:RA_CNN_geo} shows that the inclusion of additional geographic information (such as coordinates, postcode region, or OSM-derived environmental features) in the majority of experiments further improves the average predictive accuracy and stability (with the exception of the first experiment). Despite this, CNN based specifications remain outperformed by simpler models, across all robustness experiments, the lowest test errors are consistently achieved by regularized GLM specifications, showing that increased model complexity does not translate into superior performance in this setting.

\begin{table}[H]
\centering
\begin{adjustbox}{width=0.8\textwidth}
\begin{tabular}{@{}llcc@{}}
\toprule \hline
\multirow{2}{*}{}              & \multirow{2}{*}{\textbf{Model}}                                 & \multirow{2}{*}{\textbf{\begin{tabular}[c]{@{}c@{}}Avg. fold-mean test RMSE \\ (5 experiments)\end{tabular}}} & \multirow{2}{*}{\textbf{\begin{tabular}[c]{@{}c@{}}Avg. fold-std test RMSE \\ (5 experiments)\end{tabular}}} \\
                               &                                                                 &                                                                                                               &                                                                                                              \\
                               \hline \midrule
\multirow{24}{*}{\textbf{XGB}} & Only tabular feature (not geography-related)                    & 9.2201                                                                                                        & 3.5804                                                                                                       \\
                               & \texttt{+ lat\_long}                           & 9.2755                                                                                                        & 4.0177                                                                                                       \\
                               & \texttt{+ postcode2}                           & 9.2932                                                                                                        & 3.6201                                                                                                       \\
                               & \texttt{+ lat\_long + postcode\_2}             & 9.1572                                                                                                        & 3.9729                                                                                                       \\
                               & \texttt{+ osm\_r0.5}                           & 9.1314                                                                                                        & 2.9838                                                                                                       \\
                               & \texttt{+ osm\_r1}                             & 8.6256                                                                                                        & 2.3347                                                                                                       \\
                               & \texttt{+ osm\_r3}                             & 8.2906                                                                                                        & 2.3345                                                                                                       \\
                               & \texttt{+ osm\_r5}                             & 8.2259                                                                                                        & 2.2931                                                                                                       \\
                               & \texttt{+ osm\_rALL}                           & 8.3436                                                                                                        & 2.3035                                                                                                       \\
                               & \texttt{+ postcode\_2 + osm\_r0.5}             & 9.0566                                                                                                        & 3.2824                                                                                                       \\
                               & \texttt{+ postcode\_2 + osm\_r1}               & 8.5929                                                                                                        & 2.5727                                                                                                       \\
                               & \texttt{+ postcode\_2 + osm\_r3}              & 8.3595                                                                                                        & 2.6631                                                                                                       \\
                               & \texttt{+ postcode\_2 + osm\_r5}               & \textbf{8.1389}                                                                                               & 2.4579                                                                                                       \\
                               & \texttt{+ postcode\_2 + osm\_rALL}             & 8.2977                                                                                                        & 2.3696                                                                                                       \\
                               & \texttt{+ lat\_long + osm\_r0.5}               & 9.0295                                                                                                        & 3.3993                                                                                                       \\
                               & \texttt{+ lat\_long + osm\_r1}                 & 8.6256                                                                                                        & 2.3347                                                                                                       \\
                               & \texttt{+ lat\_long + osm\_r3}                 & 8.2134                                                                                                        & 2.3153                                                                                                       \\
                               & \texttt{+ lat\_long + osm\_r5}                 & 8.2414                                                                                                        & 2.4566                                                                                                       \\
                               & \texttt{+ lat\_long + osm\_rALL}               & 8.1810                                                                                                        & 2.3380                                                                                                       \\
                               & \texttt{+ lat\_long + postcode\_2 + osm\_r0.5} & 9.0566                                                                                                        & 3.2824                                                                                                       \\
                               & \texttt{+ lat\_long + postcode\_2 + osm\_r1}   & 8.7220                                                                                                        & 2.9811                                                                                                       \\
                               & \texttt{+ lat\_long + postcode\_2 + osm\_r3}   & 8.2499                                                                                                        & 2.7270                                                                                                       \\
                               & \texttt{+ lat\_long + postcode\_2 + osm\_r5}   & 8.1714                                                                                                        & 2.5541                                                                                                       \\
                               & \texttt{+ lat\_long + postcode\_2 + osm\_rALL} & 8.3775                                                                                                        & 2.5299                        \\\bottomrule                                                                              
\end{tabular}
\end{adjustbox}
\caption{Robustness analysis of the XGB model over five repeated experiments with different random seeds. In each experiment, test RMSE is averaged across six folds. The table reports the mean of the per-experiment mean RMSEs and the mean of the corresponding fold-level RMSE standard deviations.}
\label{tab:TableRAXGB}
\end{table}
\begin{table}[htb!]
\centering
\begin{subtable}[t]{0.4\textwidth}
\centering
\begin{adjustbox}{width=\textwidth}
\begin{tabular}{@{}clcc@{}}
\toprule
\multicolumn{1}{l}{\multirow{3}{*}{}} &
\multicolumn{1}{c}{\multirow{3}{*}{\textbf{\begin{tabular}[c]{@{}c@{}}CNN\\ Model \\ Family\end{tabular}}}} &
\multirow{3}{*}{\textbf{\begin{tabular}[c]{@{}c@{}}Avg. fold-mean \\ test RMSE \\ (5 experiments)\end{tabular}}} &
\multirow{3}{*}{\textbf{\begin{tabular}[c]{@{}c@{}}Avg. fold-std\\ test RMSE \\ (5 experiments)\end{tabular}}} \\
\multicolumn{1}{l}{} & \multicolumn{1}{c}{} & & \\\\
\midrule
\multirow{4}{*}{\begin{tabular}[c]{@{}c@{}}Only\\ Images\end{tabular}} &
\texttt{+ img\_0.5} & 14.5680 & 4.5733 \\
& \texttt{+ img\_1}   & 13.6625 & 4.9268 \\
& \texttt{+ img\_3}   & 12.2076 & 4.5008 \\
& \texttt{+ img\_ALL} & 14.0908 & 7.9495 \\
\midrule
\multirow{4}{*}{Multimodal} &
\texttt{+ img\_0.5} & 12.7184 & 5.6815 \\
& \texttt{+ img\_1}   & 10.8343 & 3.5124 \\
& \texttt{+ img\_3}   & 10.1565 & 2.2264 \\
& \texttt{+ img\_ALL} & \textbf{9.5525}  & 3.3185 \\
\bottomrule
\end{tabular}
\end{adjustbox}

\caption{Robustness analysis of the CNN models (ResNet18), using images only or concatenated with non-geographic tabular predictors. Image tiles with apothem length $r$ km are denoted by \texttt{img\_r}.}
\label{tab:RA_CNN_nongeo}
\end{subtable}
\hfill
\begin{subtable}[t]{0.57\textwidth}
\centering
\begin{adjustbox}{width=\textwidth}
\begin{tabular}{@{}clcc@{}}
\toprule
\multirow{3}{*}{\textbf{\begin{tabular}[c]{@{}c@{}}Model\end{tabular}}} &
&
\multirow{3}{*}{\textbf{\begin{tabular}[c]{@{}c@{}}Avg. fold-mean \\ test RMSE \\ (5 experiments)\end{tabular}}} &
\multirow{3}{*}{\textbf{\begin{tabular}[c]{@{}c@{}}Avg. fold-std \\ test RMSE \\ (5 experiments)\end{tabular}}} \\
&&&\\\\
\midrule
\multirow{7}{*}{\textbf{\begin{tabular}[c]{@{}c@{}}CNN\\ with \texttt{img\_3 +}\\ geographic \\ features\end{tabular}}} &
\texttt{+ lat\_long} & 9.4748 &	2.1435\\
& \texttt{+ postcode\_2} & 12.5501 &	6.4626\\
& \texttt{+ postcode\_2 + lat\_long} & 11.4664 & 4.5172 \\
& \texttt{+ osm\_r*} & \textbf{9.3957} &	2.4757 \\
& \texttt{+ osm\_r* + lat\_long} & 9.5695 &	2.8402 \\
& \texttt{+ osm\_r* + postcode\_2} & 11.0664 &	4.1942 \\
& \texttt{+ osm\_r* + postcode\_2 + lat\_long} &  11.3775 &	3.5920\\
\bottomrule
\end{tabular}
\end{adjustbox}

\caption{Robustness analysis when adding geographic information (coordinates, postcode region, or OSM features). Here $r^*$ denotes the radius yielding the lowest mean 5-fold CV error in the MLP model.}
\label{tab:RA_CNN_geo}
\end{subtable}

\caption{Robustness analysis results for CNN variants across five repeated experiments with different random seeds. In each experiment, test RMSE is averaged across six folds.}
\label{tab:RA_CNN_two_tables}
\end{table}

Finally, the robustness analysis provides additional insight into the role of image embeddings. Across repeated experiments, embedding based specifications often achieve lower mean test RMSE than their corresponding baseline models, particularly when explicit environmental OSM features are not included. In one experiment, an embedding based model even outperforms the best regularized GLM with OSM features. Moreover, although the main result section emphasizes improvements mainly within the GLM family, the robustness analysis shows that embeddings can also improve performance in some XGB specifications. In addition, improvements are not limited to Nomic embeddings: in several experiments, ResNet18 embeddings also provide gains relative to their corresponding baselines. In general, these results suggest that image embeddings can provide a useful additional signal when structured geographic information is limited or absent.

\section{Conclusions}
\label{sec:Conclusions}
This study examined how geographic information can be constructed and incorporated into actuarial models for MTPL claim frequency prediction when only limited location identifiers are available. Using the BeMTPL97 dataset, we focused on zonal-level modeling as a practical and interpretable framework to assess the contribution of geography beyond traditional actuarial variables. Model performance is evaluated in postcodes not observed during training, so all reported test results reflect out-of-sample prediction across geographic zones rather than within-zone fitting. Our analysis was guided by three research questions regarding the construction of geographic information from alternative data sources, its predictive value, and the sensitivity of the results to the spatial scale.

Firstly, we show that meaningful geographic information can be constructed by combining openly available spatial data sources with historical imagery aligned to the insurance exposure period. In particular, environmental indicators derived from OpenStreetMap form 2014 and CORINE Land Cover 2000, together with orthoimagery from 1995 provided by the Belgian National Geographic Institute (for academic purposes), allow the representation of local built environment characteristics despite the absence of detailed spatial variables in the raw insurance data. Moreover, this opens the door to the use of similar alternative geographic data in modern insurance settings, where such information can be obtained either from open source platforms or through commercial providers, allowing insurers to incorporate these features and assess the associated predictive gains. 

Second, the empirical results show that augmenting traditional actuarial predictors with explicitly constructed environmental features derived from OSM and CORINE leads to consistent predictive gains over models relying solely on non-geographic tabular variables. For linear specifications, including both the standard GLM and its regularized counterpart, the combination of coordinates and environmental features extracted at a 5 km neighborhood scale yields the strongest and most stable improvements in the mean test RMSEs  on average across the experiments, reflecting a balance between spatial localization and contextual richness. For XGB models, the best average performance is achieved when the postcode region is combined with environmental features on a 5km scale. However, in contrast to the GLM family, the optimal XGB specification is not invariant to the data split. In some experiments, the lowest errors are obtained when using environmental features aggregated across all radius or at a 3 km scale. This variability indicates that the non-linear structure of tree-based ensembles can exploit multiscale geographic information without suffering from the redundancy issues observed in linear models. In general, these findings highlight that both geographic positioning and environmental context are complementary and that their optimal integration depends on the modeling framework.

In addition, our findings indicate that image embeddings offer a complementary source of geographic information. When explicit environmental predictors such as OSM-CORINE features are available, image embeddings rarely improve predictive accuracy in all experiments, suggesting that handcrafted environmental indicators already capture the spatial structure most relevant to MTPL claim frequency. However, an important contribution emerges when such structured environmental variables are absent. In these settings, the Nomic-v1.5 vision-transformer embeddings, despite not being trained for insurance applications, consistently improve both predictive accuracy and cross-fold stability relative to baseline GLMs, and in some experiments also relative to XGB models. This contrasts with ResNet18 embeddings trained directly on the target variable and features related to the task, which do not yield consistent comparable gains between experiments. These results suggest that modern pretrained vision transformers can extract built environment signals that are otherwise unavailable in tabular form.

With respect to spatial scale, the results show that the predictive contribution of geographic information is highly sensitive to the neighborhood size at which features are constructed and that on average across the experiments, the 5km scale is uniformly optimal across modeling frameworks. In all experiments, for linear models, including both standard and regularized GLM, environmental features extracted at an intermediate scale of 5 km provide the most stable and accurate predictions, suggesting a balance between local specificity and contextual richness. In contrast, XGB models show greater flexibility with respect to scale. Although their lowest average errors across experiments are also achieved at the 5 km scale, in several experiments comparable or lower errors arise when environmental features are constructed using all radii or a 3 km neighborhood. In particular, smaller neighborhoods (1 km or 3 km) often yield improvements relative to models without environmental information, even when they do not achieve the lowest overall errors. This suggests that a partial geographic context still provides informative signals beyond traditional actuarial and location variables. Overall, these findings highlight that spatial scale selection is a substantive modeling decision rather than a technical detail, and that its optimal choice depends on both the nature of the geographic information and the modeling framework employ.

This study contributes to the growing literature on the use of alternative data in actuarial modeling by examining how geographic information can be constructed and incorporated into MTPL claim frequency prediction. Focusing on zonal-level modeling, we assess whether the geographic context, derived from coordinates, environmental indicators, images, and image embeddings, provides incremental predictive value beyond traditional actuarial variables. 

Across experiments, models based on raw imagery using convolutional neural networks do not outperform simpler GLM or tree-based specifications, indicating that increased model complexity alone does not ensure superior predictive performance in this setting. At the same time, robustness analysis highlights that while the exact set of predictors producing the lowest error can vary with the data split, the central findings remain stable.

These results underscore the importance of robustness checks in empirical actuarial studies, as conclusions drawn from a single data split may overstate the generality of specific model configurations. From a practical perspective, the findings support a modeling strategy that combines global geographic context at the zone level with individual-level actuarial characteristics, offering a principled way to enrich risk classification and pricing while maintaining interpretability. To our knowledge, this is the first study to systematically evaluate the predictive contribution of constructed geographic information, including environmental indicators and image-based representations, in modeling the frequency of MTPL claims. 

Although the use of a single historical dataset constrains the generalization of numerical results, it also reflects the practical reality that insurance data with detailed geographic references are rarely accessible for research. In this sense, our findings illustrate how different forms of geographic information behave when evaluated in previously unseen geographic zones. We hope this work encourages further investigation of alternative geographic data sources in more recent insurance portfolios, where high resolution satellite imagery or street-level mapping products (e.g., Google Earth or Google Maps) may be available and can be evaluated within a comparable out-of-sample framework.

\section{Acknowledgment}
The authors gratefully acknowledge the support of the Natural Sciences and Engineering Research Council of Canada (NSERC) [Discovery Grants RGPIN-2020-07114 and RGPIN-2019-06586]. This work was funded, in part, with funding from the Canada Research Chair Program [CRC-2024-00192] and was enabled, in part, by the support provided by Compute Ontario (\url{https://www.computeontario.ca}) and the Digital Research Alliance of Canada (\url{https://alliancecan.ca}).

\section{CRediT authorship contribution statement}
\textbf{Sherly Alfonso-Sánchez:}  Conceptualization, Data curation, Formal analysis, Methodology, Software, Writing – original draft, Writing – review \& editing. \textbf{Kristina G. Stankova:}  Conceptualization, Formal analysis, Funding acquisition, Methodology, Supervision, Writing – review \& editing. \textbf{Cristián Bravo:}  Conceptualization, Formal analysis, Funding acquisition, Methodology, Supervision, Writing – review \& editing.

\section{Declaration of competing interest}
We declare that there are no known conflicts of interest associated with this publication.

\section{Declaration of generative AI use}
During the preparation of this work, Sherly Alfonso-Sánchez used ChatGPT to refine the language, clarity, and structure of the manuscript. After using this tool, the authors reviewed and edited the content as needed and assume full responsibility for the content of the published article.


\begin{thebibliography}{55}
\providecommand{\natexlab}[1]{#1}
\providecommand{\url}[1]{\texttt{#1}}
\expandafter\ifx\csname urlstyle\endcsname\relax
  \providecommand{\doi}[1]{doi: #1}\else
  \providecommand{\doi}{doi: \begingroup \urlstyle{rm}\Url}\fi

\bibitem[SOA(2015)]{SOA2015LTC}
Long term care intercompany study, January 2015.
\newblock URL \url{https://www.soa.org/4a6a75/globalassets/assets/files/resources/experience-studies/2019/ltc-intercompany-study.pdf}.
\newblock Accessed: 2025-02-14.

\bibitem[Arlot and Celisse(2010)]{ArlotSylvain2010Asoc}
Arlot, S. and Celisse, A.
\newblock A survey of cross-validation procedures for model selection.
\newblock \emph{Statistics surveys}, 4\penalty0 (none):\penalty0 40--79, 2010.
\newblock ISSN 1935-7516.
\newblock \doi{10.1214/09-SS054}.

\bibitem[Asabere et~al.(2024)Asabere, Asare, Lawson, Balde, Duodu, Tsoekeku, Afriyie, and Ganiu]{asabere2024geo}
Asabere, N.~Y., Asare, I.~O., Lawson, G., Balde, F., Duodu, N.~Y., Tsoekeku, G., Afriyie, P.~O., and Ganiu, A. R.~A.
\newblock Geo-insurance: Improving big data challenges in the context of insurance services using a geographical information system (gis).
\newblock \emph{Human Behavior and Emerging Technologies}, 2024\penalty0 (1):\penalty0 9015012, 2024.
\newblock \doi{10.1155/2024/9015012}.

\bibitem[Ayuso et~al.(2019)Ayuso, Guillen, and Nielsen]{ayuso2019improving}
Ayuso, M., Guillen, M., and Nielsen, J.~P.
\newblock Improving automobile insurance ratemaking using telematics: incorporating mileage and driver behaviour data.
\newblock \emph{Transportation}, 46\penalty0 (3):\penalty0 735--752, 2019.
\newblock \doi{10.1007/s11116-018-9890-7}.

\bibitem[Benedek and Nagy(2023)]{benedek2023traditional}
Benedek, B. and Nagy, B.~Z.
\newblock Traditional versus ai-based fraud detection: cost efficiency in the field of automobile insurance.
\newblock \emph{Financial and Economic Review}, 22\penalty0 (2):\penalty0 77--98, 2023.
\newblock \doi{10.33893/FER.22.2.77}.

\bibitem[Bengio et~al.(2017)Bengio, Goodfellow, Courville, et~al.]{bengio2017deep}
Bengio, Y., Goodfellow, I., Courville, A., et~al.
\newblock \emph{Deep learning}, volume~1.
\newblock MIT press Cambridge, MA, USA, 2017.

\bibitem[Bhattacharya et~al.(2025)Bhattacharya, Castignani, Masello, and Sheehan]{bhattacharya2025ai}
Bhattacharya, S., Castignani, G., Masello, L., and Sheehan, B.
\newblock Ai revolution in insurance: bridging research and reality.
\newblock \emph{Frontiers in Artificial Intelligence}, 8:\penalty0 1568266, 2025.
\newblock \doi{10.3389/frai.2025.1568266}.

\bibitem[Blier-Wong et~al.(2022)Blier-Wong, Cossette, Lamontagne, and Marceau]{blier2022geographic}
Blier-Wong, C., Cossette, H., Lamontagne, L., and Marceau, E.
\newblock Geographic ratemaking with spatial embeddings.
\newblock \emph{ASTIN Bulletin: The Journal of the IAA}, 52\penalty0 (1):\penalty0 1--31, 2022.
\newblock \doi{10.1017/asb.2021.25}.

\bibitem[Blier-Wong et~al.(2024)Blier-Wong, Lamontagne, and Marceau]{blier2024representation}
Blier-Wong, C., Lamontagne, L., and Marceau, E.
\newblock A representation-learning approach for insurance pricing with images.
\newblock \emph{ASTIN Bulletin: The Journal of the IAA}, 54\penalty0 (2):\penalty0 280--309, 2024.
\newblock \doi{10.1017/asb.2024.9}.

\bibitem[Burka et~al.(2021)Burka, Kov{\'a}cs, and Szepesv{\'a}ry]{burka2021modelling}
Burka, D., Kov{\'a}cs, L., and Szepesv{\'a}ry, L.
\newblock Modelling mtpl insurance claim events: Can machine learning methods overperform the traditional glm approach?
\newblock \emph{Hungarian Statistical Review}, 4\penalty0 (2), 2021.
\newblock \doi{10.35618/hsr2021.02.en034}.

\bibitem[{Cartesius / National Geographic Institute (NGI Belgium)}()]{CartesiusPortal}
{Cartesius / National Geographic Institute (NGI Belgium)}.
\newblock Belgian national geospatial data portal.
\newblock \url{https://www.cartesius.be}.
\newblock Accessed 04.12.2025.

\bibitem[Chen and Guestrin(2016)]{chen2016xgboost}
Chen, T. and Guestrin, C.
\newblock Xgboost: A scalable tree boosting system.
\newblock In \emph{Proceedings of the 22nd acm sigkdd international conference on knowledge discovery and data mining}, pages 785--794, 2016.
\newblock \doi{10.1145/2939672.2939785}.

\bibitem[Clemente et~al.(2023)Clemente, Guerreiro, and Bravo]{clemente2023modelling}
Clemente, C., Guerreiro, G.~R., and Bravo, J.~M.
\newblock Modelling motor insurance claim frequency and severity using gradient boosting.
\newblock \emph{Risks}, 11\penalty0 (9):\penalty0 163, 2023.
\newblock \doi{10.3390/risks11090163}.

\bibitem[{Copernicus Land Monitoring Service / European Environment Agency}(2020)]{CLC2000}
{Copernicus Land Monitoring Service / European Environment Agency}.
\newblock {CORINE Land Cover 2000 (CLC 2000)}.
\newblock European Union’s Copernicus Land Monitoring Service, 2020.
\newblock URL \url{https://land.copernicus.eu/en/products/corine-land-cover/clc-2000}.
\newblock Accessed 02.12.2025.

\bibitem[Ding et~al.(2025)Ding, Ruan, Wang, and Liu]{DING202551}
Ding, N., Ruan, X., Wang, H., and Liu, Y.
\newblock Automobile insurance fraud detection based on pso-xgboost model and interpretable machine learning method.
\newblock \emph{Insurance: Mathematics and Economics}, 120:\penalty0 51--60, 2025.
\newblock ISSN 0167-6687.
\newblock \doi{https://doi.org/10.1016/j.insmatheco.2024.11.006}.
\newblock URL \url{https://www.sciencedirect.com/science/article/pii/S0167668724001112}.

\bibitem[Dong and Quan(2025)]{DONG202517}
Dong, P. and Quan, Z.
\newblock Automated machine learning in insurance.
\newblock \emph{Insurance: Mathematics and Economics}, 120:\penalty0 17--41, 2025.
\newblock ISSN 0167-6687.
\newblock \doi{https://doi.org/10.1016/j.insmatheco.2024.10.002}.
\newblock URL \url{https://www.sciencedirect.com/science/article/pii/S0167668724001057}.

\bibitem[Dubey et~al.(2018)Dubey, Parida, Birajdar, Prajapati, and Rane]{dubey2018smart}
Dubey, A., Parida, T., Birajdar, A., Prajapati, A.~K., and Rane, S.
\newblock Smart underwriting system: An intelligent decision support system for insurance approval \& risk assessment.
\newblock In \emph{2018 3rd International Conference for Convergence in Technology (I2CT)}, pages 1--6. IEEE, 2018.
\newblock \doi{10.1109/I2CT.2018.8529792}.

\bibitem[Dutang and Charpentier(2024)]{CASdatasets2024}
Dutang, C. and Charpentier, A.
\newblock \emph{CASdatasets: Insurance datasets}, 2024.
\newblock R package version 1.2-0.

\bibitem[Dutang et~al.(2024)Dutang, Charpentier, and Gallic]{dutang2024insurance}
Dutang, C., Charpentier, A., and Gallic, E.
\newblock Insurance dataset.
\newblock 2024.

\bibitem[{Environmental Systems Research Institute (Esri)}(2022)]{esri_belgium_postcodes}
{Environmental Systems Research Institute (Esri)}.
\newblock Belgium postcode boundaries.
\newblock \url{https://www.arcgis.com/home/item.html?id=e385aeef974a4aea8ae7fb1b0efc1341}, 2022.
\newblock GIS dataset accessed January 2026.

\bibitem[Fouad et~al.(2023)Fouad, Malawany, Osman, Amer, Abdulkhalek, and Eldin]{fouad2023automated}
Fouad, M.~M., Malawany, K., Osman, A.~G., Amer, H.~M., Abdulkhalek, A.~M., and Eldin, A.~B.
\newblock Automated vehicle inspection model using a deep learning approach.
\newblock \emph{Journal of Ambient Intelligence and Humanized Computing}, 14\penalty0 (10):\penalty0 13971--13979, 2023.
\newblock \doi{10.1007/s12652-022-04105-3}.

\bibitem[Gao et~al.(2022)Gao, Wang, and W{\"u}thrich]{gao2022boosting}
Gao, G., Wang, H., and W{\"u}thrich, M.~V.
\newblock Boosting poisson regression models with telematics car driving data.
\newblock \emph{Machine Learning}, 111\penalty0 (1):\penalty0 243--272, 2022.
\newblock \doi{10.2139/ssrn.3596034}.

\bibitem[Gupta et~al.(2022)Gupta, Ghardallou, Pandey, and Sahu]{gupta2022artificial}
Gupta, S., Ghardallou, W., Pandey, D.~K., and Sahu, G.~P.
\newblock Artificial intelligence adoption in the insurance industry: Evidence using the technology--organization--environment framework.
\newblock \emph{Research in International Business and Finance}, 63:\penalty0 101757, 2022.
\newblock \doi{10.1016/j.ribaf.2022.101757}.

\bibitem[Haberman and Renshaw(1996)]{haberman1996generalized}
Haberman, S. and Renshaw, A.~E.
\newblock Generalized linear models and actuarial science.
\newblock \emph{Journal of the Royal Statistical Society: Series D (The Statistician)}, 45\penalty0 (4):\penalty0 407--436, 1996.
\newblock \doi{10.2307/2988543}.

\bibitem[Hastie et~al.(2009)Hastie, Tibshirani, Friedman, et~al.]{hastie2009elements}
Hastie, T., Tibshirani, R., Friedman, J., et~al.
\newblock \emph{The elements of statistical learning}.
\newblock Springer, New York, 2009.
\newblock ISBN 978-0-387-84857-0.
\newblock \doi{10.1007/978-0-387-84858-7}.

\bibitem[He et~al.(2016)He, Zhang, Ren, and Sun]{he2016deep}
He, K., Zhang, X., Ren, S., and Sun, J.
\newblock Deep residual learning for image recognition.
\newblock In \emph{Proceedings of the IEEE conference on computer vision and pattern recognition}, pages 770--778, 2016.
\newblock \doi{10.1109/CVPR.2016.90}.

\bibitem[Henckaerts and Antonio(2022)]{henckaerts2022added}
Henckaerts, R. and Antonio, K.
\newblock The added value of dynamically updating motor insurance prices with telematics collected driving behavior data.
\newblock \emph{Insurance: Mathematics and Economics}, 105:\penalty0 79--95, 2022.
\newblock \doi{10.1016/j.insmatheco.2022.03.011}.

\bibitem[Henckaerts et~al.(2021)Henckaerts, C{\^o}t{\'e}, Antonio, and Verbelen]{henckaerts2021boosting}
Henckaerts, R., C{\^o}t{\'e}, M.-P., Antonio, K., and Verbelen, R.
\newblock Boosting insights in insurance tariff plans with tree-based machine learning methods.
\newblock \emph{North American Actuarial Journal}, 25\penalty0 (2):\penalty0 255--285, 2021.
\newblock \doi{10.1080/10920277.2020.1745656}.

\bibitem[Holvoet et~al.(2025)Holvoet, Antonio, and Henckaerts]{holvoet2025neural}
Holvoet, F., Antonio, K., and Henckaerts, R.
\newblock Neural networks for insurance pricing with frequency and severity data: a benchmark study from data preprocessing to technical tariff.
\newblock \emph{North American Actuarial Journal}, pages 1--44, 2025.
\newblock \doi{10.1080/10920277.2025.2451860}.

\bibitem[Ibrahim et~al.(2024)Ibrahim, Stanley, Murfi, Novkaniza, and Devila]{ibrahim2024evaluating}
Ibrahim, J., Stanley, J., Murfi, H., Novkaniza, F., and Devila, S.
\newblock Evaluating xgboost for competitive insurance pricing: A case study on motor third-party liability insurance.
\newblock In \emph{2024 International Conference on Intelligent Cybernetics Technology \& Applications (ICICyTA)}, pages 847--852. IEEE, 2024.
\newblock \doi{10.1109/icicyta64807.2024.10912952}.

\bibitem[Islam et~al.(2024)Islam, Ahamed, Matsushita, and Noguchi]{islam2024damage}
Islam, M.~M., Ahamed, T., Matsushita, S., and Noguchi, R.
\newblock A damage-based crop insurance system for flash flooding: a satellite remote sensing and econometric approach.
\newblock In \emph{Remote sensing application II: A climate change perspective in agriculture}, pages 121--163. Springer, 2024.
\newblock \doi{10.1007/978-981-97-1188-8\_5}.

\bibitem[{ISO}(2022)]{ISO19109_2022}
{ISO}.
\newblock {ISO 19109:2022 Geographic information -- Rules for application schema}.
\newblock Standard, International Organization for Standardization, Geneva, Switzerland, 2022.

\bibitem[Jaiswal(2023)]{jaiswal2023impact}
Jaiswal, R.
\newblock Impact of ai in the general insurance underwriting factors.
\newblock \emph{Central European Management Journal}, 31\penalty0 (2):\penalty0 697--705, 2023.

\bibitem[Kita-Wojciechowska and Kidzi{\'n}ski(2019)]{kita2019google}
Kita-Wojciechowska, K. and Kidzi{\'n}ski, {\L}.
\newblock Google street view image predicts car accident risk.
\newblock \emph{Central European Economic Journal}, 6\penalty0 (53):\penalty0 151--163, 2019.
\newblock \doi{10.2478/ceej-2019-0011}.

\bibitem[Lecun et~al.(1998)Lecun, Bottou, Bengio, and Haffner]{LecunY.1998Glat}
Lecun, Y., Bottou, L., Bengio, Y., and Haffner, P.
\newblock Gradient-based learning applied to document recognition.
\newblock \emph{Proceedings of the IEEE}, 86\penalty0 (11):\penalty0 2278--2324, 1998.
\newblock ISSN 0018-9219.
\newblock \doi{10.1109/5.726791}.

\bibitem[Li et~al.(2021)Li, Liu, Yang, Peng, and Zhou]{li2021survey}
Li, Z., Liu, F., Yang, W., Peng, S., and Zhou, J.
\newblock A survey of convolutional neural networks: analysis, applications, and prospects.
\newblock \emph{IEEE transactions on neural networks and learning systems}, 33\penalty0 (12):\penalty0 6999--7019, 2021.
\newblock \doi{10.1109/tnnls.2021.3084827}.

\bibitem[Longley et~al.(2015)Longley, Goodchild, Maguire, and Rhind]{longley2015geographic}
Longley, P.~A., Goodchild, M.~F., Maguire, D.~J., and Rhind, D.~W.
\newblock \emph{Geographic information science and systems}.
\newblock John Wiley \& Sons, 2015.

\bibitem[McCullagh(2019)]{mccullagh2019generalized}
McCullagh, P.
\newblock \emph{Generalized linear models}.
\newblock Routledge, 2019.
\newblock \doi{10.1201/9780203753736}.

\bibitem[Nguyen et~al.(2022)Nguyen, Belnap, Dwivedi, Deligani, Kumar, Li, Whitaker, Keralis, Mane, Yue, et~al.]{nguyen2022google}
Nguyen, Q.~C., Belnap, T., Dwivedi, P., Deligani, A. H.~N., Kumar, A., Li, D., Whitaker, R., Keralis, J., Mane, H., Yue, X., et~al.
\newblock Google street view images as predictors of patient health outcomes, 2017--2019.
\newblock \emph{Big data and cognitive computing}, 6\penalty0 (1):\penalty0 15, 2022.
\newblock \doi{10.3390/bdcc6010015}.

\bibitem[Noll et~al.(2020)Noll, Salzmann, and Wuthrich]{noll2020case}
Noll, A., Salzmann, R., and Wuthrich, M.~V.
\newblock Case study: French motor third-party liability claims.
\newblock \emph{Available at SSRN 3164764}, 2020.
\newblock \doi{10.2139/ssrn.3164764}.

\bibitem[Nussbaum et~al.(2024)Nussbaum, Duderstadt, and Mulyar]{nussbaum2024nomic}
Nussbaum, Z., Duderstadt, B., and Mulyar, A.
\newblock Nomic embed vision: Expanding the latent space.
\newblock \emph{arXiv preprint arXiv:2406.18587}, 2024.
\newblock \doi{10.48550/arXiv.2406.18587}.

\bibitem[{OpenStreetMap contributors}(2025{\natexlab{a}})]{openstreetmap}
{OpenStreetMap contributors}.
\newblock {OpenStreetMap}, 2025{\natexlab{a}}.
\newblock URL \url{https://www.openstreetmap.org}.
\newblock Data licensed under the Open Database License (ODbL).

\bibitem[{OpenStreetMap contributors}(2025{\natexlab{b}})]{osmBelgium2025}
{OpenStreetMap contributors}.
\newblock {OpenStreetMap Belgium Data Extract}.
\newblock Geofabrik GmbH, 2025{\natexlab{b}}.
\newblock URL \url{https://download.geofabrik.de/europe/belgium.html}.
\newblock Distributed by Geofabrik. Licensed under ODbL.

\bibitem[{\'O}skarsd{\'o}ttir et~al.(2022){\'O}skarsd{\'o}ttir, Ahmed, Antonio, Baesens, Dendievel, Donas, and Reynkens]{oskarsdottir2022social}
{\'O}skarsd{\'o}ttir, M., Ahmed, W., Antonio, K., Baesens, B., Dendievel, R., Donas, T., and Reynkens, T.
\newblock Social network analytics for supervised fraud detection in insurance.
\newblock \emph{Risk Analysis}, 42\penalty0 (8):\penalty0 1872--1890, 2022.
\newblock \doi{10.1111/risa.13693}.

\bibitem[P{\'e}rez-Zarate et~al.(2024)P{\'e}rez-Zarate, Corzo-Garc{\'\i}a, Pro-Mart{\'\i}n, {\'A}lvarez-Garc{\'\i}a, Mart{\'\i}nez-del Amor, and Fern{\'a}ndez-Cabrera]{perez2024automated}
P{\'e}rez-Zarate, S.~A., Corzo-Garc{\'\i}a, D., Pro-Mart{\'\i}n, J.~L., {\'A}lvarez-Garc{\'\i}a, J.~A., Mart{\'\i}nez-del Amor, M.~A., and Fern{\'a}ndez-Cabrera, D.
\newblock Automated car damage assessment using computer vision: Insurance company use case.
\newblock \emph{Applied Sciences}, 14\penalty0 (20):\penalty0 9560, 2024.
\newblock \doi{10.3390/app14209560}.

\bibitem[Qazvini(2019)]{qazvini2019validation}
Qazvini, M.
\newblock On the validation of claims with excess zeros in liability insurance: A comparative study.
\newblock \emph{Risks}, 7\penalty0 (3):\penalty0 71, 2019.
\newblock \doi{10.3390/risks7030071}.

\bibitem[Rababaah(2023)]{rababaah2023investigation}
Rababaah, A.~R.
\newblock Investigation of deep learning models for vehicle damage classification.
\newblock In \emph{2023 10th International Conference on Signal Processing and Integrated Networks (SPIN)}, pages 25--30. IEEE, 2023.
\newblock \doi{10.1109/spin57001.2023.10116703}.

\bibitem[Seyam(2025)]{seyam2025predicting}
Seyam, E.~A.
\newblock Predicting motor insurance claim incidence using generalized and tree-based models: A comparative statistical approach.
\newblock \emph{Insurance Markets and Companies}, 16\penalty0 (2):\penalty0 38, 2025.
\newblock \doi{10.21511/ins.16(2).2025.04}.

\bibitem[Stevenson et~al.(2022)Stevenson, Mues, and Bravo]{STEVENSON2022378}
Stevenson, M., Mues, C., and Bravo, C.
\newblock Deep residential representations: Using unsupervised learning to unlock elevation data for geo-demographic prediction.
\newblock \emph{ISPRS Journal of Photogrammetry and Remote Sensing}, 187:\penalty0 378--392, 2022.
\newblock ISSN 0924-2716.
\newblock \doi{https://doi.org/10.1016/j.isprsjprs.2022.03.015}.
\newblock URL \url{https://www.sciencedirect.com/science/article/pii/S0924271622000880}.

\bibitem[Thiran and Thomas(1997)]{thiran1997accidents}
Thiran, P. and Thomas, I.
\newblock Accidents de la route et distance au domicile. approche quantitative pour bruxelles.
\newblock \emph{Les Cahiers Scientifiques du Transport-Scientific Papers in Transportation}, 32, 1997.
\newblock \doi{10.46298/cst.11958}.

\bibitem[Tufvesson et~al.(2019)Tufvesson, Lindstr{\"o}m, and Lindstr{\"o}m]{tufvesson2019spatial}
Tufvesson, O., Lindstr{\"o}m, J., and Lindstr{\"o}m, E.
\newblock Spatial statistical modelling of insurance risk: a spatial epidemiological approach to car insurance.
\newblock \emph{Scandinavian Actuarial Journal}, 2019\penalty0 (6):\penalty0 508--522, 2019.
\newblock \doi{10.1080/03461238.2019.1576146}.

\bibitem[Vít et~al.(2025)Vít, Seif, and Štěpánek]{OndřejVít2025CFEi}
Vít, O., Seif, L., and Štěpánek, L.
\newblock Claim frequency estimation in motor third-party liability (mtpl): Classical statistical models versus machine learning methods.
\newblock In \emph{Annals of Computer Science and Information Systems}, volume~45, pages 161--166. Polish Information Processing Society, 2025.
\newblock \doi{10.15439/2025f5118}.

\bibitem[Zail(2019)]{zail2019predictive}
Zail, H.
\newblock Predictive analytics in long term care.
\newblock In \emph{Actuarial Aspects of Long Term Care}, pages 309--336. Springer, 2019.
\newblock \doi{10.1007/978-3-030-05660-5\_13}.

\bibitem[Zhang et~al.(2023)Zhang, Lipton, Li, and Smola]{zhang2023dive}
Zhang, A., Lipton, Z.~C., Li, M., and Smola, A.~J.
\newblock \emph{Dive into deep learning}.
\newblock Cambridge University Press, 2023.

\bibitem[Zou and Hastie(2005)]{zou2005regularization}
Zou, H. and Hastie, T.
\newblock Regularization and variable selection via the elastic net.
\newblock \emph{Journal of the Royal Statistical Society Series B: Statistical Methodology}, 67\penalty0 (2):\penalty0 301--320, 2005.
\newblock \doi{10.1111/j.1467-9868.2005.00503.x}.

\end{thebibliography}
\end{document}